\def\LRDcleanversion{1}
\documentclass{article}

\usepackage[main,final]{neurips_2026}

\usepackage[utf8]{inputenc} % allow utf-8 input
\usepackage[T1]{fontenc}    % use 8-bit T1 fonts
\usepackage{hyperref}       % hyperlinks
\usepackage{url}            % simple URL typesetting
\usepackage{booktabs}       % professional-quality tables
\usepackage{amsfonts}       % blackboard math symbols
\usepackage{nicefrac}       % compact symbols for 1/2, etc.
\usepackage{microtype}      % microtypography
\usepackage{xcolor}         % colors

\usepackage{graphicx}
\usepackage{subcaption}

\usepackage{hyperref}

\usepackage{xspace}
\usepackage{comment}
\usepackage{xcolor}        
\usepackage{enumitem}
\usepackage{lscape}
\usepackage{adjustbox}
\usepackage{tabularx}
\usepackage{graphicx}
\usepackage{float} 
\usepackage{placeins}
\usepackage{color}
\usepackage{colortbl}
\usepackage{multirow}

\usepackage[utf8]{inputenc}
\usepackage[T1]{fontenc}
\usepackage{url}
\usepackage{booktabs}
\usepackage{amsfonts}
\usepackage{nicefrac}
\usepackage{microtype}
\usepackage{makecell}
\usepackage{wrapfig} % 前导加上这个包
\usepackage{arydshln}
\definecolor{LightCyan}{rgb}{0.88,1,1}
\definecolor{LightRed}{rgb}{1,0.5,0.5}
\definecolor{LightYellow}{rgb}{1,1,0.88}
\definecolor{Grey}{rgb}{0.75,0.75,0.75}
\definecolor{DarkGrey}{rgb}{0.55,0.55,0.55}
\definecolor{LightGreen}{rgb}{0.0, 0.70, 0.0}

\newif\ifshowchanges
\showchangesfalse
\ifdefined\LRDblueversion\showchangestrue\fi
\ifdefined\LRDcleanversion\showchangesfalse\fi
\newcommand{\rev}[1]{{\ifshowchanges\color{blue}\fi #1}}
\newenvironment{revision}{\par\begingroup\ifshowchanges\color{blue}\fi}{\par\endgroup}

\makeatletter
\@for\lrd@key:=hersche2025softmasked,zhong2026evotoken,hu2026rcd,kim2026stveto,wang2025creditdecoding,deschenaux2026psi,sahoo2025duality,cao2026soar,fu2026nemotron,diffusiongemma2026,rein2023gpqa,suzgun2022bbh\do{%
  \expandafter\gdef\csname lrd@newref@\lrd@key\endcsname{1}}
\def\lrd@markedbibitem[#1]#2{%
  \lrd@originalbibitem[#1]{#2}%
  \ifshowchanges
    \ifcsname lrd@newref@#2\endcsname\color{blue}\else\color{black}\fi
  \fi
  \ignorespaces}
\AtBeginDocument{%
  \let\lrd@originalbibitem\bibitem
  \renewcommand\bibitem{\@ifnextchar[{\lrd@markedbibitem}{\lrd@markedbibitem[]}}}
\makeatother

\usepackage{amsmath}
\usepackage{amssymb}
\usepackage{mathtools}
\usepackage{amsthm}
\DeclareMathOperator*{\argmax}{arg\,max}

\theoremstyle{plain}
\newtheorem{theorem}{Theorem}[section]

\newtheorem{lemma}[theorem]{Lemma}
\newtheorem{corollary}[theorem]{Corollary}
\theoremstyle{definition}
\newtheorem{definition}[theorem]{Definition}

\theoremstyle{remark}
\newtheorem{remark}[theorem]{Remark}

\title{Latent Refinement Decoding: Enhancing Diffusion Language Models by Refining Belief States}

\author{%
  \rev{\bfseries Qinglin Zhu$^1$\thanks{\rev{Equal contribution.\quad \textsuperscript{\(\dagger\)}Corresponding author.}} \quad Yizhen Yao$^1$\footnotemark[1] \quad Runcong Zhao$^1$\footnotemark[2] \quad Yanzheng Xiang$^1$}\\
  \rev{\bfseries Siya Qi$^1$ \quad Amrutha Saseendran$^3$ \quad Chen Jin$^3$ \quad Philip Teare$^3$}\\
  \rev{\bfseries Bin Liang$^4$ \quad Yulan He$^{1,2}$ \quad Lin Gui$^1$}\\[3pt]
  \rev{\normalfont\small $^1$King's College London, UK \quad $^2$The Alan Turing Institute, UK}\\
  \rev{\normalfont\small $^3$AstraZeneca, UK \quad $^4$The Chinese University of Hong Kong}\\[2pt]
  \rev{\normalfont\footnotesize\texttt{\{qinglin.1.zhu,runcong.zhao,lin.1.gui\}@kcl.ac.uk}}
}

\begin{document}

\maketitle

\begin{abstract}
Autoregressive (AR) models remain the standard for natural language generation but still suffer from high latency due to strictly sequential decoding.
Recent diffusion-inspired approaches, such as LLaDA and Dream, mitigate this by generating in parallel, yet they suffer from two core limitations: information loss, as predictive distributions for non-finalised tokens are discarded at each step, and a lack of well-behaved commitment dynamics, where local decisions are not properly coordinated at the global level.
We introduce Latent Refinement Decoding (LRD), a two-stage framework with \textit{Latent Refinement} and a \textit{Predictive Feedback Loop}. The first stage maintains masked positions as distributional mixtures of predicted tokens and the mask embedding, allowing the model to establish more globally consistent beliefs. The second stage progressively finalises confident tokens while retaining uncertain ones for iterative feedback. KL-divergence dynamics provide an effective and stable criterion for convergence and early stopping.
Experiments show LRD improves accuracy in coding (HumanEval +6.3, MBPP +2.6) and reasoning (GSM8K +2.9, MATH500 +3.8), while delivering up to 10.6$\times$ speedup overall.
LRD is compatible with single-token decoding, multi-token threshold commitment, and dynamic/system accelerators; Fast-dLLM+LRD reaches up to 18.4$\times$ speedup over vanilla decoding while improving accuracy.
\end{abstract}

\section{Introduction}

Autoregressive (AR) models have long defined the standard for natural language generation~\citep{brown2020language, fei2025advancing,achiam2023gpt, yang2025qwen3}, but their inherently sequential token-by-token decoding imposes a fundamental bottleneck on inference latency~\citep{touvron2023llama,sun2024moss}. This constraint has motivated the development of parallel decoding paradigms. Among them, diffusion-inspired approaches such as \textit{LLaDA}~\citep{nie2025largelanguagediffusionmodels} and \textit{Dream}~\citep{ye2025dream} offer a particularly promising direction. By formulating text generation as an iterative refinement process and updating all token positions in parallel at each step, these methods provide a compelling alternative to traditional AR decoding, achieving significant speedups while maintaining competitive quality~\citep{labs2025mercuryultrafastlanguagemodels, deepmind2025geminidiffu}.
Despite recent progress, diffusion language models employ hard assignment strategies~\citep{DBLP:conf/iclr/GongA0YZLAZB00K25,nie2025largelanguagediffusionmodels,ye2025dream}: at each denoising step, they commit high-confidence positions to specific tokens while resetting remaining positions to uniform \texttt{[MASK]} tokens. The predictive distributions from earlier steps are discarded, limiting the model's ability to build upon partial beliefs established in earlier iterations. 

This design introduces two limitations: (i) \textbf{Information loss from hard masking}~\citep{li2024promisespitfallsgenerativemasked}: At each denoising step, positions below confidence thresholds are reset to uniform \texttt{[MASK]} embeddings, completely discarding their predictive distributions. %\wrong{This prevents uncertain positions from sharing probabilistic information through self-attention, forcing each masked position to be predicted in isolation(point1)} \reviewone{
This prevents uncertain positions from sharing probabilistic information through self-attention across steps, forcing them to be repeatedly predicted from the same  
\texttt{[MASK]} state. When mispredictions occur, the hard assignment yields infinite KL divergence from the true posterior, as it assigns zero probability mass to the correct token.
(ii) \textbf{Lack of well-behaved commitment dynamics}~\citep{luxembourg2025planspeeddilatedscheduling,li2025convergencetheorydiffusionlanguage}: The binary nature of hard assignment creates a dilemma: 
\textbf{aggressive} selection commits early and can lock in incorrect predictions, propagating errors through later steps; \textbf{conservative} selection keeps many positions masked, which slows progress and requires many denoising iterations.
Moreover, using a fixed number of iterations ignores the varying complexity across different generation tasks, wasting computation on simple cases while potentially underserving complex ones.

\begin{figure*}
    \centering
    \includegraphics[width=0.9 \linewidth]{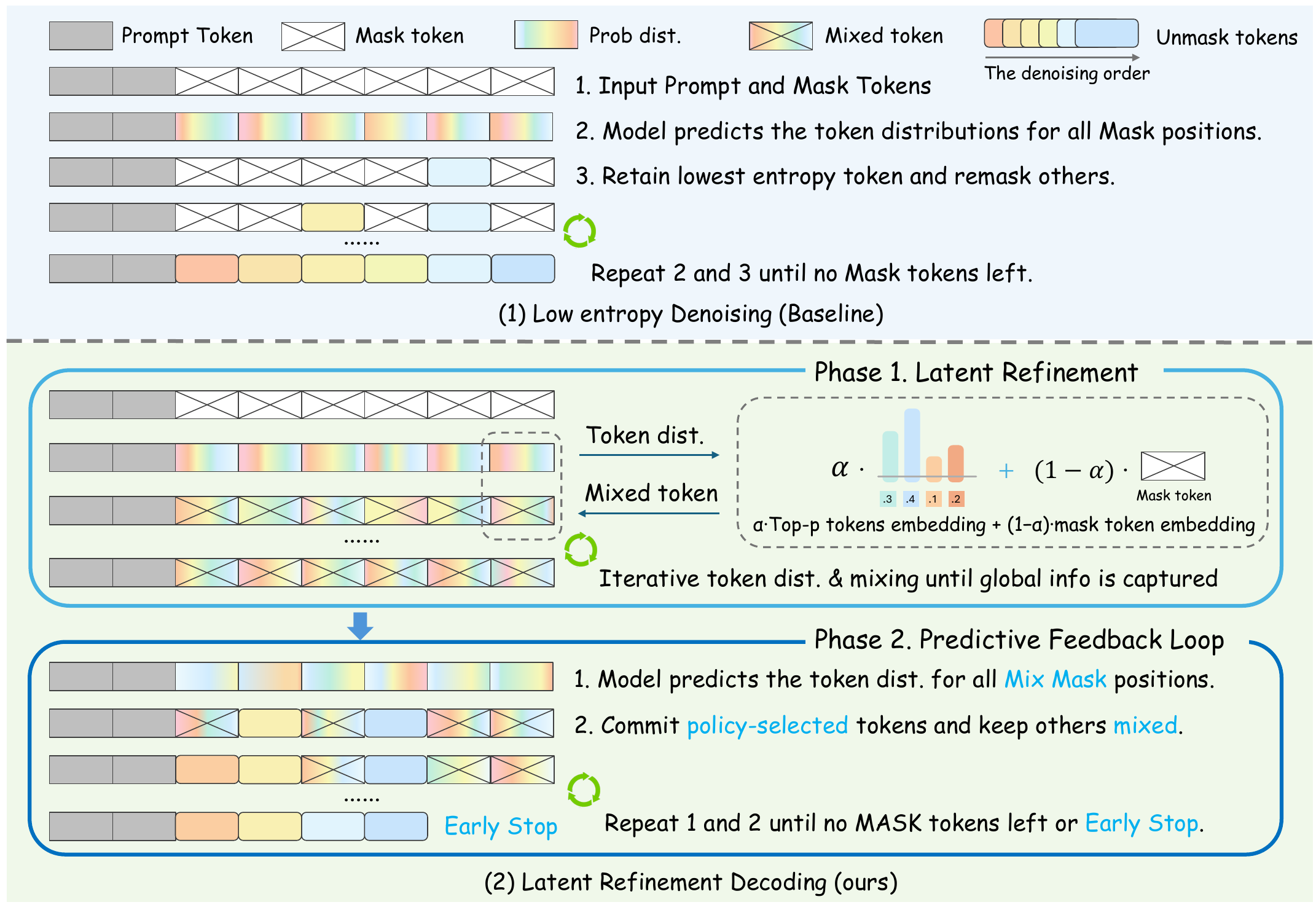}
\caption{Comparison between the existing decoding strategy and the proposed method. Different \textcolor{red}{c}\textcolor{orange}{o}\textcolor{blue}{l}\textcolor{green}{o}\textcolor{magenta}{u}\textcolor{purple}{r}s represent distinct tokens, while gradient colours indicate predicted token representations. \textbf{Top:} In the existing strategy, all \texttt{[MASK]} tokens share the same embedding and are repeatedly remasked if not selected. \textbf{Bottom:} In LRD, Phase 1 refines each \texttt{[MASK]} embedding, and Phase 2 progressively commits confident tokens while keeping uncertain ones soft for context-aware decoding.}

    \label{fig:framework}
\end{figure*}

To overcome these limitations, we move beyond purely discrete denoising and introduce \textbf{Latent Refinement Decoding (LRD)}, a hybrid framework that operates in both embedding and token spaces. LRD restructures the denoising process into two coordinated stages. \textbf{Phase 1: Latent Refinement} performs \rev{belief-guided} updates entirely in the embedding space: for each masked position, we form a mix embedding by mixing the \texttt{[MASK]} embedding with the entropy-normalised expectation over top-$p$ predicted token embeddings, allowing
the model to ``think latently'' in continuous embedding space, establishing globally coherent beliefs before committing to discrete decisions.
Once the predictive distributions stabilise, \textbf{Phase 2: Predictive Feedback Loop} progressively converts policy-selected confident positions into discrete tokens while keeping the remaining positions in soft form, feeding each step’s predictions back into the next soft mixture; KL-based monitors govern the soft-to-hard transition and enable adaptive early stopping.
\textbf{The main contributions of LRD are:}

\begin{enumerate} 

\item \textbf{Soft diffusion} that enables continuous denoising in embedding space by mixing \texttt{[MASK]} with weighted token representations. This preserves distributional information across steps and enables cross-position refinement through self-attention.

\item \textbf{Adaptive two-phase sampling} that combines soft refinement for global coherence with hard decoding for precise convergence. KL-based monitoring enables automatic phase transitions and early stopping based on actual convergence rather than fixed iteration counts.

\item We validate LRD across diverse model families, generation lengths, and benchmarks spanning coding (HumanEval: +6.3, MBPP: +2.6) and reasoning (GSM8K: +2.9, MATH500: +3.8), and general language tasks (reading comprehension and summarisation), improving accuracy across all domains and achieving speedups of up to 10.6$\times$.

\item \textbf{Commitment-policy composability.} LRD supports single-token decoding, multi-token threshold commitment, and dynamic/system accelerators such as Fast-dLLM and DAEDAL. With Fast-dLLM, LRD reaches 15.2$\times$ speedup on Dream-Base and 18.4$\times$ on LLaDA-Ins, showing its value as a drop-in decoding module.

\end{enumerate}

\section{Preliminary}
For dLLMs \citep{ou2024your, zheng2024reparameterized, shi2024simplified, DBLP:conf/iclr/GongA0YZLAZB00K25}, 
the forward process corrupts data $\mathbf{x}_0 \in \{1, ..., |V|\}^L$ (a sequence of $L$ tokens from vocabulary size $|V|$) 
into progressively noisier versions $\mathbf{x}_1, ..., \mathbf{x}_T$. 
At each timestep, the forward process is defined as a categorical distribution:
\begin{equation}
q(\mathbf{x}_t|\mathbf{x}_{t-1}) = \text{Cat}(\mathbf{x}_t; \mathbf{Q}_t^\top \mathbf{x}_{t-1})
\end{equation}
where $\mathbf{x}_t \in \{0,1\}^{|V| \times L}$ is the one-hot representation of tokens at time $t$, 
and $\mathbf{Q}_t \in [0,1]^{|V| \times |V|}$ is the transition matrix. 
Each token either remains unchanged with probability $1-\beta_t$ or transitions to the special \texttt{[MASK]} token with probability $\beta_t \in (0,1)$: 
$\mathbf{Q}_t = (1 - \beta_t)\mathbf{I} + \beta_t \mathbf{1}\mathbf{m}^\top$, 
where $\mathbf{I} \in \mathbb{R}^{|V| \times |V|}$ is the identity matrix, 
$\mathbf{1} \in \mathbb{R}^{|V|}$ is an all-ones vector, 
and $\mathbf{m} \in \{0,1\}^{|V|}$ is the one-hot encoding of \texttt{[MASK]}. 
The cumulative transition matrix from $\mathbf{x}_0$ to $\mathbf{x}_t$ is
$\overline{\mathbf{Q}}_t = \alpha_t^*\mathbf{I} + (1-\alpha_t^*)\mathbf{1}\mathbf{m}^\top$,
where $\alpha_t^* = \prod_{s=1}^{t}(1-\beta_s)$ denotes the probability that a token remains unmasked from step 0 to step $t$ (for $t=1,\dots,T$).
% The cumulative transition matrix from $\mathbf{x}_0$ to $\mathbf{x}_t$ becomes: 
% $\overline{\mathbf{Q}}_t = \alpha_t^*\mathbf{I} + (1-\alpha_t^*)\mathbf{1}\mathbf{m}^\top$, 
% where $\alpha_t^* = \prod_{s=1}^{t}(1-\beta_s)$ is the probability of a token remaining unmasked from time 0 to time $t$.

The reverse process $p_\theta(\mathbf{x}_{t-1}|\mathbf{x}_t)$ aims to reconstruct the original data by iteratively denoising from $\mathbf{x}_T$ (fully masked) to $\mathbf{x}_0$ (clean text). At each denoising step $t$, the model predicts a distribution over tokens for each position: $p_\theta^{(i)}(\mathbf{x}_0|\mathbf{x}_t)$ for position $i$.

In transformer-based diffusion models, each token is represented by a learnable embedding vector. Let $\mathbf{e}_v \in \mathbb{R}^d$ denote the embedding for token $v \in V$, and $\mathbf{e}_{[\text{MASK}]} \in \mathbb{R}^d$ the embedding for the \texttt{[MASK]} token. During the reverse process, traditional sampling strategies employ \textbf{hard assignment}, selecting tokens based on prediction confidence:
\begin{small}
\begin{equation}
v_t^{(i)} =
\begin{cases}
\argmax_{v \in V} \; p_\theta^{(i)}(v|\mathbf{x}_t), 
& \text{if } i \in \text{top-1}(\{H_t^{(j)}\}_{j=1}^L) \\
\texttt{[MASK]}, & \text{otherwise}
\end{cases}
\label{eq:hard_assignment}
\end{equation}
\end{small}

where $H_t^{(j)} = -\sum_{v} p_\theta^{(j)}(v|\mathbf{x}_t) \log p_\theta^{(j)}(v|\mathbf{x}_t)$ is the entropy at position $j$, and top-$1$ selects the position with lowest entropy (highest confidence). 
This creates a binary embedding assignment: each position uses either $\mathbf{e}_{[\text{MASK}]}$ or a specific token embedding $\mathbf{e}_{v_t^{(i)}}$, resulting in complete information loss for positions not selected. This binary decision mechanism creates a discontinuous mapping from probability distributions to discrete embeddings: positions below the confidence threshold are reset to pure \texttt{[MASK]} embeddings, completely discarding their distributional information $p_\theta(\cdot|\mathbf{x}_t)$, resulting in abrupt information loss and suboptimal exploration of the posterior distribution.

\section{Methodology}\label{sec:method}

Figure~\ref{fig:framework} illustrates the two-phase LRD pipeline. Effective discrete diffusion sampling requires maintaining sufficient uncertainty for exploration while incorporating token-specific information for convergence. To achieve this balance, we propose LRD: instead of binary decisions that abruptly switch between pure noise (\texttt{[MASK]}) and deterministic tokens, we create intermediate representations through continuous embedding interpolation. Specifically, we construct mixed embeddings that blend \texttt{[MASK]} and token embeddings weighted by prediction uncertainty, where high-entropy positions retain more mask-like characteristics (preserving exploration) while low-entropy positions incorporate more token information (enabling commitment). This enables a gradual denoising trajectory where the noise-signal ratio smoothly decreases, yielding better-calibrated probability distributions for subsequent sampling steps.

\subsection{Soft Diffusion}\label{sec:soft_diffu}
Our method operates in the embedding space rather than discrete token space. At each timestep $t$, we maintain a set of \textit{soft embeddings} $\mathcal{E}_t = \{\tilde{\mathbf{e}}_t^{(1)}, \ldots, \tilde{\mathbf{e}}_t^{(L)}\}$ where
\begin{small}
\begin{equation}
\tilde{\mathbf{e}}_t^{(i)} = (1-\alpha_t^{(i)}) \cdot \mathbf{e}_{\texttt{[MASK]}} + \alpha_t^{(i)} \cdot \sum_{v \in \mathcal{T}_{t+1}^{(i)}} \bar{p}_{t+1}^{(i)}(v) \cdot \mathbf{e}_v
\label{eq:mix-embedding}
\end{equation}
\end{small}
Here, $\mathbf{e}_v \in \mathbb{R}^d$ denotes the embedding of the token $v \in \mathcal{T}_{t+1}^{(i)}$, where $\mathcal{T}_{t+1}^{(i)}$ is the top-$p$ nucleus set. $\mathbf{e}_{\texttt{[MASK]}}$ is the \texttt{[MASK]} embedding, 
$\bar{p}_{t+1}^{(i)}(v)$ denotes the probability mass of token $v$ at position $i$, renormalised to the nucleus set $\mathcal{T}_{t+1}^{(i)}$.
% and $p_{t+1}^{(i)}(v)$ is the predicted probability of token $v$ at position $i$ from the previous timestep.
The coefficient $\alpha_t^{(i)} \in [0,1]$ controls the interpolation strength.
The mixing weight $\alpha_t$ is controlled by entropy: 
\begin{small}
\begin{equation}
\alpha_t^{(i)} = r_f \cdot ( 1 - \bar{H}_{t+1}^{(i)} ) = r_f \cdot \left(1 + \frac{\sum_{k=1}^{|V|} p_{t+1}^{(i)}(k) \log p_{t+1}^{(i)}(k)}{\log |V|}\right)
\label{eq:alpha}
\end{equation}
\end{small}
where $p_{t+1}^{(i)}(k)$ refers to the probability distribution over the full vocabulary, $\bar{H}_{t+1}^{(i)}$ is the normalised entropy of this distribution, and $r_f \in (0,1]$ sets the maximum interpolation strength. Since the entropy of a categorical distribution over a vocabulary of size $|V|$ lies in $[0,\log |V|]$, we divide by $\log |V|$ to normalise it into $[0,1]$.
This design ensures that uncertain positions stay mask-like while confident ones commit to tokens. Formal justification and stability analysis are deferred to Appendix~\ref{appendix:stability}.

Consider the absorbing discrete diffusion process where the true posterior distribution $q^*(x_{t-1}|x_t, x_0)$ represents optimal denoising. For masked positions where $x_t^{(i)} = \texttt{[MASK]}$, Bayes' rule yields:
\begin{align}
q^*(x_{t-1}^{(i)}&|x_t^{(i)}=\texttt{[MASK]}, x_0^{(i)}) \notag \\
&= \frac{\alpha_{t-1}^* - \alpha_t^*}{1 - \alpha_t^*}\delta_{x_0^{(i)}} + \frac{1 - \alpha_{t-1}^*}{1 - \alpha_t^*}\delta_{\texttt{[MASK]}}
\label{eq:true-posterior}
\end{align}
where the detailed derivation is provided in Appendix~\ref{appendix:posterior-derivation}. Hard assignment approximates this by a degenerate distribution $q_{\mathrm{hard}} \in \{\delta_{\hat{x}_0^{(i)}}, \delta_{\texttt{[MASK]}}\}$, where $\hat{x}_0^{(i)}$ denotes the model's predicted estimate of the original token $x_0^{(i)}$. 
If $\hat{x}_0^{(i)} \neq x_0^{(i)}$, then $q_{\mathrm{hard}}$ assigns zero probability where $q^*$ is positive, leading to $\mathrm{KL}(q^*\|q_{\mathrm{hard}})=\infty$. Moreover, positions that remain masked are represented by a fixed embedding $\mathbf{e}_{\texttt{[MASK]}}$, which conveys no distributional information to neighbouring positions.

Latent Refinement Decoding mitigates both issues. First, \rev{the soft mixture retains probability mass on alternative candidates within the top-$p$ set, including the correct token when it remains in that set.} Second, the weighted mixture $\sum_v \bar{p}_t^{(i)}(v)\mathbf{e}_v$ can be viewed as the expected embedding under the model’s belief at position $i$. Since self-attention propagates these mixtures through weighted aggregation, the soft mixture \rev{can retain} partial information about the correct token, which can be transmitted to other positions via attention and guide contextual disambiguation.

\subsection{Adaptive Sampling with Soft-to-Hard Scheduling}
The optimal denoising strategy must balance two objectives: preserving sufficient uncertainty for exploration while progressively reducing entropy for convergence. Latent Refinement Decoding provides a smooth relaxation in the embedding space, where continuous embedding updates are well behaved and %\wrong{guarantee contraction toward} 
\rev{often stabilise empirically. } However, Latent Refinement Decoding cannot fully collapse distributions to one-hot states, since embeddings always encode mixtures rather than discrete commitments. As a result, convergence slows in later stages when sharper updates are required for final token generation.

To overcome this limitation, we adopt a two-phase schedule. \textbf{Phase 1} \rev{refines the soft state towards approximate self-consistency}. Once the model’s predictive distributions stabilise ($\mathcal{D}_{\text{KL}}^{(t)} < \tau_{\text{refine}}$), \textbf{Phase 2} transitions to hard assignment, which \rev{enables discrete commitment from the refined state}. This design \rev{is analogous to} graduated optimisation: begin with a smooth relaxation to encourage global exploration, then progressively sharpen the objective to encourage convergence.

\paragraph{Phase 1: Latent Refinement via Soft Embeddings.}
During the initial refinement phase, the model iteratively refines predictive distributions through soft embedding propagation without committing to any discrete tokens. Starting from $t=T$ (fully masked), we compute soft embeddings using Equation~\ref{eq:mix-embedding}, where predictions $p_{t}^{(i)}(v)$ are conditioned on the prior soft embeddings rather than discrete tokens.

As refinement progresses, the soft embeddings approach a self-consistent state of Eq.~\ref{eq:mix-embedding}, where the model's predictions become {aligned with the soft context}, i.e., the output distribution given the current soft embeddings closely matches the distribution encoded in those embeddings. \rev{Small changes in consecutive predictions provide a practical stopping signal, without certifying correctness or exhaustion of all useful information.} We detect this saturation by monitoring the KL divergence between consecutive predictions:
\begin{equation}
\mathcal{D}_{\text{KL}}^{(t)} = \frac{1}{L}\sum_{i=1}^{L} D_{\text{KL}}(p_{t}^{(i)} \| p_{t+1}^{(i)})
\label{eq:kl-gap}
\end{equation}
When $\mathcal{D}_{\text{KL}}^{(t)} < \tau_{\text{refine}}$, \rev{we treat the predictive state as approximately stable and switch to discrete commitments}.

This triggers the transition to Phase 2, where discrete token generation can exploit the well-initialised distributions from Phase 1 (Appendix~\ref{appendix:fixed_point_view}).
Alternatively, if convergence is not achieved within $T_{\text{refine}}$ steps, we still transition to Phase 2 for computational efficiency, as extended refinement shows diminishing returns while incurring additional computational cost.

\paragraph{Phase 2: Predictive Feedback Loop.}
Once convergence is detected at timestep $t^*$, we switch to Predictive Feedback decoding for the remaining timesteps $t \in [t^*, 0]$. Let $\mathcal C_t$ be the commitment set selected by the decoding policy; by default, LRD uses the lowest-entropy position, while multi-token or dynamic policies can provide a larger $\mathcal C_t$. We modify the standard hard assignment (Equation~\ref{eq:hard_assignment}) as:
\begin{equation}
\mathbf{e}_t^{(i)} =
\begin{cases}
\mathbf{e}_{\arg\max_{v} p_t^{(i)}(v)}, & \text{if } i \in \mathcal C_t \\
\tilde{\mathbf{e}}_t^{(i)}, & \text{otherwise}
\end{cases}
\label{eq:soft_hard_assignment}
\end{equation}
This preserves the distributional information from Phase 1's refinement in uncommitted positions, providing richer context for subsequent decoding steps while still allowing confident positions to make discrete commitments.

During decoding, we continue monitoring $\mathcal{D}_{\text{KL}}^{(t)}$ (Equation~\ref{eq:kl-gap}). If $\mathcal{D}_{\text{KL}}^{(t)} < \tau_{\text{decode}}$, \rev{we terminate generation and output the final sequence, using predictive stability as a practical efficiency criterion rather than a correctness guarantee}. In practice, this allows the model to adaptively adjust its \rev{number of decoding steps} based on the problem complexity rather than using a fixed number of steps.

\paragraph{Extending to multi-token commitment.}
Equation~\ref{eq:soft_hard_assignment} admits arbitrary commitment sets $\mathcal{C}_t$; in practice, committing all positions above a confidence threshold \rev{enables additional parallelism}. However, because self-attention couples all positions, a large batch of simultaneous soft-to-hard transitions could destabilise the remaining soft embeddings and cause confidence oscillations, \rev{in addition to the perturbations present under hard masking}. \rev{The analysis below bounds soft-update perturbations through $r_f$; composability with multi-token accelerators is evaluated empirically.}

\subsection{Threshold-Oriented Stability for Multi-Token Commitment}\label{sec:tcr}

Because LRD is training-free, the relevant theoretical question is inference-time stability rather than optimisation. Transformer encoders are not globally Lipschitz, but prior work shows that \emph{local} Lipschitz bounds can hold on small-norm balls~\citep{yudin2025payattentionattentiondistribution,hu2024specformerguardingvisiontransformer,dasoulas2021lipschitznormalizationselfattentionlayers}. The \texttt{[MASK]} embedding lies in the lower tail of the token-norm distribution, so choosing a small mixing ratio through $\alpha_t^{(i)}$ and $r_f$ in Eq.~\ref{eq:alpha} keeps mixed embeddings near this stable region (Figure~\ref{fig:base_rate} and Appendix~\ref{appendix:stability}). Unlike token-wise local bounds, self-attention couples every position to every other position; therefore, any useful Lipschitz statement for the per-position confidence $c_i^{(t)}=\max_v p_i^{(t)}(v)$ must depend on the entire sequence. Let $\Delta\tilde{\mathcal E}_t \!\coloneqq\! \big(\tilde{\mathbf e}_j^{(t+1)}-\tilde{\mathbf e}_j^{(t)}\big)_{j=1}^{L}$ and $\|\Delta\tilde{\mathcal E}_t\|_{2,\infty}\!\coloneqq\!\max_j \|\tilde{\mathbf e}_j^{(t+1)}-\tilde{\mathbf e}_j^{(t)}\|_2$. We use $\eta\in(0,1)$ for the failure probability (renamed from $\alpha$ to avoid clashing with the mixing coefficient $\alpha_t^{(i)}$ of Eq.~\ref{eq:alpha}) and $K$ for the Lipschitz constant (to avoid clashing with the sequence length $L$).

\begin{definition}[Local Lipschitz continuity, sequence-level]\label{def:local-lip-seq}
The confidence dynamics satisfy a \emph{$(K,\eta)$-local Lipschitz condition} over the LRD operating regime if, for each position $i$ and step $t$,
\begin{equation}
\Pr_{\mathcal{P}}\Big(\,\big|c_i^{(t+1)}-c_i^{(t)}\big|\le K\,\big\|\Delta\tilde{\mathcal E}_t\big\|_{2,\infty}\,\Big)\ge 1-\eta,
\end{equation}
where $\mathcal{P}$ is the joint distribution over (prompt, position, step) triples on a held-out validation set, and the soft embeddings remain inside a small-norm ball $\{\|\tilde{\mathbf e}\|_2\le R\}$.
\end{definition}

\rev{The following soft-update bounds hold with the committed context fixed. Hard commitments can introduce additional perturbations and are not bounded by $r_f$ alone; Appendix~\ref{appendix:tcr-proofs} states this scope explicitly.}

The next lemma shows that the very same $r_f$ that controls the soft mixture also caps the per-step embedding move, so the standard ``bounded update'' premise is not a free assumption but a property of LRD. Let $E_{\max}\!\coloneqq\!\max_v\|\mathbf{e}_v\|_2$ and $E_{\texttt{[MASK]}}\!\coloneqq\!\|\mathbf{e}_{\texttt{[MASK]}}\|_2$.

\begin{lemma}[Per-step embedding bound]\label{lem:rho}
For every $i$ and $t$,
\(
\big\|\tilde{\mathbf e}_i^{(t+1)}-\tilde{\mathbf e}_i^{(t)}\big\|_2\le 2\,r_f\,(E_{\max}+E_{\texttt{[MASK]}})\eqqcolon\rho.
\)
\end{lemma}

The proof (Appendix~\ref{appendix:tcr-proofs}) uses \rev{the triangle inequality and the distance of each soft embedding from the mask anchor}. Note $\rho\!\to\!0$ as $r_f\!\to\!0$.

\begin{definition}[Threshold-Oriented Convergence Rate (TCR)]\label{def:tcr}
With commitment threshold $\tau$, distance $d_i^{(t)}=[\tau-c_i^{(t)}]_+$, and $M_t$ the set of uncommitted positions,
\(
\mathrm{TCR}^{(t)}=\tfrac{1}{|M_t|}\sum_{i\in M_t}\big(d_i^{(t)}-d_i^{(t+1)}\big).
\)
\end{definition}

\begin{theorem}[High-probability bound on backward updates]\label{thm:back}
Under Definition~\ref{def:local-lip-seq} and Lemma~\ref{lem:rho}, for any $\delta>K\rho$ and every $i\in M_t$,
\begin{equation}
\Pr_{\mathcal{P}}\big(\,d_i^{(t+1)}-d_i^{(t)}>\delta\,\big)\le\eta.
\end{equation}
Hence the expected number of large backward updates over $M_t$ is at most $\eta\,|M_t|$.
\end{theorem}

The proof plugs $\rho$ into the Lipschitz bound on the high-probability event and uses that $[\tau-\cdot]_+$ is $1$-Lipschitz. Theorem~\ref{thm:back} controls oscillation; to translate it into expected forward progress we add a measurable confidence-gap condition, replacing the assumption-laden positivity of expected progress used in earlier drafts.

\begin{corollary}[Lower bound on expected TCR]\label{cor:tcr}
Let $\gamma\in(0,\tau]$ and define the \emph{progress event}
\(
G_i^{(t)}\!\coloneqq\!\big\{c_i^{(t+1)}-c_i^{(t)}\ge\gamma\big\}\cap\big\{c_i^{(t)}\le\tau-\gamma\big\}.
\)
If $\Pr_{\mathcal{P}}\!\big(G_i^{(t)}\big)\ge q_t$ for some $q_t\in(\eta,1]$, then under the assumptions of Theorem~\ref{thm:back},
\begin{equation}
\mathbb{E}\big[\mathrm{TCR}^{(t)}\big]\;\ge\;\gamma\,(q_t-\eta)\;-\;K\rho\,(1-q_t)\;-\;\tau\,\eta.
\end{equation}
\end{corollary}

The side condition $c_i^{(t)}\le\tau-\gamma$ ensures the position has at least $\gamma$ ``room'' to make progress. In the small-$r_f$ regime $K\rho\ll\tau$ and the bound simplifies to $\gamma(q_t-\eta)-\tau\eta\;-\;O(K\rho)$, which is strictly positive \rev{when $q_t>[K\rho+\eta(\gamma+\tau)]/(\gamma+K\rho)$}. The quantities $\gamma$, $q_t$, and $\eta$ admit a measurement protocol on a held-out validation set under any chosen $r_f$, so Corollary~\ref{cor:tcr} can be checked rather than postulated; Appendix~\ref{appendix:tcr-proofs} provides \rev{the measurement protocol and empirical threshold-progress curves}, with Figure~\ref{fig:tcr-verification} showing that LRD moves positions toward the commitment threshold faster and yields larger positive TCR than the baseline.

\rev{These conditional bounds motivate controlling soft-feedback strength. Table~\ref{tab:main_result} separately demonstrates empirical composability with confidence-threshold policies; the bounds do not certify every simultaneous hard-commitment step.}

\begin{table*}[!htbp]
\caption{Main comparison at sequence lengths 256 and 512. Speed is normalized to vanilla decoding (1.0$\times$; higher is faster). Entropy-min is the standard hard-assignment decoder that commits the minimum-entropy token each step. Green deltas compare LRD with Entropy-min and Fast-dLLM+LRD with Conf-thresh (confidence threshold $0.9$). Additional results are in Appendix~\ref{appendix:results}.}
\centering
\renewcommand{\arraystretch}{1}
\resizebox{0.9 \linewidth}{!}{
\begin{tabular}{lcclllllllll}
\toprule[1pt]
\multirow{2}{*}{\textbf{Model}} & \multirow{2}{*}{\textbf{Len}} & \multirow{2}{*}{\textbf{Commit}} & \multirow{2}{*}{\textbf{Method}} & \multicolumn{2}{c}{\textbf{HumanEval}} & \multicolumn{2}{c}{\textbf{MBPP}} & \multicolumn{2}{c}{\textbf{GSM8K}} & \multicolumn{2}{c}{\textbf{MATH500}} \\
\cmidrule(lr){5-6} \cmidrule(lr){7-8} \cmidrule(lr){9-10} \cmidrule(lr){11-12}
& & & & Acc & Speed & Acc & Speed & Acc & Speed & Acc & Speed \\
\midrule
\multirow{16}{*}{\makecell[l]{\textbf{Dream}\\\textbf{Base-7B}}}
& \multirow{8}{*}{256} & \multirow{3}{*}{Single} & Entropy-min & \textcolor{DarkGrey}{50.6} & \textcolor{DarkGrey}{1.0$\times$} & \textcolor{DarkGrey}{55.8} & \textcolor{DarkGrey}{1.0$\times$} & \textcolor{DarkGrey}{75.3} & \textcolor{DarkGrey}{1.0$\times$} & \textcolor{DarkGrey}{36.9} & \textcolor{DarkGrey}{1.0$\times$} \\
& & & Prophet & 51.1 & 1.2$\times$ & 55.1 & 2.1$\times$ & 75.6 & 1.5$\times$ & 36.5 & 1.4$\times$ \\
& & & LRD & \textbf{56.9}$_{\textcolor{LightGreen}{+6.3}}$ & \textbf{1.2$\times$} & \textbf{57.6}$_{\textcolor{LightGreen}{+1.8}}$ & \textbf{2.3$\times$} & \textbf{78.2}$_{\textcolor{LightGreen}{+2.9}}$ & \textbf{1.8$\times$} & \textbf{39.8}$_{\textcolor{LightGreen}{+2.9}}$ & \textbf{1.4$\times$} \\
\cmidrule(lr){3-12}
& & \multirow{5}{*}{Multi} & Conf-thresh & \textcolor{DarkGrey}{48.9} & \textcolor{DarkGrey}{1.4$\times$} & \textcolor{DarkGrey}{55.5} & \textcolor{DarkGrey}{1.9$\times$} & \textcolor{DarkGrey}{74.7} & \textcolor{DarkGrey}{2.5$\times$} & \textcolor{DarkGrey}{36.7} & \textcolor{DarkGrey}{2.4$\times$} \\
& & & KLASS & 51.4 & 1.5$\times$ & 56.6 & 1.4$\times$ & 75.2 & 1.8$\times$ & 37.2 & 1.2$\times$ \\
& & & SlowFast & 47.4 & 1.6$\times$ & 53.8 & 2.0$\times$ & 74.7 & 1.7$\times$ & 36.3 & 1.9$\times$ \\
& & & Fast-dLLM & 49.6 & 2.4$\times$ & 54.6 & 4.2$\times$ & 73.5 & 4.8$\times$ & 36.5 & 3.8$\times$ \\
& & & Fast-dLLM+LRD & \textbf{56.1}$_{\textcolor{LightGreen}{+7.2}}$ & \textbf{2.8$\times$} & \textbf{57.4}$_{\textcolor{LightGreen}{+1.9}}$ & \textbf{6.8$\times$} & \textbf{77.4}$_{\textcolor{LightGreen}{+2.7}}$ & \textbf{6.4$\times$} & \textbf{39.4}$_{\textcolor{LightGreen}{+2.7}}$ & \textbf{5.2$\times$} \\
\cmidrule(lr){2-12}
& \multirow{8}{*}{512} & \multirow{3}{*}{Single} & Entropy-min & \textcolor{DarkGrey}{54.4} & \textcolor{DarkGrey}{1.0$\times$} & \textcolor{DarkGrey}{55.8} & \textcolor{DarkGrey}{1.0$\times$} & \textcolor{DarkGrey}{76.2} & \textcolor{DarkGrey}{1.0$\times$} & \textcolor{DarkGrey}{37.5} & \textcolor{DarkGrey}{1.0$\times$} \\
& & & Prophet & 55.1 & 2.3$\times$ & 55.3 & 4.2$\times$ & 76.6 & 3.2$\times$ & 37.2 & 1.7$\times$ \\
& & & LRD & \textbf{58.8}$_{\textcolor{LightGreen}{+4.4}}$ & \textbf{2.6$\times$} & \textbf{58.4}$_{\textcolor{LightGreen}{+2.6}}$ & \textbf{4.5$\times$} & \textbf{77.4}$_{\textcolor{LightGreen}{+1.2}}$ & \textbf{3.4$\times$} & \textbf{40.8}$_{\textcolor{LightGreen}{+3.3}}$ & \textbf{1.8$\times$} \\
\cmidrule(lr){3-12}
& & \multirow{5}{*}{Multi} & Conf-thresh & \textcolor{DarkGrey}{52.6} & \textcolor{DarkGrey}{2.7$\times$} & \textcolor{DarkGrey}{54.5} & \textcolor{DarkGrey}{3.6$\times$} & \textcolor{DarkGrey}{76.2} & \textcolor{DarkGrey}{4.5$\times$} & \textcolor{DarkGrey}{37.7} & \textcolor{DarkGrey}{2.4$\times$} \\
& & & KLASS & 53.2 & 2.1$\times$ & 56.6 & 1.8$\times$ & 76.1 & 2.1$\times$ & 40.4 & 1.5$\times$ \\
& & & SlowFast & 52.2 & 2.2$\times$ & 53.8 & 2.8$\times$ & 75.6 & 3.4$\times$ & 36.9 & 2.6$\times$ \\
& & & Fast-dLLM & 52.8 & 3.0$\times$ & 55.1 & 5.4$\times$ & 74.9 & 6.0$\times$ & 37.1 & 4.5$\times$ \\
& & & Fast-dLLM+LRD & \textbf{56.3}$_{\textcolor{LightGreen}{+3.7}}$ & \textbf{6.7$\times$} & \textbf{57.1}$_{\textcolor{LightGreen}{+2.6}}$ & \textbf{7.6$\times$} & \textbf{77.3}$_{\textcolor{LightGreen}{+1.1}}$ & \textbf{15.2$\times$} & \textbf{40.2}$_{\textcolor{LightGreen}{+2.5}}$ & \textbf{6.8$\times$} \\
\midrule
\multirow{16}{*}{\makecell[l]{\textbf{LLaDA}\\\textbf{Ins-8B}}}
& \multirow{8}{*}{256} & \multirow{3}{*}{Single} & Entropy-min & \textcolor{DarkGrey}{38.7} & \textcolor{DarkGrey}{1.0$\times$} & \textcolor{DarkGrey}{36.9} & \textcolor{DarkGrey}{1.0$\times$} & \textcolor{DarkGrey}{77.4} & \textcolor{DarkGrey}{1.0$\times$} & \textcolor{DarkGrey}{33.8} & \textcolor{DarkGrey}{1.0$\times$} \\
& & & Prophet & 39.3 & 1.1$\times$ & 36.1 & 1.2$\times$ & 77.6 & 1.4$\times$ & 33.3 & 1.3$\times$ \\
& & & LRD & \textbf{43.3}$_{\textcolor{LightGreen}{+4.6}}$ & \textbf{1.2$\times$} & \textbf{40.0}$_{\textcolor{LightGreen}{+3.1}}$ & \textbf{1.3$\times$} & \textbf{78.8}$_{\textcolor{LightGreen}{+1.4}}$ & \textbf{1.5$\times$} & \textbf{35.8}$_{\textcolor{LightGreen}{+2.0}}$ & \textbf{1.4$\times$} \\
\cmidrule(lr){3-12}
& & \multirow{5}{*}{Multi} & Conf-thresh & \textcolor{DarkGrey}{36.0} & \textcolor{DarkGrey}{2.3$\times$} & \textcolor{DarkGrey}{36.6} & \textcolor{DarkGrey}{2.6$\times$} & \textcolor{DarkGrey}{77.0} & \textcolor{DarkGrey}{3.1$\times$} & \textcolor{DarkGrey}{33.0} & \textcolor{DarkGrey}{2.4$\times$} \\
& & & KLASS & 38.9 & 1.6$\times$ & 38.1 & 1.5$\times$ & 78.8 & 1.8$\times$ & 34.2 & 1.5$\times$ \\
& & & SlowFast & 36.9 & 2.1$\times$ & 35.1 & 1.8$\times$ & 77.2 & 2.2$\times$ & 33.8 & 1.7$\times$ \\
& & & Fast-dLLM & 38.1 & 2.6$\times$ & 34.6 & 3.6$\times$ & 75.1 & 4.5$\times$ & 31.6 & 3.1$\times$ \\
& & & Fast-dLLM+LRD & \textbf{42.6}$_{\textcolor{LightGreen}{+6.6}}$ & \textbf{3.1$\times$} & \textbf{39.2}$_{\textcolor{LightGreen}{+2.6}}$ & \textbf{4.5$\times$} & \textbf{78.5}$_{\textcolor{LightGreen}{+1.5}}$ & \textbf{5.7$\times$} & \textbf{35.3}$_{\textcolor{LightGreen}{+2.3}}$ & \textbf{4.2$\times$} \\
\cmidrule(lr){2-12}
& \multirow{8}{*}{512} & \multirow{3}{*}{Single} & Entropy-min & \textcolor{DarkGrey}{43.9} & \textcolor{DarkGrey}{1.0$\times$} & \textcolor{DarkGrey}{38.2} & \textcolor{DarkGrey}{1.0$\times$} & \textcolor{DarkGrey}{81.3} & \textcolor{DarkGrey}{1.0$\times$} & \textcolor{DarkGrey}{37.7} & \textcolor{DarkGrey}{1.0$\times$} \\
& & & Prophet & 44.7 & 1.3$\times$ & 37.6 & 1.4$\times$ & 81.8 & 1.7$\times$ & 37.5 & 1.5$\times$ \\
& & & LRD & \textbf{48.4}$_{\textcolor{LightGreen}{+4.5}}$ & \textbf{1.3$\times$} & \textbf{40.6}$_{\textcolor{LightGreen}{+2.4}}$ & \textbf{1.5$\times$} & \textbf{84.5}$_{\textcolor{LightGreen}{+3.2}}$ & \textbf{2.0$\times$} & \textbf{39.8}$_{\textcolor{LightGreen}{+2.1}}$ & \textbf{1.6$\times$} \\
\cmidrule(lr){3-12}
& & \multirow{5}{*}{Multi} & Conf-thresh & \textcolor{DarkGrey}{41.8} & \textcolor{DarkGrey}{2.1$\times$} & \textcolor{DarkGrey}{37.0} & \textcolor{DarkGrey}{3.1$\times$} & \textcolor{DarkGrey}{80.1} & \textcolor{DarkGrey}{6.5$\times$} & \textcolor{DarkGrey}{37.4} & \textcolor{DarkGrey}{2.1$\times$} \\
& & & KLASS & 44.1 & 2.1$\times$ & 37.9 & 2.0$\times$ & 81.7 & 4.6$\times$ & 38.2 & 1.9$\times$ \\
& & & SlowFast & 42.1 & 2.8$\times$ & 36.4 & 2.5$\times$ & 81.1 & 5.1$\times$ & 37.8 & 2.4$\times$ \\
& & & Fast-dLLM & 41.6 & 3.2$\times$ & 36.2 & 4.2$\times$ & 78.6 & 11.3$\times$ & 35.1 & 3.8$\times$ \\
& & & Fast-dLLM+LRD & \textbf{45.7}$_{\textcolor{LightGreen}{+3.9}}$ & \textbf{4.7$\times$} & \textbf{38.8}$_{\textcolor{LightGreen}{+1.8}}$ & \textbf{5.8$\times$} & \textbf{83.7}$_{\textcolor{LightGreen}{+3.6}}$ & \textbf{18.4$\times$} & \textbf{39.2}$_{\textcolor{LightGreen}{+1.8}}$ & \textbf{5.3$\times$} \\
\bottomrule[1pt]
\end{tabular}
}
\label{tab:main_result}
\end{table*}

\section{Experiments}\label{sec:experiments}

\subsection{Implementation Details}
We evaluate our method on representative diffusion-based language models, with the main comparison focused on Dream-Base-7B and LLaDA-Ins-8B. Additional LRD-only results cover Dream-Instruct-7B, LLaDA-Base-8B, LLaDA-1.5-8B, and LLaDA2.0-mini-preview 16B~\citep{nie2025largelanguagediffusionmodels,zhu2025llada15variancereducedpreference,ye2025dream}. To ensure robustness, we fix the temperature to 0 and always select the token with the minimum entropy at each decoding step; detailed evaluation settings, including the integration with the semi-AR framework used by LLaDA-family models, are provided in Appendix~\ref{appendix:Experiment-settings}. All experiments are conducted on a server equipped with 8 NVIDIA A100 80GB GPUs.

\subsection{Benchmarks and Metrics}
To evaluate our approach, we conduct main experiments on four benchmarks covering mathematical reasoning and code generation. For mathematical reasoning, we use GSM8K~\citep{cobbe2021training} and the more challenging MATH500~\citep{lightman2024lets}. For code generation, we evaluate on MBPP~\citep{austin2021programsynthesislargelanguage} and HumanEval~\citep{chen2021evaluatinglargelanguagemodels}. 
Following prior work, all Instruct models are evaluated zero-shot. For Base models, we use standard few-shot settings: zero-shot for HumanEval, 3-shot for MBPP, 4-shot for MATH500, and 8-shot for GSM8K. We report accuracy on math reasoning and pass@1 on code generation. Appendix~\ref{appendix:general_qa} further evaluates general language quality on RACE~\citep{lai-etal-2017-race}, CNN/DailyMail~\citep{see-etal-2017-get}, NQ-Open~\citep{kwiatkowski-etal-2019-natural}, and TruthfulQA~\citep{lin-etal-2022-truthfulqa}.
%using accuracy for RACE, BERTScore and ROUGE-1/L for CNN/DailyMail, exact match for NQ-Open, and the standard TruthfulQA accuracy variants.

\paragraph{Baselines.}
We compare against baselines that represent different decoding choices. Entropy-min is the standard hard-assignment decoder that commits the lowest-entropy token at each step. Prophet~\citep{li2025diffusion} is a single-token acceleration method that uses early prediction signals before final decoding. Conf-thresh commits all positions whose confidence exceeds $0.9$, giving a simple parallel-commitment baseline. KLASS~\citep{kim2025klass} uses KL-guided criteria for faster masked-diffusion inference, while SlowFast Sampling~\citep{wei2025slowfast} alternates slow and fast denoising schedules. Fast-dLLM~\citep{wu2025fastdllmtrainingfreeaccelerationdiffusion} combines KV caching with confidence-threshold parallel commitment at $0.9$. DAEDAL~\citep{li2025beyond} is a dynamic acceleration baseline that adapts denoising length using EOS confidence and low-confidence region detection.

\subsection{Main Results}
\textbf{Performance on Benchmarks.} Table~\ref{tab:main_result} reports the main comparison across four representative benchmarks at sequence lengths 256 and 512. We focus on Dream-Base-7B and LLaDA-Ins-8B and include both LRD-only decoding and Fast-dLLM with dual cache and parallel decoding. LRD improves the baseline on all four benchmarks at both lengths; when integrated with Fast-dLLM, it improves accuracy and adds further speedups over Fast-dLLM alone.
Appendix~\ref{appendix:results} follows the same result story, covering single-token LRD-only decoding, multi-token commitment strategies, dynamic acceleration with DAEDAL, and additional model/dataset results.
\rev{Appendix~\ref{appendix:recent_comparisons} additionally reports training-based comparisons, further model families, and GPQA/BBH. Accelerator combinations test composability.}

\begin{revision}
\textbf{Recent training-free decoders.}
Table~\ref{tab:main_soar_credit} compares SOAR~\citep{cao2026soar} and CreditDecoding~\citep{wang2025creditdecoding} under matched evaluation and hardware on LLaDA-Instruct-8B ($L=512$). LRD is more accurate on all four tasks; SOAR is faster on both coding tasks. Appendix~\ref{appendix:soar_credit} gives results at $L=256$.

\begin{table}[!htbp]
\ifshowchanges\color{blue}\captionsetup{font+={color=blue},labelfont+={color=blue}}\fi
\centering
\vspace{-6pt}
\caption{LLaDA-Instruct-8B ($L=512$). Speed is relative to native decoding.}
\label{tab:main_soar_credit}
\small
\resizebox{0.8\textwidth}{!}{%
\begin{tabular}{lcccccccc}
\toprule
\multirow{2}{*}{\textbf{Method}} & \multicolumn{2}{c}{\textbf{HumanEval}} & \multicolumn{2}{c}{\textbf{MBPP}} & \multicolumn{2}{c}{\textbf{GSM8K}} & \multicolumn{2}{c}{\textbf{MATH500}} \\
\cmidrule(lr){2-3} \cmidrule(lr){4-5} \cmidrule(lr){6-7} \cmidrule(lr){8-9}
& Acc & Speed & Acc & Speed & Acc & Speed & Acc & Speed \\
\midrule
Baseline & \textcolor{DarkGrey}{43.9} & \textcolor{DarkGrey}{1.0$\times$} & \textcolor{DarkGrey}{38.2} & \textcolor{DarkGrey}{1.0$\times$} & \textcolor{DarkGrey}{81.3} & \textcolor{DarkGrey}{1.0$\times$} & \textcolor{DarkGrey}{37.7} & \textcolor{DarkGrey}{1.0$\times$} \\
SOAR & 46.6 & \textbf{1.7$\times$} & 39.2 & \textbf{1.6$\times$} & 82.1 & 1.4$\times$ & 39.0 & 1.2$\times$ \\
CreditDecoding & 45.6 & 1.3$\times$ & 38.9 & 1.5$\times$ & 80.7 & 1.9$\times$ & 37.6 & 1.5$\times$ \\
LRD & \textbf{48.4} & 1.3$\times$ & \textbf{40.6} & 1.5$\times$ & \textbf{84.5} & \textbf{2.0$\times$} & \textbf{39.8} & \textbf{1.6$\times$} \\
\bottomrule
\end{tabular}}
\vspace{-6pt}
\end{table}
\end{revision}

\begin{figure}[h!]
\centering
\begin{minipage}{0.49\textwidth}
    \centering
    \includegraphics[width=\linewidth]{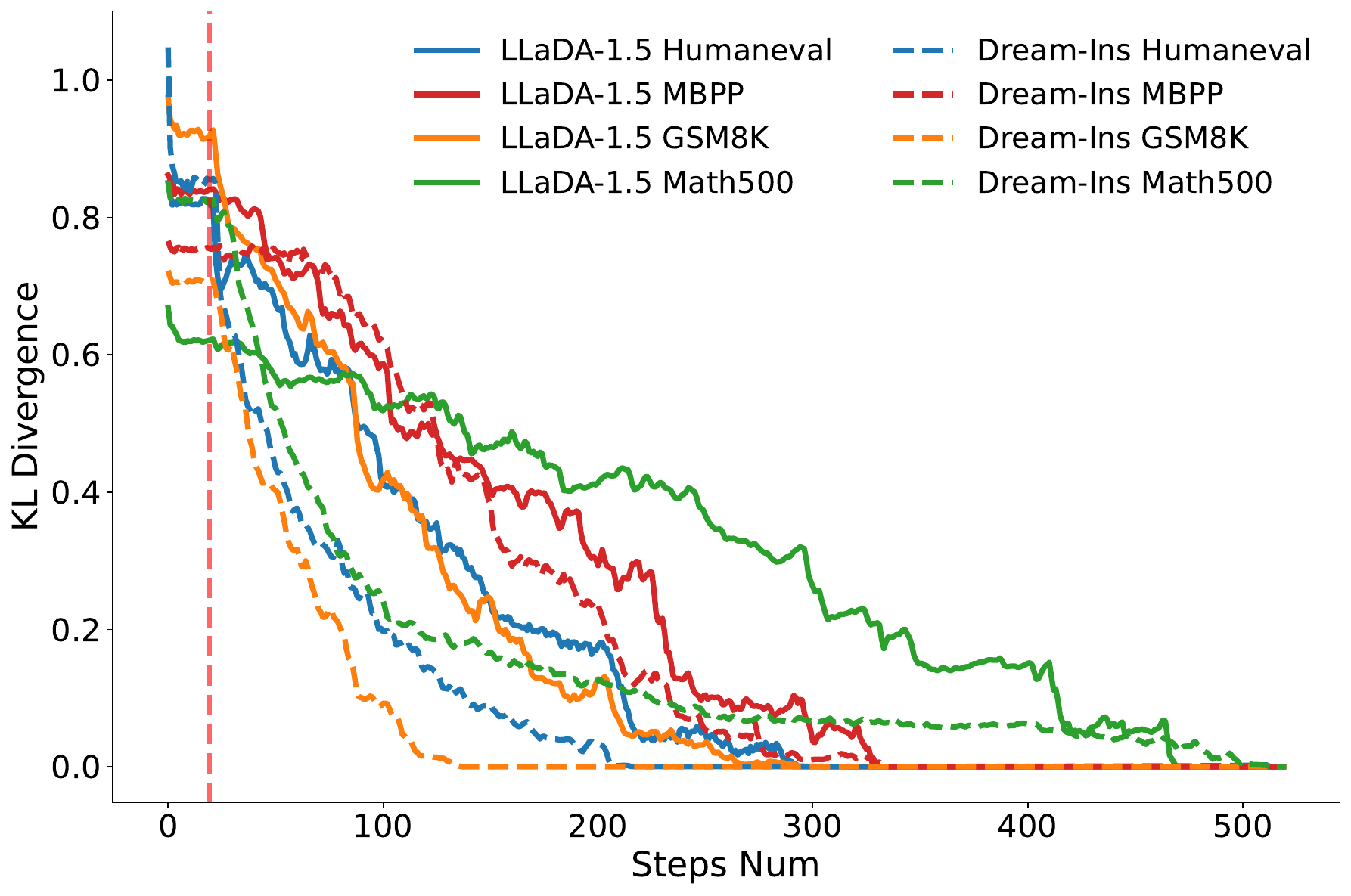}
    \caption{KL divergence between step-wise predictive distributions and final decoded results. The red vertical line marks where decoding begins after a fixed 20-step latent refinement.}
    \label{fig:kl}
\end{minipage}
\hfill
\begin{minipage}{0.49\textwidth}
    \centering
    \includegraphics[width=\linewidth]{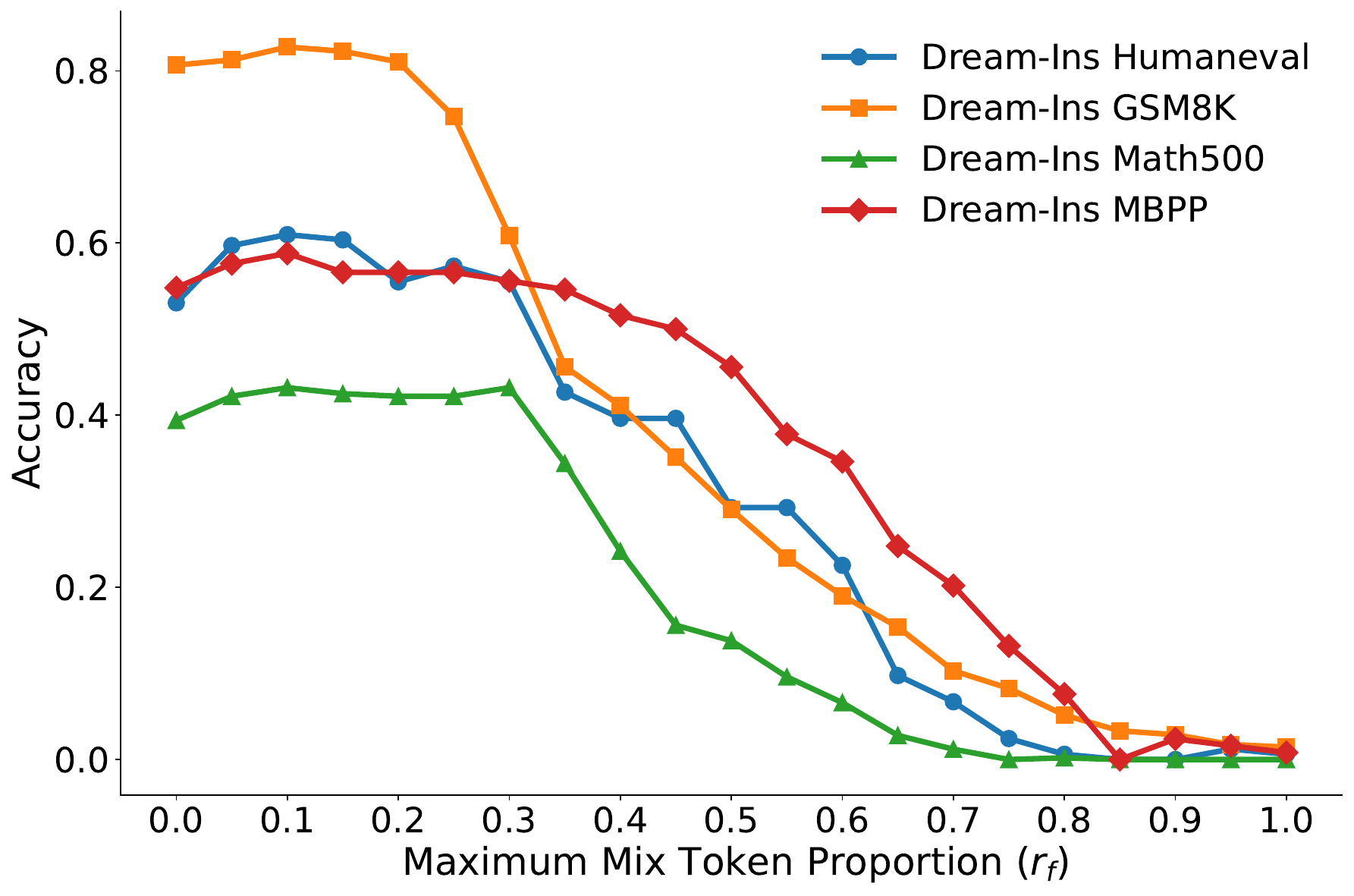}
    \caption{Accuracy of Dream-Ins on four benchmarks under different maximum token proportions, where $r_f=0$ corresponds to no mixing.}
    \label{fig:base_rate}
\end{minipage}

\end{figure}

% % --- Figure 2: Hotstart / Convergence ratio ---
% \begin{figure}[t]
% \centering
% \resizebox{0.85 \linewidth}{!}{%
%     \includegraphics{images/hotstart.pdf}%
% }
% \caption{Convergence ratios across latent refinement steps for LLaDA-1.5 and Dream-Ins on four benchmarks. Since computing the difference in KL divergence requires at least three consecutive steps, the curves are plotted starting from step 2.}
% \label{fig:hotstart}
% \vspace{-3mm}
% \end{figure}

% % --- Figure 1: KL ---
% \begin{figure}[t]
% \centering
% \resizebox{0.85 \linewidth}{!}{%
%     \includegraphics{images/kl.pdf}%
% }
% \caption{KL divergence between step-wise predictive distributions and final decoded results for LLaDA-1.5 and Dream-Ins across benchmarks. The red vertical line marks where decoding begins after a fixed 20-step latent refinement.}
% \label{fig:kl}
% \vspace{-3mm}
% \end{figure}

\textbf{Efficiency and Decoding Speed.}
Beyond accuracy, LRD substantially accelerates inference. As shown in Table~\ref{tab:main_result}, LRD improves speed over the baseline on all four benchmarks, and the Fast-dLLM+LRD combination gives the strongest speedups where matched measurements are available. For example, Fast-dLLM(dual cache+parallel)+LRD reaches $6.7\times$ speedup on HumanEval and $15.2\times$ on GSM8K for Dream-Base-7B, and $4.7\times$ and $18.4\times$ respectively for LLaDA-Ins-8B. 
\rev{These end-to-end ratios include sequence-level early stopping. On Dream-Instruct-7B, HumanEval at $L=512$, LRD increases per-step latency from 81.8 to 94.3\,ms, but reduces average steps from 512 to 153 and total runtime from 41.8 to 14.4\,s (Appendix~\ref{appendix:cost_pareto}). At $L=1024$, content-length-normalised speedups with only Phase~2 early stopping disabled are 1.06--1.09$\times$ for LRD and 1.11--1.21$\times$ for Fast-dLLM+LRD (Appendix~\ref{appendix:content_speed}). Large-window headline gains therefore arise mainly from sequence-level early stopping and should not be interpreted as proportional increases in reasoning speed.}
% The gain comes from two complementary effects: (i) predictive mixing improves the converged belief state and decoding stability (Section~\ref{sec:ab_study}; Appendix~\ref{appendix:unified_ablation_results}), and (ii) KL-divergence-based early stopping prevents unnecessary refinement and decoding steps, especially in long sequences (Appendix~\ref{appendix:cost_pareto}).

\subsection{Convergence Analysis}

\textbf{KL divergence decreases steadily during refinement and decoding.}  
Figure~\ref{fig:kl} shows the KL divergence between step-wise predictive distributions and the final decoded outputs for LLaDA-1.5 and Dream-Ins across four benchmarks. For ease of observation, we fix the latent refinement phase to 20 steps. The divergence exhibits a clear downward trend: during the latent refinement phase, the KL values drop rapidly and stabilise, indicating that the latent belief state quickly converges before decoding begins. Once decoding starts, the KL divergence continues to decrease with mild fluctuations, reflecting the model’s progressive confidence sharpening. 
% For most benchmarks, the divergence approaches zero within about 300 steps, whereas Dream-Ins converges even faster, reaching near-zero divergence around 140 steps. The MATH500 benchmark proves more challenging, with non-negligible divergence persisting until the full 512-step horizon. 
Overall, these patterns are consistent with our expectations: the refinement phase provides a stable initialisation, and the subsequent decoding stage steadily drives the system toward convergence. Appendix~\ref{appendix:convergence_analysis} further groups the convergence-ratio curves and metric comparison.

\rev{Under adaptive refinement, 68.9\% of Dream-Instruct-7B HumanEval examples converge by step 2 and more than 85\% by step 3 (Appendix~\ref{appendix:latent_refinement_ratio}). Thus most examples transition well before the 20-step cap, limiting initial refinement overhead. The weak sensitivity to the refinement cap in Appendix~\ref{appendix:hyperparam} is consistent with this early convergence.}

% % --- Figure 2: Hotstart / Convergence ratio ---
% \begin{figure}[t]
% \centering
% \resizebox{0.85 \linewidth}{!}{%
%     \includegraphics{images/hotstart.pdf}%
% }
% \caption{Convergence ratios across latent refinement steps for LLaDA-1.5 and Dream-Ins on four benchmarks. Since computing the difference in KL divergence requires at least three consecutive steps, the curves are plotted starting from step 2.}
% \label{fig:hotstart}
% \vspace{-3mm}
% \end{figure}

\subsection{Ablation Study}\label{sec:ab_study}

\begin{wraptable}{r}{0.5\textwidth}
\vspace{-1.2em}
\centering
\scriptsize
\captionsetup{width=\linewidth}
\caption{Random Gaussian noise (GN) vs.\ soft embeddings on Dream-Instruct-7B ($L=512$).}
\label{tab:noise_ablation}
\setlength{\tabcolsep}{2.4pt}
\renewcommand{\arraystretch}{1}

\begin{tabular*}{\linewidth}{@{\extracolsep{\fill}}lcccc@{}}
\toprule
Method & HE & MBPP & GSM8K & MATH \\
\midrule
Baseline                  & 56.1 & 56.7 & 80.2 & 38.6 \\
GN, $r_f=0.1$             & 49.2 & 52.2 & 70.4 & 33.4 \\
GN, $r_f=0.2$             & 47.0 & 51.4 & 69.8 & 32.4 \\
KL-stop only              & 55.8 & 56.9 & 80.4 & 38.3 \\
LRD w/o ent.\ wt.         & 57.7 & 57.9 & 76.8 & 40.2 \\
LRD (full)                & \textbf{60.9} & \textbf{58.8} & \textbf{82.7} & \textbf{41.8} \\
\bottomrule
\end{tabular*}
\end{wraptable}

\textbf{LRD's accuracy gains come from predictive soft mixtures.}
Table~\ref{tab:noise_ablation} provides a direct control experiment.
Replacing the predictive soft mixture with Gaussian noise of comparable magnitude catastrophically degrades accuracy, while KL-stop only remains close to the baseline. The comparison also separates the mixture itself from its entropy weighting, showing that the gains depend on token-level belief information rather than random nudging, stopping criteria, or a single tuning detail. Appendix~\ref{appendix:unified_ablation_results} reports component ablations with effective token counts ($E_{\text{token}}$), Appendix~\ref{appendix:repeated_refinement} studies repeated refinement, and Appendix~\ref{appendix:convergence_analysis}--\ref{appendix:hyperparameter_analysis} detail convergence ratios, convergence-metric comparisons, top-$p$ mixing, and sweeps over $\tau_{\text{refine}}$, $\tau_{\text{decode}}$, and $T_{\text{refine}}$.
\rev{At matched pre-commit states, soft feedback increases correct-token probability by 0.12--0.13 on probed error positions from examples LRD ultimately corrects, and by 0.04--0.06 when both final responses remain wrong (Appendix~\ref{appendix:matched_state}). Small-budget and cross-window robustness results are in Appendices~\ref{appendix:small_budgets} and~\ref{appendix:cross_window}.}

\begin{revision}
\textbf{Component contributions.}
Table~\ref{tab:main_component_ablation} shows that removing initial refinement or mixed embeddings reduces accuracy on all four tasks. Disabling early stopping largely preserves accuracy but reduces speedups to 1.0--1.2$\times$, separating the sources of quality improvement and acceleration.

\begin{table}[!htbp]
\color{black}\captionsetup{font+={color=black},labelfont+={color=black}}
\centering
\vspace{-6pt}
\caption{Dream-Instruct-7B ablations ($L=512$). Speed is relative to the baseline.}
\label{tab:main_component_ablation}
\small
\resizebox{0.8\textwidth}{!}{%
\begin{tabular}{lcccccccc}
\toprule
\multirow{2}{*}{\textbf{Method}} & \multicolumn{2}{c}{\textbf{HumanEval}} & \multicolumn{2}{c}{\textbf{MBPP}} & \multicolumn{2}{c}{\textbf{GSM8K}} & \multicolumn{2}{c}{\textbf{MATH500}} \\
\cmidrule(lr){2-3} \cmidrule(lr){4-5} \cmidrule(lr){6-7} \cmidrule(lr){8-9}
& Acc & Speed & Acc & Speed & Acc & Speed & Acc & Speed \\
\midrule
Baseline & 56.1 & 1.0$\times$ & 56.7 & 1.0$\times$ & 80.2 & 1.0$\times$ & 38.6 & 1.0$\times$ \\
LRD (full) & 60.9 & 2.9$\times$ & 58.8 & 4.6$\times$ & 82.7 & 3.6$\times$ & 41.8 & 1.2$\times$ \\
w/o latent refinement & 59.9 & 3.1$\times$ & 57.8 & 4.9$\times$ & 82.2 & 3.8$\times$ & 41.0 & 1.2$\times$ \\
w/o mixed embeddings & 58.0 & 2.7$\times$ & 57.8 & 4.3$\times$ & 80.7 & 3.0$\times$ & 40.8 & 1.0$\times$ \\
w/o early stopping & 61.2 & 1.2$\times$ & 58.8 & 1.1$\times$ & 82.9 & 1.1$\times$ & 41.9 & 1.0$\times$ \\
\bottomrule
\end{tabular}}
\vspace{-6pt}
\end{table}
\end{revision}

% --- Figure 1: base_rate ---
% \begin{figure}[t]
% \centering
% \resizebox{0.88 \linewidth}{!}{%
%     \includegraphics{images/base_rate.pdf}%
% }
% \caption{Accuracy of Dream-Instruct model  on four benchmarks under different Maximum mix token , where $r_f$=0 corresponds to no mixing.}
% \label{fig:base_rate}
% \vspace{-3mm}
% \end{figure}
% --- Figure 2: topp ---
% \begin{figure}[t]
% \centering
% \resizebox{0.95 \linewidth}{!}{%
%     \includegraphics{images/topp.pdf}%
% }
% \caption{Effect of top-$p$ mixing on Dream-Ins across four benchmarks. The purple curve shows the log fraction of tokens included in the mixture.}
% \label{fig:topp}
% \vspace{-2mm}
% \end{figure}

\textbf{Full mixing collapses the model, while intermediate $r_f$ works best.}  
We further investigate the effect of the maximum mix ratio $r_f$, which scales the interpolation between predicted token embeddings and the \texttt{[MASK]} embedding during refinement (Eq.~\ref{eq:alpha}). When $r_f$=0, the model falls back to always using the \texttt{[MASK]} token for unfinalised positions, equivalent to the baseline. At the other extreme, setting $r_f$=1 allows the mixing weight to fully follow the entropy schedule, meaning that in low-entropy cases the \texttt{[MASK]} embedding may vanish. As shown in Figure~\ref{fig:base_rate}, both extremes are suboptimal: the baseline propagates information slowly, while overly aggressive mixing destabilises refinement and leads to collapse. Intermediate values of $r_f$ achieve the best trade-off, providing sufficient mask guidance while still leveraging predictive feedback. This matches the stability analysis: Lemma~\ref{lem:rho} shows that the per-step embedding movement grows linearly with $r_f$, while Appendix~\ref{appendix:stability} using Local Lipschitz Continuity explains why bounded mixing keeps soft embeddings near the stable \texttt{[MASK]} anchor.
The corresponding top-$p$ sensitivity analysis is deferred to the hyperparameter analysis in Appendix~\ref{appendix:hyperparameter_analysis}.

\FloatBarrier

\section{Related Work}
\textbf{Diffusion LLMs (dLLMs).}
Diffusion models first achieved strong results in continuous domains such as image~\citep{song2020score, ho2020denoising, nichol2021glide, rombach2022high} and speech generation~\citep{huang2023make, yang2023diffsound}, but applying them to language is challenging because text is discrete. Masked Diffusion Models (MDMs)~\citep{austin2021structured, ou2024your, shi2024simplified, lou2024discrete} address this by iteratively predicting masked tokens from bidirectional context, and recent work has scaled this paradigm into large language models. DiffuLLaMA~\citep{DBLP:conf/iclr/GongA0YZLAZB00K25} shows that diffusion LLMs can approach autoregressive models through continual pre-training, while commercial and open-source systems such as Mercury~\citep{labs2025mercuryultrafastlanguagemodels}, Gemini Diffusion~\citep{deepmind2025geminidiffu}, LLaDA~\citep{nie2025largelanguagediffusionmodels, zhu2025llada15variancereducedpreference}, and Dream~\citep{ye2025dream} demonstrate the practical promise of parallel text generation. However, current dLLMs remain limited by expensive repeated full-sequence computation and hard masking decisions: reducing denoising steps improves latency but often degrades quality. This creates the central question addressed by our work: how to improve generation quality while also reducing inference cost.

\textbf{Efficient and concurrent dLLM decoding.}
Recent efficient dLLM methods mainly reduce redundant computation through caching or adaptive decoding. dLLM-Cache~\citep{liu2025dllmcacheacceleratingdiffusionlarge}, Fast-dLLM~\citep{wu2025fastdllmtrainingfreeaccelerationdiffusion}, and Sparse-dLLM~\citep{song2025sparse} introduce KV-cache reuse, sparse attention, or cache eviction mechanisms tailored to bidirectional diffusion decoding. Other methods reduce the number of decoding steps: Prophet~\citep{li2025diffusion} uses confidence-gap-based early commitment, DAEDAL~\citep{li2025beyond} adapts generation length through EOS confidence and low-confidence region detection, and related training-free accelerators explore alternative scheduling strategies~\citep{ben2025accelerated, yu2025dimplediscretediffusionmultimodal, ma2025dkv, israel2025accelerating}.  SlowFast Sampling~\citep{wei2025slowfast} reports large speedups under a different decoding protocol. Orthogonal training-time work such as HDLM~\citep{zhou2025hdlm} and MolHIT~\citep{jung2026molhit} modifies the discrete diffusion state space, while LRD is training-free and acts only at inference time. Inspired by mixed-token reasoning improvements in autoregressive models~\citep{zhang2025softthinkingunlockingreasoning, wang2024guidinglanguagemodelreasoning, hao2024training}, LRD targets a different point in the design space: it improves dLLM belief states through soft embedding refinement while using KL convergence to avoid unnecessary computation.

\rev{\textbf{Predictive feedback and continuous states.} Soft-Masked DLMs~\citep{hersche2025softmasked}, EvoToken-DLM~\citep{zhong2026evotoken}, and RCD~\citep{hu2026rcd} also retain intermediate predictive information, using training to support soft or residual states. LRD's contribution is a training-free, instance-level refinement framework for frozen dLLMs, combining convergence control, progressive commitment, and conditional stability analysis. Appendix~\ref{appendix:recent_comparisons} compares these approaches and distinguishes direct decoding comparisons from accelerator composability; Appendix~\ref{appendix:corrector_connection} discusses the connection to predictor--corrector sampling.}

\section{Conclusion}
We introduced Latent Refinement Decoding, a two-stage framework that mitigates information loss from hard masking by refining global beliefs in embedding space before progressively finalising confident tokens under KL-based convergence monitoring.
Across code generation (HumanEval +6.3, MBPP +2.6), mathematical reasoning (GSM8K +2.9, MATH500 +3.8), QA, and summarisation benchmarks, LRD improves both quality and inference efficiency, works with single-token, multi-token, and dynamic/system decoding policies, and reaches 15.2$\times$ speedup on Dream-Base and 18.4$\times$ on LLaDA-Ins when combined with Fast-dLLM. While LRD consistently improves both quality and efficiency over existing dLLM decoding strategies, the broader comparison between diffusion and autoregressive language models remains an open question for future research.

\clearpage
\begin{revision}
\renewcommand{\acksection}{\section*{Acknowledgments}}
\begin{ack}
This work was supported in part by the UK Engineering and Physical Sciences Research Council (EPSRC) through a Turing AI Fellowship (grant nos.\ EP/V020579/1 and EP/V020579/2) and King's College London's Impact Acceleration Account (grant no.\ EP/X525571/1). The authors also acknowledge the use of the King's Computational Research, Engineering, and Technology Environment (CREATE) at King's College London.
\end{ack}
\end{revision}

\bibliographystyle{unsrtnat}
\bibliography{example_paper,references_revision}

\appendix
\newpage
\begingroup
\hypersetup{linkcolor=black,linktoc=all}
\renewcommand{\contentsname}{Appendix}
\setlength{\parskip}{4.8pt}
% Hide main-paper entries until this marker in the generated contents file.
\edef\appendixsavedtocdepth{\number\value{tocdepth}}
\setcounter{tocdepth}{-1}
\addtocontents{toc}{\protect\setcounter{tocdepth}{3}}
\tableofcontents
\setcounter{tocdepth}{\appendixsavedtocdepth}
\endgroup
\newpage

\FloatBarrier
\section{Additional Experiments and Implementation Details}
\label{appendix:Experiment}

This appendix provides material complementary to the main paper: additional experiments and implementation details (App.~\ref{appendix:Experiment}), ablations and hyperparameter analyses (App.~\ref{appendix:ablations_full}), and theoretical support including stability, true-posterior derivation, and consistency proofs (App.~\ref{appendix:theory}).

\FloatBarrier
\subsection{Experiment Settings}
\label{appendix:Experiment-settings}
For Base models, we follow standard few-shot settings for each benchmark: zero-shot for HumanEval, 3-shot for MBPP, 4-shot for MATH500, and 8-shot for GSM8K. For all benchmarks, we report accuracy for mathematical reasoning and pass@1 for code generation. We set the nucleus threshold to top-$p=0.9$. The hyperparameter $r_f$ is varied between $0.1$ and $0.2$. The thresholds for stopping latent refinement and early decoding are $\tau_{\text{refine}} = 0.1$ and $\tau_{\text{decode}} = 0.1$, respectively. We cap the latent refinement stage at a maximum of $T_{\text{refine}}=20$ steps.
For the Fast-dLLM rows in the main experiments, we report the dual cache + parallel decoding configuration.
For LLaDA-Instruct and LLaDA-1.5 models, generation is conducted under the official semi-AR framework~\citep{nie2025largelanguagediffusionmodels}, where the sequence is divided into blocks and decoded autoregressively at the block level. Within each block, instead of the standard hard masking used in the original work, we integrate our Latent Refinement and Predictive Feedback Loop, enabling refinement of token distributions before discrete commitment. The integration steps are described below.

\paragraph{Integration with Semi-AR Framework.}
\label{appendix:semi-ar}

In the semi-AR setting in LLaDA~\citep{nie2025largelanguagediffusionmodels}, a sequence of length $L$ is partitioned into $B$ blocks $\{b_1, b_2, ..., b_B\}$. While their original work uses standard hard masking within each block, we apply soft embeddings as follows:

For each block $b_i$ conditioned on previously generated blocks $\{b_1, ..., b_{i-1}\}$:
\begin{enumerate}
    \item \textbf{Soft Refinement:} Initialise positions in $b_i$ with \texttt{[MASK]} embeddings, then apply soft embedding refinement (Equation~\ref{eq:mix-embedding}) until convergence.
    \item \textbf{Progressive Decoding:} Use the converged soft embeddings to guide token selection within the block.
\end{enumerate}

\paragraph{Recovering hard-assignment baselines.}
The original hard-masking decoders in Dream and LLaDA can be recovered as a special case of our framework by setting the mixing ratio to zero ($r_f = 0$, so undecided positions use the plain $[\texttt{MASK}]$ embedding), disabling latent refinement ($T_{\text{refine}} = 0$), and turning off KL-based early stopping (e.g., $\tau_{\text{decode}} = 0$). Under $(r_f = 0,\; T_{\text{refine}} = 0,\; \tau_{\text{decode}} = 0)$, our sampler is equivalent to the original hard-assignment baselines.

\FloatBarrier
\subsection{Results}
\label{appendix:results}
\label{appendix:recent_comparisons}

This section organises the main appendix results around the commitment regimes supported by LRD: single-token LRD-only decoding, multi-token threshold commitment, and dynamic/system acceleration with DAEDAL. We then report additional datasets and model scales\rev{, including integrations with model-specific native decoding policies}.

\subsubsection{Full LRD-only Results}
\label{appendix:full_lrd_results}

Table~\ref{tab:full_lrd_results} reports single-token LRD decoding across additional model variants and lengths.

\begin{table*}[!htbp]
\caption{Full LRD-only results across additional models and sequence lengths. Speed denotes relative runtime (baseline = 1.0$\times$), where larger values indicate faster inference. Accuracy deltas in green are relative to the corresponding baseline row.}
\centering
\renewcommand{\arraystretch}{1}
\resizebox{1\linewidth}{!}{
\begin{tabular}{lcllclclclc}
\toprule[1pt]
\multirow{2}{*}{\textbf{Model}} & \multirow{2}{*}{\textbf{Len}} & \multirow{2}{*}{\textbf{Method}} & \multicolumn{2}{c}{\textbf{HumanEval}} & \multicolumn{2}{c}{\textbf{MBPP}} & \multicolumn{2}{c}{\textbf{GSM8K}} & \multicolumn{2}{c}{\textbf{MATH500}} \\
\cmidrule(lr){4-5} \cmidrule(lr){6-7} \cmidrule(lr){8-9} \cmidrule(lr){10-11}
& & & Acc & Speed & Acc & Speed & Acc & Speed & Acc & Speed \\
\hline
\multirow{6}{*}{\textbf{Dream-Base-7B}} & \multirow{2}{*}{256} & baseline & \textcolor{DarkGrey}{50.6} & \textcolor{DarkGrey}{1.0$\times$} & \textcolor{DarkGrey}{55.8} & \textcolor{DarkGrey}{1.0$\times$} & \textcolor{DarkGrey}{75.3} & \textcolor{DarkGrey}{1.0$\times$} & \textcolor{DarkGrey}{36.9} & \textcolor{DarkGrey}{1.0$\times$} \\
& & LRD & 56.9$_{\textcolor{LightGreen}{+6.3}}$ & 1.2$\times$ & 57.6$_{\textcolor{LightGreen}{+1.8}}$ & 2.3$\times$ & 78.2$_{\textcolor{LightGreen}{+2.9}}$ & 1.8$\times$ & 39.8$_{\textcolor{LightGreen}{+2.9}}$ & 1.4$\times$\\
& \multirow{2}{*}{512} & baseline & \textcolor{DarkGrey}{54.4} & \textcolor{DarkGrey}{1.0$\times$} & \textcolor{DarkGrey}{55.8} & \textcolor{DarkGrey}{1.0$\times$} & \textcolor{DarkGrey}{76.2} & \textcolor{DarkGrey}{1.0$\times$} & \textcolor{DarkGrey}{37.5} & \textcolor{DarkGrey}{1.0$\times$} \\
& & LRD & 58.8$_{\textcolor{LightGreen}{+4.4}}$ & 2.6$\times$ & 58.4$_{\textcolor{LightGreen}{+2.6}}$ & 4.5$\times$ & 77.4$_{\textcolor{LightGreen}{+1.2}}$ & 3.4$\times$ & 40.8$_{\textcolor{LightGreen}{+3.3}}$ & 1.8$\times$ \\
& \multirow{2}{*}{1024} & baseline & \textcolor{DarkGrey}{54.8} & \textcolor{DarkGrey}{1.0$\times$} & \textcolor{DarkGrey}{58.0} & \textcolor{DarkGrey}{1.0$\times$} & \textcolor{DarkGrey}{76.8} & \textcolor{DarkGrey}{1.0$\times$} & \textcolor{DarkGrey}{39.1} & \textcolor{DarkGrey}{1.0$\times$} \\
& & LRD & 59.1$_{\textcolor{LightGreen}{+4.3}}$ & 4.4$\times$ & 58.8$_{\textcolor{LightGreen}{+0.8}}$ & 7.6$\times$ & 77.8$_{\textcolor{LightGreen}{+1.0}}$ & 4.2$\times$ & 42.4$_{\textcolor{LightGreen}{+3.3}}$ & 2.2$\times$\\
\hline
\multirow{6}{*}{\textbf{Dream-Ins-7B}} & \multirow{2}{*}{256} & baseline & \textcolor{DarkGrey}{55.4} & \textcolor{DarkGrey}{1.0$\times$} & \textcolor{DarkGrey}{57.4} & \textcolor{DarkGrey}{1.0$\times$} & \textcolor{DarkGrey}{80.8} & \textcolor{DarkGrey}{1.0$\times$} & \textcolor{DarkGrey}{37.9} & \textcolor{DarkGrey}{1.0$\times$} \\
& & LRD & 61.6$_{\textcolor{LightGreen}{+6.2}}$ & 1.4$\times$ & 59.4$_{\textcolor{LightGreen}{+2.0}}$ & 2.4$\times$ & 83.0$_{\textcolor{LightGreen}{+2.2}}$ & 1.4$\times$ & 40.6$_{\textcolor{LightGreen}{+2.7}}$ & 1.1$\times$ \\
& \multirow{2}{*}{512} & baseline & \textcolor{DarkGrey}{56.1} & \textcolor{DarkGrey}{1.0$\times$} & \textcolor{DarkGrey}{56.7} & \textcolor{DarkGrey}{1.0$\times$} & \textcolor{DarkGrey}{80.2} & \textcolor{DarkGrey}{1.0$\times$} & \textcolor{DarkGrey}{38.6} & \textcolor{DarkGrey}{1.0$\times$} \\
& & LRD & 60.9$_{\textcolor{LightGreen}{+4.8}}$ & 2.9$\times$ & 58.8$_{\textcolor{LightGreen}{+2.1}}$ & 4.6$\times$ & 82.7$_{\textcolor{LightGreen}{+2.5}}$ & 3.6$\times$ & 41.8$_{\textcolor{LightGreen}{+3.2}}$ & 1.2$\times$ \\
& \multirow{2}{*}{1024} & baseline & \textcolor{DarkGrey}{56.0} & \textcolor{DarkGrey}{1.0$\times$} & \textcolor{DarkGrey}{57.3}  & \textcolor{DarkGrey}{1.0$\times$} & \textcolor{DarkGrey}{81.3} & \textcolor{DarkGrey}{1.0$\times$} & \textcolor{DarkGrey}{40.1} & \textcolor{DarkGrey}{1.0$\times$} \\
& & LRD & 61.0$_{\textcolor{LightGreen}{+5.0}}$ & 9.3$\times$ & 59.0$_{\textcolor{LightGreen}{+1.7}}$ & 10.6$\times$ & 83.5$_{\textcolor{LightGreen}{+2.2}}$ & 5.5$\times$ & 43.9$_{\textcolor{LightGreen}{+3.8}}$ & 1.7$\times$ \\
\hline
\multirow{6}{*}{\textbf{LLaDA-Base-8B}} & \multirow{2}{*}{256} & baseline & \textcolor{DarkGrey}{32.9} & \textcolor{DarkGrey}{1.0$\times$} & \textcolor{DarkGrey}{39.7} & \textcolor{DarkGrey}{1.0$\times$} & \textcolor{DarkGrey}{69.1} & \textcolor{DarkGrey}{1.0$\times$} &\textcolor{DarkGrey}{30.2} & \textcolor{DarkGrey}{1.0$\times$} \\
& & LRD & 36.0$_{\textcolor{LightGreen}{+3.1}}$ & 1.3$\times$ & 41.4$_{\textcolor{LightGreen}{+1.7}}$ & 1.5$\times$ & 71.2$_{\textcolor{LightGreen}{+2.1}}$ & 1.6$\times$ & 32.4$_{\textcolor{LightGreen}{+2.2}}$ & 1.4$\times$ \\
& \multirow{2}{*}{512} & baseline & \textcolor{DarkGrey}{32.8} & \textcolor{DarkGrey}{1.0$\times$} & \textcolor{DarkGrey}{39.8} & \textcolor{DarkGrey}{1.0$\times$} & \textcolor{DarkGrey}{70.8} & \textcolor{DarkGrey}{1.0$\times$} & \textcolor{DarkGrey}{30.8} & \textcolor{DarkGrey}{1.0$\times$} \\
& & LRD & 36.0$_{\textcolor{LightGreen}{+3.2}}$ & 1.7$\times$ & 41.4$_{\textcolor{LightGreen}{+1.6}}$ & 1.9$\times$ & 72.5$_{\textcolor{LightGreen}{+1.7}}$ & 2.2$\times$ &32.4$_{\textcolor{LightGreen}{+1.6}}$ & 1.6$\times$\\
& \multirow{2}{*}{1024} & baseline & \textcolor{DarkGrey}{31.7} & \textcolor{DarkGrey}{1.0$\times$} & \textcolor{DarkGrey}{39.8} & \textcolor{DarkGrey}{1.0$\times$} & \textcolor{DarkGrey}{71.4} & \textcolor{DarkGrey}{1.0$\times$} & \textcolor{DarkGrey}{30.1} & \textcolor{DarkGrey}{1.0$\times$} \\
& & LRD & 34.8$_{\textcolor{LightGreen}{+3.1}}$ & 2.2$\times$ & 40.8$_{\textcolor{LightGreen}{+1.0}}$ & 3.6$\times$ & 72.1$_{\textcolor{LightGreen}{+0.7}}$ & 3.3$\times$ & 32.2$_{\textcolor{LightGreen}{+2.1}}$ & 2.1$\times$\\
\hline
\multirow{6}{*}{\textbf{LLaDA-Ins-8B}} & \multirow{2}{*}{256} & baseline & \textcolor{DarkGrey}{38.7} & \textcolor{DarkGrey}{1.0$\times$} & \textcolor{DarkGrey}{36.9} & \textcolor{DarkGrey}{1.0$\times$} & \textcolor{DarkGrey}{77.4} & \textcolor{DarkGrey}{1.0$\times$} & \textcolor{DarkGrey}{33.8} & \textcolor{DarkGrey}{1.0$\times$} \\
& & LRD & 43.3$_{\textcolor{LightGreen}{+4.6}}$ & 1.2$\times$ & 40.0$_{\textcolor{LightGreen}{+3.1}}$ & 1.3$\times$ & 78.8$_{\textcolor{LightGreen}{+1.4}}$ & 1.5$\times$ & 35.8$_{\textcolor{LightGreen}{+2.0}}$ & 1.4$\times$ \\
& \multirow{2}{*}{512} & baseline & \textcolor{DarkGrey}{43.9} & \textcolor{DarkGrey}{1.0$\times$} & \textcolor{DarkGrey}{38.2} & \textcolor{DarkGrey}{1.0$\times$} & \textcolor{DarkGrey}{81.3} & \textcolor{DarkGrey}{1.0$\times$} & \textcolor{DarkGrey}{37.7} & \textcolor{DarkGrey}{1.0$\times$} \\
& & LRD & 48.4$_{\textcolor{LightGreen}{+4.5}}$ & 1.3$\times$ & 40.6$_{\textcolor{LightGreen}{+2.4}}$ & 1.5$\times$ & 84.5$_{\textcolor{LightGreen}{+3.2}}$ & 2.0$\times$ & 39.8$_{\textcolor{LightGreen}{+2.1}}$ & 1.6$\times$\\
& \multirow{2}{*}{1024} & baseline & \textcolor{DarkGrey}{44.6} & \textcolor{DarkGrey}{1.0$\times$} & \textcolor{DarkGrey}{37.4} & \textcolor{DarkGrey}{1.0$\times$} & \textcolor{DarkGrey}{82.3} & \textcolor{DarkGrey}{1.0$\times$} & \textcolor{DarkGrey}{39.4} & \textcolor{DarkGrey}{1.0$\times$} \\
& & LRD & 49.5$_{\textcolor{LightGreen}{+4.9}}$ & 1.7$\times$ &39.6$_{\textcolor{LightGreen}{+2.2}}$ & 3.7$\times$&83.7$_{\textcolor{LightGreen}{+1.4}}$ &4.3$\times$ & 42.2$_{\textcolor{LightGreen}{+2.8}}$ &2.0$\times$ \\
\hline
\multirow{6}{*}{\textbf{LLaDA-1.5-8B}} & \multirow{2}{*}{256} & baseline & \textcolor{DarkGrey}{38.4} & \textcolor{DarkGrey}{1.0$\times$} & \textcolor{DarkGrey}{38.6} & \textcolor{DarkGrey}{1.0$\times$} & \textcolor{DarkGrey}{79.2} & \textcolor{DarkGrey}{1.0$\times$} & \textcolor{DarkGrey}{33.4} & \textcolor{DarkGrey}{1.0$\times$} \\
& & LRD & 44.5$_{\textcolor{LightGreen}{+6.1}}$ & 1.2$\times$ & 39.8$_{\textcolor{LightGreen}{+1.2}}$ & 1.3$\times$ & 80.4$_{\textcolor{LightGreen}{+1.2}}$ & 1.5$\times$ & 36.6$_{\textcolor{LightGreen}{+3.2}}$ & 1.3$\times$ \\
& \multirow{2}{*}{512} & baseline & \textcolor{DarkGrey}{45.1} & \textcolor{DarkGrey}{1.0$\times$} & \textcolor{DarkGrey}{37.6} & \textcolor{DarkGrey}{1.0$\times$} & \textcolor{DarkGrey}{82.9} & \textcolor{DarkGrey}{1.0$\times$} & \textcolor{DarkGrey}{38.6} & \textcolor{DarkGrey}{1.0$\times$} \\
& & LRD & 49.6$_{\textcolor{LightGreen}{+4.5}}$ & 1.2$\times$ & 40.2$_{\textcolor{LightGreen}{+2.6}}$ & 1.5$\times$ & 84.5$_{\textcolor{LightGreen}{+1.6}}$ & 1.9$\times$ & 41.0$_{\textcolor{LightGreen}{+2.4}}$ & 1.4$\times$\\
& \multirow{2}{*}{1024} & baseline & \textcolor{DarkGrey}{45.7} & \textcolor{DarkGrey}{1.0$\times$} & \textcolor{DarkGrey}{37.4} & \textcolor{DarkGrey}{1.0$\times$} & \textcolor{DarkGrey}{82.5} & \textcolor{DarkGrey}{1.0$\times$} & \textcolor{DarkGrey}{39.6} & \textcolor{DarkGrey}{1.0$\times$} \\
& & LRD & 50.6$_{\textcolor{LightGreen}{+4.9}}$ & 1.7$\times$ & 39.6$_{\textcolor{LightGreen}{+2.2}}$ &3.5$\times$ & 83.9$_{\textcolor{LightGreen}{+1.4}}$ &4.0$\times$ & 41.8$_{\textcolor{LightGreen}{+2.2}}$ &1.9$\times$ \\
\bottomrule[1pt]
\end{tabular}
}
\label{tab:full_lrd_results}
\end{table*}

\subsubsection{Comparison with Confidence/Entropy Sampling Strategies}
\label{appendix:sampling_strategies}

We further ask whether more calibrated confidence- or entropy-based commitment rules can avoid the quality and speed trade-off caused by committing too many tokens per step. We sweep four families of token-commitment strategies on Dream-Instruct-7B with $L=512$:
\begin{itemize}
    \item \textbf{Top $k$ by confidence:} commit the $k$ positions with highest confidence at each step, $k\in\{1,\dots,5\}$.
    \item \textbf{Top $k$ by entropy:} commit the $k$ positions with lowest entropy at each step, $k\in\{1,\dots,5\}$.
    \item \textbf{Confidence threshold:} commit positions whose confidence exceeds threshold $t\in\{0.6, 0.7, 0.8, 0.9, 0.95\}$. Note that confidence threshold $t{=}0.9$ is the parallel-decoding criterion adopted by Fast-dLLM~\citep{wu2025fastdllmtrainingfreeaccelerationdiffusion}, which is why Fast-dLLM and LRD compose naturally.
    \item \textbf{Entropy threshold:} commit positions whose entropy is below threshold $t\in\{0.1, 0.3, 0.8, 1.0, 2.0\}$.
\end{itemize}
Table~\ref{tab:sampling_strategies} reports accuracy and the corresponding speed for all 19 configurations across HumanEval, MBPP, GSM8K, and MATH-500. The rows for the top $k$ rules serve as reference points for naive parallel commitment; the main comparison here is whether rules based on thresholds provide a better alternative.

\begin{table*}[!htbp]
\caption{Sampling strategy sweep on Dream-Instruct-7B ($L=512$). Speed is normalised to the baseline. ``Top $k=1$ by confidence'' and ``Top $k=1$ by entropy'' coincide with the standard baseline decoder.}
\label{tab:sampling_strategies}
\centering
\small
\resizebox{0.9 \linewidth}{!}{
\begin{tabular}{lcccccccc}
\toprule
& \multicolumn{2}{c}{HumanEval} & \multicolumn{2}{c}{MBPP} & \multicolumn{2}{c}{GSM8K} & \multicolumn{2}{c}{MATH-500} \\
\cmidrule(lr){2-3} \cmidrule(lr){4-5} \cmidrule(lr){6-7} \cmidrule(lr){8-9}
Method & Acc. & Speed & Acc. & Speed & Acc. & Speed & Acc. & Speed \\
\midrule
Baseline       & 56.1 & 1.0$\times$ & 56.7 & 1.0$\times$ & 80.2 & 1.0$\times$ & 38.6 & 1.0$\times$ \\
\midrule
Top $k=1$ by confidence   & 55.1 & 1.0$\times$ & 55.0 & 1.0$\times$ & 80.7 & 1.0$\times$ & 37.4 & 1.0$\times$ \\
Top $k=2$ by confidence   & 41.9 & 1.9$\times$ & 43.0 & 1.9$\times$ & 75.9 & 2.1$\times$ & 29.0 & 1.9$\times$ \\
Top $k=3$ by confidence   & 26.7 & 2.9$\times$ & 27.2 & 2.8$\times$ & 66.1 & 3.1$\times$ & 26.2 & 2.9$\times$ \\
Top $k=4$ by confidence   & 21.6 & 3.8$\times$ & 22.2 & 3.9$\times$ & 57.7 & 4.1$\times$ & 14.0 & 3.9$\times$ \\
Top $k=5$ by confidence   & 12.9 & 4.7$\times$ & 20.8 & 4.9$\times$ & 47.5 & 5.1$\times$ &  8.4 & 4.8$\times$ \\
\midrule
Top $k=1$ by entropy    & 56.1 & 1.0$\times$ & 56.7 & 1.0$\times$ & 80.2 & 1.0$\times$ & 38.6 & 1.0$\times$ \\
Top $k=2$ by entropy    & 43.3 & 1.9$\times$ & 44.2 & 2.1$\times$ & 77.1 & 1.9$\times$ & 30.0 & 2.1$\times$ \\
Top $k=3$ by entropy    & 27.9 & 2.8$\times$ & 28.0 & 2.9$\times$ & 66.9 & 3.1$\times$ & 27.4 & 2.9$\times$ \\
Top $k=4$ by entropy    & 22.5 & 3.9$\times$ & 23.0 & 3.8$\times$ & 58.9 & 4.1$\times$ & 15.2 & 3.9$\times$ \\
Top $k=5$ by entropy    & 13.5 & 4.8$\times$ & 21.6 & 4.9$\times$ & 49.1 & 5.1$\times$ &  8.8 & 4.9$\times$ \\
\midrule
Confidence threshold $t=0.95$ & 52.4 & 2.7$\times$ & 57.4 & 4.6$\times$ & 80.4 & 4.5$\times$ & 38.8 & 2.4$\times$ \\
Confidence threshold $t=0.9$  & 52.4 & 3.0$\times$ & 57.4 & 4.9$\times$ & 80.0 & 4.9$\times$ & 38.4 & 2.6$\times$ \\
Confidence threshold $t=0.8$  & 50.0 & 3.4$\times$ & 56.6 & 5.1$\times$ & 79.6 & 5.2$\times$ & 37.8 & 3.0$\times$ \\
Confidence threshold $t=0.7$  & 46.3 & 3.6$\times$ & 56.0 & 5.4$\times$ & 77.2 & 5.6$\times$ & 37.0 & 3.3$\times$ \\
Confidence threshold $t=0.6$  & 42.7 & 4.0$\times$ & 50.8 & 5.6$\times$ & 74.8 & 6.0$\times$ & 35.6 & 3.7$\times$ \\
\midrule
Entropy threshold $t=0.1$   & 53.0 & 2.2$\times$ & 56.6 & 4.1$\times$ & 80.6 & 3.6$\times$ & 39.0 & 1.9$\times$ \\
Entropy threshold $t=0.3$   & 53.0 & 2.5$\times$ & 56.8 & 4.4$\times$ & 80.4 & 4.0$\times$ & 38.8 & 2.2$\times$ \\
Entropy threshold $t=0.8$   & 47.6 & 3.0$\times$ & 54.4 & 4.8$\times$ & 77.8 & 4.9$\times$ & 36.4 & 2.9$\times$ \\
Entropy threshold $t=1.0$   & 45.7 & 3.2$\times$ & 51.6 & 5.0$\times$ & 74.0 & 5.1$\times$ & 33.6 & 3.0$\times$ \\
Entropy threshold $t=2.0$   & 29.9 & 4.1$\times$ & 33.2 & 5.3$\times$ & 56.4 & 6.0$\times$ & 21.8 & 4.1$\times$ \\
\midrule
\textbf{LRD (ours)} & 60.9 & 2.9$\times$ & 58.8 & 4.6$\times$ & 82.7 & 3.6$\times$ & 41.8 & 1.2$\times$ \\
\bottomrule
\end{tabular}}
\end{table*}

The threshold methods trace a smoother speed and accuracy trade-off than fixed top $k$ commitment, but the configurations that remain close to baseline accuracy (confidence threshold $t=0.95$, entropy threshold $t=0.1$, and entropy threshold $t=0.3$) do not meaningfully improve quality. Lowering the confidence threshold or raising the entropy threshold buys speed by committing more positions, but accuracy degrades. LRD is the only configuration in the table that simultaneously \emph{improves} accuracy across all four benchmarks ($+4.8$, $+2.1$, $+2.5$, $+3.2$) while accelerating decoding from $1.2\times$ to $4.6\times$, \rev{achieving the highest accuracy on all four tasks, while some threshold configurations are faster}.

\begin{revision}
\FloatBarrier
\subsubsection{Additional Training-Free Decoding Comparisons}
\label{appendix:soar_credit}

SOAR~\citep{cao2026soar} searches alternative commitment orders when confidence is low and decodes in parallel when confidence is high. CreditDecoding~\citep{wang2025creditdecoding} accumulates historical evidence to accelerate commitment. Both are direct training-free decoding comparisons.
Table~\ref{tab:soar_credit} compares SOAR and CreditDecoding on LLaDA-Instruct-8B under matched evaluation and hardware. Each cell reports accuracy and wall-clock speed relative to same-length native decoding. LRD achieves the highest accuracy in all eight model-length/task settings; SOAR is faster in several coding settings. These results separate quality improvement from the distinct objective of maximising throughput.

\begin{table}[htbp]
\ifshowchanges\color{blue}\captionsetup{font+={color=blue},labelfont+={color=blue}}\fi
\centering
\caption{Additional training-free decoders on LLaDA-Instruct-8B. Cells are accuracy / speedup.}
\label{tab:soar_credit}
\small
\resizebox{\textwidth}{!}{%
\begin{tabular}{clcccc}
\toprule
Length & Method & HumanEval & MBPP & GSM8K & MATH500 \\
\midrule
256 & Baseline & 38.7 / 1.0$\times$ & 36.9 / 1.0$\times$ & 77.4 / 1.0$\times$ & 33.8 / 1.0$\times$ \\
256 & SOAR & 40.4 / 1.3$\times$ & 37.6 / 1.3$\times$ & 78.2 / 1.2$\times$ & 34.7 / 1.0$\times$ \\
256 & CreditDecoding & 40.3 / 1.2$\times$ & 37.5 / 1.3$\times$ & 76.9 / 1.4$\times$ & 33.6 / 1.3$\times$ \\
256 & LRD & 43.3 / 1.2$\times$ & 40.0 / 1.3$\times$ & 78.8 / 1.5$\times$ & 35.8 / 1.4$\times$ \\
\midrule
512 & Baseline & 43.9 / 1.0$\times$ & 38.2 / 1.0$\times$ & 81.3 / 1.0$\times$ & 37.7 / 1.0$\times$ \\
512 & SOAR & 46.6 / 1.7$\times$ & 39.2 / 1.6$\times$ & 82.1 / 1.4$\times$ & 39.0 / 1.2$\times$ \\
512 & CreditDecoding & 45.6 / 1.3$\times$ & 38.9 / 1.5$\times$ & 80.7 / 1.9$\times$ & 37.6 / 1.5$\times$ \\
512 & LRD & 48.4 / 1.3$\times$ & 40.6 / 1.5$\times$ & 84.5 / 2.0$\times$ & 39.8 / 1.6$\times$ \\
\bottomrule
\end{tabular}}
\end{table}

\FloatBarrier
\subsubsection{Training-Based Predictive-Feedback Methods}
\label{appendix:feedback_positioning}

Soft-Masked DLMs~\citep{hersche2025softmasked} train mask--prediction embedding mixtures, EvoToken-DLM~\citep{zhong2026evotoken} supervises evolving soft token distributions, and RCD~\citep{hu2026rcd} learns contextual residuals. These methods adapt model parameters, whereas LRD refines a frozen model's beliefs separately for each input.

Table~\ref{tab:trained_feedback} uses a generation length of 512 and the full forward-pass budget. The corresponding Dream coding checkpoint for Soft-Masked DLMs was unavailable, so its coding numbers are paper-reported and are not a matched-checkpoint comparison. EvoToken-DLM uses its released 10k-step LoRA on LLaDA-Instruct-8B; RCD uses the released RCD-LLaDA-8B-Instruct weights. LRD uses the corresponding frozen Dream-Instruct-7B or LLaDA-Instruct-8B model, with the same prompts, answer extraction, decoding budgets, and evaluators in the reproduced comparisons. Since the trained methods modify the model, this table compares adaptation strategies rather than isolating a decoder on identical weights. LRD does not dominate every entry: EvoToken-DLM is slightly better on MATH500.

\begin{table}[htbp]
\ifshowchanges\color{blue}\captionsetup{font+={color=blue},labelfont+={color=blue}}\fi
\centering
\caption{Training-based predictive feedback versus frozen-model LRD. Soft-Masked DLM results are quoted; the other comparisons use released weights.}
\label{tab:trained_feedback}
\small
\begin{tabular}{llcc}
\toprule
Method & Benchmark & Method accuracy & LRD accuracy \\
\midrule
Soft-Masked DLMs & HumanEval & 57.8 & 60.9 \\
Soft-Masked DLMs & MBPP & 56.4 & 58.8 \\
EvoToken-DLM & GSM8K & 84.0 & 84.5 \\
EvoToken-DLM & MATH500 & 40.0 & 39.8 \\
RCD & GSM8K & 79.1 & 84.5 \\
RCD & MATH500 & 37.0 & 39.8 \\
\bottomrule
\end{tabular}
\end{table}
\end{revision}

\FloatBarrier
\subsubsection{Comparison with Acceleration Baselines}
\label{appendix:accel_baselines}

We further compare LRD against dynamic training-free acceleration strategies for masked diffusion LMs. DAEDAL~\citep{li2025beyond} performs length-adaptive denoising by terminating sequences via EOS-confidence detection. Table~\ref{tab:accel_compare} reports matched-compute results on Dream-Instruct-7B with $L=512$ on a single A100; Prophet~\citep{li2025diffusion} and SlowFast Sampling~\citep{wei2025slowfast} are already included in the main comparison table where matched measurements are available.

\begin{table}[!htbp]
\caption{Comparison with acceleration baselines on Dream-Instruct-7B ($L=512$). All speedups are normalised to the same baseline. Combinations such as DAEDAL+LRD and Fast-dLLM+LRD demonstrate composability.}
\label{tab:accel_compare}
\centering
\small
\begin{tabular}{lcccccccc}
\toprule
& \multicolumn{2}{c}{HumanEval} & \multicolumn{2}{c}{MBPP} & \multicolumn{2}{c}{GSM8K} & \multicolumn{2}{c}{MATH500} \\
\cmidrule(lr){2-3} \cmidrule(lr){4-5} \cmidrule(lr){6-7} \cmidrule(lr){8-9}
Method & Acc. & Speed & Acc. & Speed & Acc. & Speed & Acc. & Speed \\
\midrule
Baseline & 56.1 & 1.0$\times$ & 56.7 & 1.0$\times$ & 80.2 & 1.0$\times$ & 38.6 & 1.0$\times$ \\
DAEDAL~\citep{li2025beyond} & 54.3 & 3.2$\times$ & 56.0 & \textbf{7.7$\times$} & 79.1 & 8.4$\times$ & 38.2 & 1.2$\times$ \\
DAEDAL+LRD & 58.8 & \textbf{3.3$\times$} & 58.4 & 7.6$\times$ & 81.9 & \textbf{8.6$\times$} & 41.0 & 1.2$\times$ \\
LRD (ours) & \textbf{60.9} & 2.9$\times$ & \textbf{58.8} & 4.6$\times$ & \textbf{82.7} & 3.6$\times$ & \textbf{41.8} & 1.2$\times$ \\
\bottomrule
\end{tabular}
\end{table}

DAEDAL alone trades accuracy for substantially larger speedups on some tasks; combining DAEDAL with LRD recovers accuracy above the baseline while preserving most of DAEDAL's speed advantage, mirroring the composability we observe with Fast-dLLM in the main results.
\rev{These combinations test composability with system or dynamic acceleration; the direct decoding comparisons are reported in Appendix~\ref{appendix:soar_credit}.}

\FloatBarrier
\subsubsection{Additional Results on General QA, Summarisation, and Knowledge-Based QA}
\label{appendix:general_qa}

\textbf{LRD maintains general QA and summarisation quality, and improves knowledge-based QA.}
To verify that LRD is not over-specialised to reasoning benchmarks, we evaluate it on four non-code and non-math benchmarks using Dream-Ins-7B. These benchmarks cover context-grounded tasks, including reading comprehension on RACE~\cite{lai-etal-2017-race} and summarisation on CNN/DailyMail~\cite{see-etal-2017-get}, as well as knowledge-oriented QA tasks, including NQ-Open~\cite{kwiatkowski-etal-2019-natural} and TruthfulQA~\cite{lin-etal-2022-truthfulqa}. This broader evaluation tests whether LRD preserves general language quality beyond the reasoning and coding benchmarks used in the main results.

Tables~\ref{tab:race_truthfulqa_table} and~\ref{tab:cnn_nqopen_table} report these general-language results. Across all context lengths, LRD matches or improves the baseline on RACE accuracy ($+1.1$ to $+2.1$) and CNN/DailyMail metrics (BERTScore, ROUGE-1/L), while NQ-Open EM rises by $+1.4$ to $+3.0$ and TruthfulQA improves by $+1.4$ to $+3.0$ across the three accuracy variants. Speed gains differ markedly between task types: shorter-answer tasks where early stopping fires aggressively (TruthfulQA, NQ-Open) reach from $2.1\times$ to $2.9\times$, whereas longer-output tasks (RACE selection over long passages, CNN/DailyMail summaries) see more modest speedups from $1.3\times$ to $1.7\times$. The improvements on knowledge-based QA are notable because they cannot be attributed to better use of in-prompt context, suggesting that the soft-embedding refinement helps the model produce more coherent factual completions in addition to better contextual reasoning. Together, these results indicate that LRD preserves and in some cases enhances both context-based and knowledge-based language abilities, rather than trading off language quality for efficiency.

\begin{table*}[!htbp]
\caption{Closed-selection benchmarks: reading comprehension (RACE) and TruthfulQA on Dream-Ins-7B. Baseline results are shown in \textcolor{DarkGrey}{grey}, and improvements of our method are highlighted in \textcolor{LightGreen}{green}. Speed is reported per task.}
\label{tab:race_truthfulqa_table}
\centering
\renewcommand{\arraystretch}{1}
\resizebox{0.9\linewidth}{!}{
\begin{tabular}{cl llllll}
\toprule[1pt]
\multirow{2}{*}{\textbf{Len}} & \multirow{2}{*}{\textbf{Method}}
 & \multicolumn{2}{c}{\textbf{RACE}}
 & \multicolumn{4}{c}{\textbf{TruthfulQA}} \\
\cmidrule(lr){3-4} \cmidrule(lr){5-8}
 & & Acc & Speed & bleu\_acc & rouge1\_acc & rouge2\_acc & Speed \\
\hline
\multirow{2}{*}{256}
 & baseline
 & \textcolor{DarkGrey}{81.1} & \textcolor{DarkGrey}{1.0$\times$}
 & \textcolor{DarkGrey}{40.3} & \textcolor{DarkGrey}{45.3} & \textcolor{DarkGrey}{32.8} & \textcolor{DarkGrey}{1.0$\times$} \\
 & Ours
 & 82.2$_{\textcolor{LightGreen}{+1.1}}$ & 1.3$\times$
 & 42.5$_{\textcolor{LightGreen}{+2.2}}$ & 47.6$_{\textcolor{LightGreen}{+2.3}}$ & 34.2$_{\textcolor{LightGreen}{+1.4}}$ & 2.1$\times$ \\
\hline
\multirow{2}{*}{512}
 & baseline
 & \textcolor{DarkGrey}{80.9} & \textcolor{DarkGrey}{1.0$\times$}
 & \textcolor{DarkGrey}{41.9} & \textcolor{DarkGrey}{45.4} & \textcolor{DarkGrey}{33.1} & \textcolor{DarkGrey}{1.0$\times$} \\
 & Ours
 & 82.4$_{\textcolor{LightGreen}{+1.5}}$ & 1.4$\times$
 & 44.3$_{\textcolor{LightGreen}{+2.4}}$ & 48.3$_{\textcolor{LightGreen}{+2.9}}$ & 35.1$_{\textcolor{LightGreen}{+2.0}}$ & 2.6$\times$ \\
\hline
\multirow{2}{*}{1024}
 & baseline
 & \textcolor{DarkGrey}{81.2} & \textcolor{DarkGrey}{1.0$\times$}
 & \textcolor{DarkGrey}{41.7} & \textcolor{DarkGrey}{45.5} & \textcolor{DarkGrey}{33.6} & \textcolor{DarkGrey}{1.0$\times$} \\
 & Ours
 & 83.3$_{\textcolor{LightGreen}{+2.1}}$ & 1.7$\times$
 & 44.7$_{\textcolor{LightGreen}{+3.0}}$ & 47.5$_{\textcolor{LightGreen}{+2.0}}$ & 35.1$_{\textcolor{LightGreen}{+1.5}}$ & 2.9$\times$ \\
\bottomrule[1pt]
\end{tabular}
}
\end{table*}

\begin{table*}[!htbp]
\caption{Open-generation benchmarks: summarisation (CNN/DailyMail) and open-domain QA (NQ-Open) on Dream-Ins-7B. Baseline results are shown in \textcolor{DarkGrey}{grey}, and improvements of our method are highlighted in \textcolor{LightGreen}{green}. Speed is reported per task.}
\label{tab:cnn_nqopen_table}
\centering
\renewcommand{\arraystretch}{1}
\resizebox{0.9 \linewidth}{!}{
\begin{tabular}{cl cccc cc}
\toprule[1pt]
\multirow{2}{*}{\textbf{Len}} & \multirow{2}{*}{\textbf{Method}}
 & \multicolumn{4}{c}{\textbf{CNN/DailyMail}}
 & \multicolumn{2}{c}{\textbf{NQ-Open}} \\
\cmidrule(lr){3-6} \cmidrule(lr){7-8}
 & & BERTScore & ROUGE-1 & ROUGE-L & Speed & EM & Speed \\
\hline
\multirow{2}{*}{256}
 & baseline
 & \textcolor{DarkGrey}{85.6} & \textcolor{DarkGrey}{21.7} & \textcolor{DarkGrey}{15.0} & \textcolor{DarkGrey}{1.0$\times$}
 & \textcolor{DarkGrey}{12.0} & \textcolor{DarkGrey}{1.0$\times$} \\
 & Ours
 & 85.7$_{\textcolor{LightGreen}{+0.1}}$ & 22.3$_{\textcolor{LightGreen}{+0.6}}$ & 15.5$_{\textcolor{LightGreen}{+0.5}}$ & 1.3$\times$
 & 15.0$_{\textcolor{LightGreen}{+3.0}}$ & 2.2$\times$ \\
\hline
\multirow{2}{*}{512}
 & baseline
 & \textcolor{DarkGrey}{85.5} & \textcolor{DarkGrey}{22.6} & \textcolor{DarkGrey}{15.6} & \textcolor{DarkGrey}{1.0$\times$}
 & \textcolor{DarkGrey}{12.8} & \textcolor{DarkGrey}{1.0$\times$} \\
 & Ours
 & 85.6$_{\textcolor{LightGreen}{+0.1}}$ & 23.0$_{\textcolor{LightGreen}{+0.4}}$ & 15.9$_{\textcolor{LightGreen}{+0.3}}$ & 1.4$\times$
 & 14.4$_{\textcolor{LightGreen}{+1.6}}$ & 2.6$\times$ \\
\hline
\multirow{2}{*}{1024}
 & baseline
 & \textcolor{DarkGrey}{85.7} & \textcolor{DarkGrey}{22.8} & \textcolor{DarkGrey}{15.7} & \textcolor{DarkGrey}{1.0$\times$}
 & \textcolor{DarkGrey}{12.8} & \textcolor{DarkGrey}{1.0$\times$} \\
 & Ours
 & 85.8$_{\textcolor{LightGreen}{+0.1}}$ & 23.3$_{\textcolor{LightGreen}{+0.5}}$ & 16.5$_{\textcolor{LightGreen}{+0.8}}$ & 1.7$\times$
 & 14.2$_{\textcolor{LightGreen}{+1.4}}$ & 2.9$\times$ \\
\bottomrule[1pt]
\end{tabular}
}
\end{table*}

\begin{revision}
\FloatBarrier
\subsubsection{Additional Reasoning Benchmarks}
\label{appendix:additional_reasoning}

We also evaluate GPQA~\citep{rein2023gpqa} and BBH~\citep{suzgun2022bbh} on Dream-Instruct-7B (Table~\ref{tab:gpqa_bbh}). Across the four benchmark-length settings, accuracy increases by 2.5--2.9 points, with 1.2--2.9$\times$ wall-clock speedups over native decoding.

\begin{table}[htbp]
\ifshowchanges\color{blue}\captionsetup{font+={color=blue},labelfont+={color=blue}}\fi
\centering
\caption{Additional benchmarks on Dream-Instruct-7B.}
\label{tab:gpqa_bbh}
\small
\begin{tabular}{lcccc}
\toprule
Benchmark & Length & Native accuracy & LRD accuracy & LRD speedup \\
\midrule
GPQA & 256 & 34.2 & 36.8 & 1.2$\times$ \\
GPQA & 512 & 35.5 & 38.4 & 1.5$\times$ \\
BBH & 256 & 59.1 & 61.6 & 1.5$\times$ \\
BBH & 512 & 60.4 & 63.2 & 2.9$\times$ \\
\bottomrule
\end{tabular}
\end{table}
\end{revision}

\FloatBarrier
\subsubsection{Additional Results on LLaDA2.0-mini-preview (16B)}
\label{appendix:llada_2}

\begin{table*}[!htbp]
\caption{Performance of LLaDA2.0-mini-preview (16B) across benchmarks.
Speed denotes relative runtime (baseline = 1.0×), where larger values indicate faster inference.
Baseline results are shown in \textcolor{DarkGrey}{grey}, and our improvements in \textcolor{LightGreen}{green}.}
\centering
\renewcommand{\arraystretch}{1}
\resizebox{0.9 \linewidth}{!}{
\begin{tabular}{lcllclclclc}
\toprule[1pt]
\multirow{2}{*}{\textbf{Model}} & \multirow{2}{*}{\textbf{Len}} & \multirow{2}{*}{\textbf{Method}}
& \multicolumn{2}{c}{\textbf{HumanEval}}
& \multicolumn{2}{c}{\textbf{MBPP}}
& \multicolumn{2}{c}{\textbf{GSM8K}}
& \multicolumn{2}{c}{\textbf{MATH500}} \\
\cmidrule(lr){4-5} \cmidrule(lr){6-7} \cmidrule(lr){8-9} \cmidrule(lr){10-11}
& & & Acc & Speed & Acc & Speed & Acc & Speed & Acc & Speed \\
\hline
\multirow{6}{*}{\makecell{\textbf{LLaDA2.0-}\\\textbf{mini-preview}}}
& \multirow{2}{*}{256}
& baseline
& \textcolor{DarkGrey}{5.5} & \textcolor{DarkGrey}{1.0×}
& \textcolor{DarkGrey}{19.2} & \textcolor{DarkGrey}{1.0×}
& \textcolor{DarkGrey}{63.8} & \textcolor{DarkGrey}{1.0×}
& \textcolor{DarkGrey}{16.2} & \textcolor{DarkGrey}{1.0×} \\
& & Ours
& 7.3$_{\textcolor{LightGreen}{+1.8}}$ & 1.0×
& 22.2$_{\textcolor{LightGreen}{+3.0}}$ & 1.1×
& 65.6$_{\textcolor{LightGreen}{+1.8}}$ & 1.2×
& 17.5$_{\textcolor{LightGreen}{+1.3}}$ & 1.0× \\
\cmidrule(lr){2-11}

& \multirow{2}{*}{512}
& baseline
& \textcolor{DarkGrey}{54.3} & \textcolor{DarkGrey}{1.0×}
& \textcolor{DarkGrey}{54.7} & \textcolor{DarkGrey}{1.0×}
& \textcolor{DarkGrey}{86.5} & \textcolor{DarkGrey}{1.0×}
& \textcolor{DarkGrey}{44.6} & \textcolor{DarkGrey}{1.0×} \\
& & Ours
& 56.9$_{\textcolor{LightGreen}{+2.6}}$ & 1.1×
& 58.2$_{\textcolor{LightGreen}{+3.5}}$ & 1.6×
& 88.1$_{\textcolor{LightGreen}{+1.6}}$ & 2.1×
& 46.3$_{\textcolor{LightGreen}{+1.7}}$ & 1.1× \\
\cmidrule(lr){2-11}

& \multirow{2}{*}{1024}
& baseline
& \textcolor{DarkGrey}{74.2} & \textcolor{DarkGrey}{1.0×}
& \textcolor{DarkGrey}{63.2} & \textcolor{DarkGrey}{1.0×}
& \textcolor{DarkGrey}{87.7} & \textcolor{DarkGrey}{1.0×}
& \textcolor{DarkGrey}{61.2} & \textcolor{DarkGrey}{1.0×} \\
& & Ours
& 78.7$_{\textcolor{LightGreen}{+4.5}}$ & 1.1×
& 65.7$_{\textcolor{LightGreen}{+2.5}}$ & 2.2×
& 89.5$_{\textcolor{LightGreen}{+1.8}}$ & 3.8×
& 63.2$_{\textcolor{LightGreen}{+2.0}}$ & 1.8× \\
\bottomrule[1pt]
\end{tabular}}
\label{tab:llada2mini}
\end{table*}

Table~\ref{tab:llada2mini} reports results on the larger \texttt{LLaDA2.0-mini-preview} (16B)~\citep{nie2025largelanguagediffusionmodels} across the four benchmarks. Across all context lengths, LRD consistently improves accuracy and provides additional speedups over the baseline decoder. At length 256, LLaDA2.0 tends to produce very long outputs, which makes performance less stable in this short-context regime, but LRD still yields gains (e.g., +1.8 on HumanEval and +3.0 on MBPP); at 512 and 1024, where the model is better aligned with the available budget, LRD brings improvements of up to +4.5 points and from $2.2\times$ to $3.8\times$ faster decoding, indicating that the refinement scheme scales smoothly to larger dLLMs.

\begin{revision}
\FloatBarrier
\subsubsection{Additional Model Families}
\label{appendix:recent_models}

We additionally evaluate two model families without parameter updates. Nemotron-Labs-Diffusion~\citep{fu2026nemotron} uses block-wise parallel confidence sampling; LRD retains this native commitment policy. DiffusionGemma~\citep{diffusiongemma2026} uses noisy-token canvases, soft self-conditioning, Entropy-Bounded denoising, and adaptive stopping. The DiffusionGemma experiment therefore uses an LRD-style predictive residual alongside these native mechanisms, rather than replacing them with the standard two-phase masked decoder. Native adaptive stopping limits the additional speed gain. Each result in Table~\ref{tab:recent_models} is relative to that model's own native decoder; cross-model speed ratios are not comparable.

\begin{table}[htbp]
\ifshowchanges\color{blue}\captionsetup{font+={color=blue},labelfont+={color=blue}}\fi
\centering
\caption{Additional model families. Cells are accuracy / speedup over the corresponding native decoder. DiffusionGemma uses a model-specific LRD-style adaptation.}
\label{tab:recent_models}
\resizebox{\textwidth}{!}{%
\begin{tabular}{lcccc}
\toprule
Method & HumanEval & MBPP & GSM8K & MATH500 \\
\midrule
\multicolumn{5}{l}{Nemotron-Labs-Diffusion-8B-Instruct} \\
Native diffusion & 69.8 / 1.0$\times$ & 70.9 / 1.0$\times$ & 87.4 / 1.0$\times$ & 65.6 / 1.0$\times$ \\
+LRD & 72.4 / 1.3$\times$ & 73.6 / 1.5$\times$ & 89.0 / 1.2$\times$ & 68.0 / 1.1$\times$ \\
\midrule
\multicolumn{5}{l}{DiffusionGemma-26B-A4B-it} \\
Native EB + adaptive stopping & 76.3 / 1.0$\times$ & 73.8 / 1.0$\times$ & 89.6 / 1.0$\times$ & 68.2 / 1.0$\times$ \\
+LRD-style adaptation & 77.5 / 1.2$\times$ & 75.3 / 1.1$\times$ & 90.3 / 1.2$\times$ & 69.6 / 1.1$\times$ \\
\bottomrule
\end{tabular}}
\end{table}
\end{revision}

\FloatBarrier
\subsection{Efficiency and System Integration}
\label{appendix:efficiency}

\subsubsection{Efficiency Analysis}
\label{appendix:cost_pareto}

\begin{table*}[!htbp]
    \centering
    \caption{\textbf{Efficiency Analysis Breakdown (HumanEval, length=512).} We compare the per-step theoretical cost and wall-clock time between the Baseline and our method (Dream-Instruct-7B). Although our method introduces a minor overhead per step, the significant reduction in generation steps results in a substantial total speedup.}
    \label{tab:efficiency_breakdown}
    % \vspace{2mm}

    \resizebox{0.9\textwidth}{!}{
    \begin{tabular}{l c c l}
        \toprule
        \textbf{Per-Step Cost} & \textbf{FLOPs (G)} & \textbf{Latency} & \textbf{Remark / $\Delta$} \\
        \midrule

        \textbf{Baseline (Standard)} & \textbf{14.14} & \textbf{81.8 ms} & \textbf{N/A} \\
        \quad Backbone & 14.14 & 68.6 ms & MatMul dominated \\
        \quad System Overhead & N/A & 13.2 ms & Data dispatch \\

        \vspace{0.5mm}

        \textbf{LRD (Proposed)} & $\sim$\textbf{14.14} & \textbf{94.3 ms} & \textbf{N/A} \\
        \quad Backbone & 14.14 & 68.6 ms & Identical to baseline \\
        \quad KL / Mixing / Aux & $\approx 0$ & 12.6 ms & Memory-bound overhead \\
        \quad System Overhead & N/A & 13.1 ms & Similar to baseline \\

        \midrule
        % \multicolumn{4}{l}{\textit{\textbf{End-to-End Efficiency}}} \\
        \midrule

        Avg. Steps Needed & N/A & $512 \rightarrow 153$ & {\textbf{$\sim$70\% Reduction}} \\
        \textbf{Total Wall-clock} & N/A & \textbf{41.8s $\rightarrow$ 14.4s} & {\textbf{$\sim$2.9$\times$ Speedup}} \\
        \bottomrule
    \end{tabular}}
\end{table*}

\paragraph{Per-step overhead and wall-clock speed.}
To better characterise the efficiency of LRD beyond relative ``Speed'' ratios, we perform a detailed cost accounting on Dream-Instruct-7B using absolute wall-clock measurements and per-step FLOP estimates (Table~\ref{tab:efficiency_breakdown}). Per step, LRD adds only negligible additional FLOPs compared to the baseline diffusion decoder, since the backbone forward pass is identical (14.14G FLOPs, 68.6\,ms). However, the extra KL computation over the 152K-token vocabulary and the mixing operations introduce a memory-bound overhead, leading to a modest per-step latency. Crucially, LRD substantially reduces the number of required generation steps: the average number of steps drops from 512 to 153 (approximately 70\% fewer steps), which translates into a net end-to-end speedup of roughly 2.9$\times$ in total wall-clock time (41.8s $\rightarrow$ 14.4s).

\begin{figure}[!htbp]
    \centering
    \includegraphics[width=0.5\textwidth]{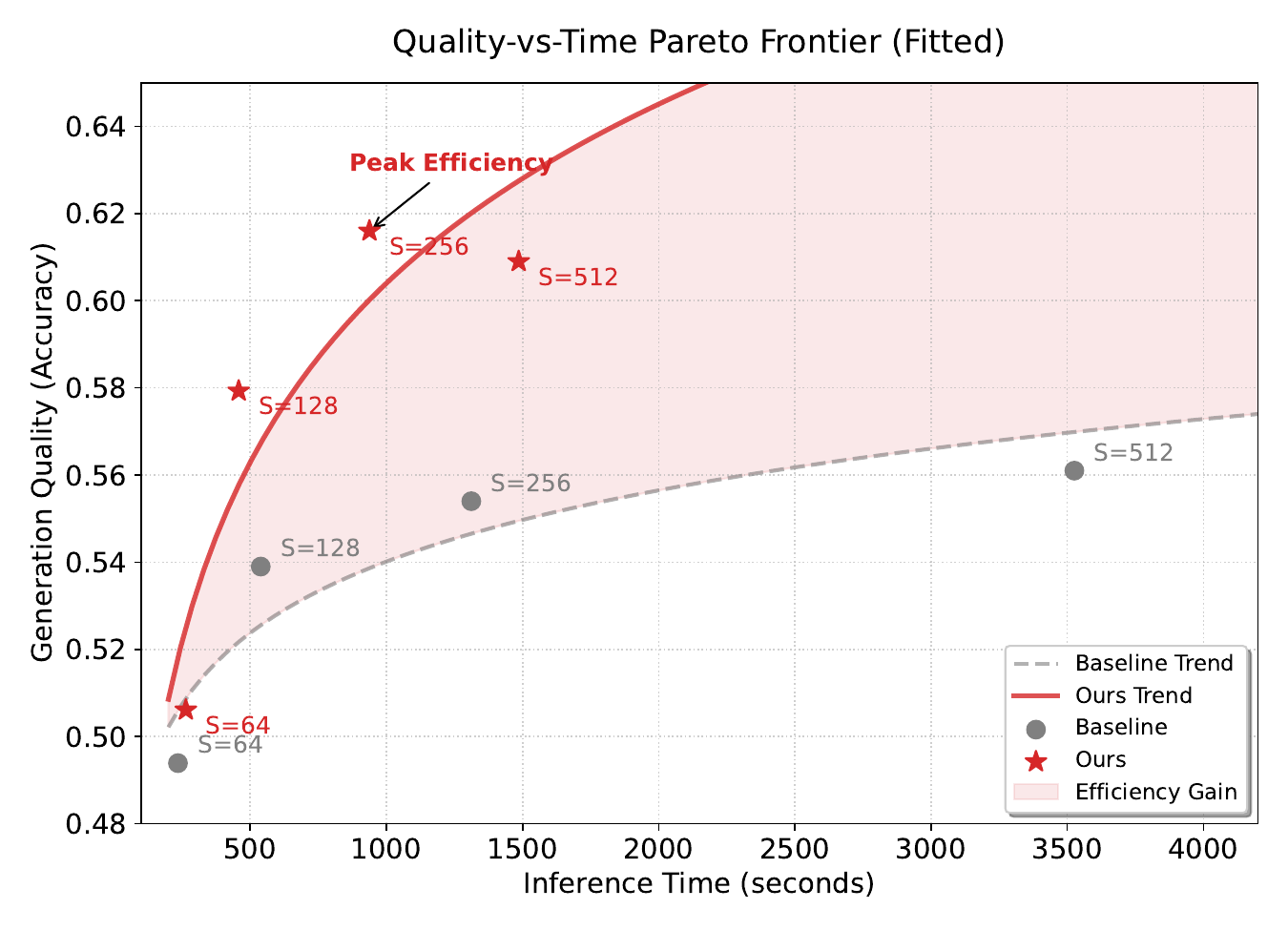}
    \caption{
        Quality vs. time Pareto analysis comparing our method with the baseline on HumanEval using Dream-Instruct under different diffusion-step budgets (2×A100 GPUs).
    }
    \label{fig:Pareto}
\end{figure}

\paragraph{Quality and Time Pareto.}
Beyond scalar speed, we also examine the \emph{quality and time} trade-off by varying the maximum number of diffusion steps $S \in \{64, 128, 256, 512\}$ and plotting accuracy against absolute wall-clock time on HumanEval with Dream-Instruct-7B (Figure~\ref{fig:Pareto}). Each point in the figure corresponds to running either the baseline decoder or LRD with a given step budget $S$, and we fit smooth trend lines to visualise the resulting Pareto frontiers. Across all budgets, LRD lies on a strictly better frontier: for any given time budget it attains higher accuracy, and for any fixed accuracy level it reaches that quality faster.
This shows that the small per-step overhead from KL-based refinement is more than offset by the reduced number of steps, leading to a better overall quality and efficiency trade-off.

\begin{revision}
\FloatBarrier
\subsubsection{Content-Length-Normalised Efficiency}
\label{appendix:content_speed}

Large fixed decoding windows can contain many trailing \texttt{[EOS]} positions. Avoiding their remaining updates is useful end-to-end acceleration, but it does not by itself establish faster generation of reasoning content. We therefore distinguish window-level runtime savings from content-normalised efficiency. Let $E$ be the number of generated tokens before the first \texttt{[EOS]}, and let $t$ be wall-clock generation time. The content-normalised speedup is
\begin{equation}
S_{\mathrm{content}}=\frac{E_{\mathrm{method}}/t_{\mathrm{method}}}{E_{\mathrm{baseline}}/t_{\mathrm{baseline}}}
=\frac{t_{\mathrm{baseline}}}{t_{\mathrm{method}}}\frac{E_{\mathrm{method}}}{E_{\mathrm{baseline}}}.
\end{equation}
Table~\ref{tab:content_speed} reports a matched analysis at $L=1024$. In addition to effective content lengths under ordinary decoding, we report normalised content speed after disabling only Phase~2 sequence-level KL stopping; KL-guided Phase~1 transition and the other components are retained. The length columns describe the ordinary-decoding runs, whereas the no-stop speed columns use the corresponding lengths and times measured in the no-stop runs.

\begin{table}[htbp]
\ifshowchanges\color{blue}\captionsetup{font+={color=blue},labelfont+={color=blue}}\fi
\centering
\caption{Content-length controls at $L=1024$. $E$ counts tokens before \texttt{[EOS]}; $S_{\mathrm{content}}$ is measured with Phase~2 early stopping disabled.}
\label{tab:content_speed}
\small
\resizebox{\linewidth}{!}{%
\begin{tabular}{lccccc}
\toprule
& & \multicolumn{2}{c}{LRD} & \multicolumn{2}{c}{Fast-dLLM+LRD} \\
\cmidrule(lr){3-4}\cmidrule(lr){5-6}
Benchmark & Baseline $E$ & $E$ (change) & $S_{\mathrm{content}}$ (no stop) & $E$ (change) & $S_{\mathrm{content}}$ (no stop) \\
\midrule
HumanEval & 90.4 & 84.6 ($-6.4\%$) & 1.08$\times$ & 98.6 ($+9.1\%$) & 1.11$\times$ \\
MBPP & 60.5 & 57.2 ($-5.5\%$) & 1.06$\times$ & 49.4 ($-18.3\%$) & 1.19$\times$ \\
GSM8K & 135.5 & 123.7 ($-8.7\%$) & 1.07$\times$ & 115.3 ($-14.9\%$) & 1.13$\times$ \\
MATH500 & 482.3 & 437.3 ($-9.3\%$) & 1.09$\times$ & 377.2 ($-21.8\%$) & 1.21$\times$ \\
\bottomrule
\end{tabular}}
\end{table}

The no-stop gains are modest: 1.06--1.09$\times$ for LRD and 1.11--1.21$\times$ for Fast-dLLM+LRD. Thus the large headline speedups arise mainly from sequence-level early stopping under fixed windows. The remaining gains and the higher accuracy support the usefulness of predictive feedback, but do not imply an equally large improvement in reasoning speed. Effective output length can increase as well as decrease: Fast-dLLM+LRD generates 9.1\% more non-EOS tokens on HumanEval in the ordinary-decoding comparison.

\FloatBarrier
\subsubsection{Small Decoding Budgets}
\label{appendix:small_budgets}

Table~\ref{tab:small_budgets} evaluates budgets of 64 and 128 denoising steps. LRD improves aggregate accuracy in all 16 model-budget/task comparisons, but refinement overhead is not always amortised at 64 steps: several settings run at 0.9$\times$ baseline speed. By 128 steps, LRD is faster on all reported tasks. Here the denoising-step budget is varied explicitly; it should not be confused with the generation-window lengths used in the main table.

\begin{table}[htbp]
\ifshowchanges\color{blue}\captionsetup{font+={color=blue},labelfont+={color=blue}}\fi
\centering
\caption{Small-budget decoding. Cells are accuracy / speed relative to the baseline at the same budget.}
\label{tab:small_budgets}
\resizebox{\linewidth}{!}{%
\begin{tabular}{lclcccc}
\toprule
Model & Steps & Method & HumanEval & MBPP & GSM8K & MATH500 \\
\midrule
Dream-Base-7B & 64 & Baseline & 43.2 / 1.0$\times$ & 50.2 / 1.0$\times$ & 33.2 / 1.0$\times$ & 10.9 / 1.0$\times$ \\
& 64 & LRD & 44.4 / 0.9$\times$ & 50.6 / 1.0$\times$ & 33.7 / 1.1$\times$ & 11.6 / 0.9$\times$ \\
& 128 & Baseline & 47.2 / 1.0$\times$ & 54.4 / 1.0$\times$ & 63.4 / 1.0$\times$ & 28.9 / 1.0$\times$ \\
& 128 & LRD & 49.0 / 1.1$\times$ & 55.4 / 1.1$\times$ & 64.6 / 1.3$\times$ & 30.4 / 1.1$\times$ \\
\midrule
LLaDA-Ins-8B & 64 & Baseline & 31.2 / 1.0$\times$ & 36.0 / 1.0$\times$ & 61.0 / 1.0$\times$ & 24.3 / 1.0$\times$ \\
& 64 & LRD & 31.8 / 0.9$\times$ & 36.7 / 0.9$\times$ & 61.5 / 1.0$\times$ & 24.8 / 0.9$\times$ \\
& 128 & Baseline & 35.9 / 1.0$\times$ & 37.3 / 1.0$\times$ & 71.2 / 1.0$\times$ & 30.3 / 1.0$\times$ \\
& 128 & LRD & 36.4 / 1.1$\times$ & 38.1 / 1.1$\times$ & 71.7 / 1.2$\times$ & 31.0 / 1.2$\times$ \\
\bottomrule
\end{tabular}}
\end{table}
\end{revision}

\FloatBarrier
\subsection{Temperature Sampling and pass@\texorpdfstring{$k$}{k}}
\label{appendix:temperature}

Reasoning benchmarks combined with test-time scaling commonly require stochastic decoding. The only implementation difference from the deterministic setting is how temperature is applied when a token is committed; the latent refinement mixture itself remains based on the unscaled model distribution.
Let $\mathbf{z}_t^{(i)}$ denote the logits at position $i$.
LRD first computes the unscaled belief distribution $p_t^{(i)}=\mathrm{softmax}(\mathbf{z}_t^{(i)})$ and uses it for latent refinement exactly as in Eq.~\ref{eq:mix-embedding} and Eq.~\ref{eq:alpha}.
Only when Phase~2 commits a token do we sample from the temperature-scaled distribution
\[
p_{t_\text{temp}}^{(i)}=\mathrm{softmax}(\mathbf{z}_t^{(i)}/T).
\]
Thus temperature affects the stochastic token choice, while the soft mixture remains tied to the model's unscaled belief state and does not amplify sampling noise during refinement.

Table~\ref{tab:passk_temperature} evaluates this setting on LLaDA-Ins-8B ($L=256$, $T=0.8$). LRD improves pass@$k$ for all $k\in\{1,2,5,10\}$ across HumanEval, MBPP, GSM8K, and MATH-500, showing that the method remains compatible with stochastic test-time scaling.

\begin{table}[!htbp]
\centering
\caption{Pass@$k$ under temperature sampling ($T=0.8$) on LLaDA-Ins-8B (L=256). LRD consistently improves pass@$k$ across all benchmarks and all $k$, confirming compatibility with temperature sampling.}
\label{tab:passk_temperature}
\resizebox{0.9\linewidth}{!}{
\begin{tabular}{c|cc|cc|cc|cc}
\toprule
\multirow{2}{*}{$k$} & \multicolumn{2}{c|}{HumanEval} & \multicolumn{2}{c|}{MBPP} & \multicolumn{2}{c|}{GSM8K} & \multicolumn{2}{c}{MATH-500} \\
& Baseline & LRD & Baseline & LRD & Baseline & LRD & Baseline & LRD \\
\midrule
1  & 37.8 & 42.7 & 36.4 & 39.7 & 77.2 & 78.6 & 33.9 & 35.9 \\
2  & 45.7 & 50.1 & 44.1 & 47.1 & 86.5 & 87.7 & 41.5 & 43.4 \\
5  & 52.8 & 56.5 & 52.6 & 54.7 & 92.4 & 93.0 & 50.3 & 52.0 \\
10 & 56.3 & 58.7 & 58.1 & 60.2 & 94.2 & 95.2 & 54.9 & 56.7 \\
\bottomrule
\end{tabular}}
\end{table}

\FloatBarrier
\subsection{Batched Inference}
\label{appendix:batched}

LRD is fully compatible with batched inference. The only complication is that different sequences in a batch may transition between phases or stop early at different steps. This is the same situation that autoregressive models face when sequences emit \texttt{[EOS]} at different times; the standard remedy is a per-sequence \texttt{unfinished\_sequences} flag that pads finished sequences while the batch continues until all sequences complete~\citep{yu2022orca}.

LRD handles its two divergence points with the same mechanism:
\begin{enumerate}
    \item \textbf{Phase 1$\rightarrow$2 transition.} Sequences may reach $\mathcal{D}_\text{KL}^{(t)} < \tau_\text{refine}$ at different steps. We maintain a per-sequence \texttt{phase1\_converged} flag. Both phases execute the identical transformer forward pass (embedding sequence in, logits out). The only difference is post-forward bookkeeping: Phase~1 does not commit tokens, while Phase~2 commits the lowest-entropy position (Eq.~\ref{eq:soft_hard_assignment}). This is per-sequence and does not affect the batched forward pass, so sequences in different phases coexist in the same batch naturally.
    \item \textbf{Early stopping.} Sequences may reach $\mathcal{D}_\text{KL}^{(t)} < \tau_\text{decode}$ at different steps. We maintain a per-sequence \texttt{is\_finished} flag, identical to the post-\texttt{[EOS]} padding strategy used in autoregressive batched generation~\citep{yu2022orca}.
\end{enumerate}
Consequently, LRD adds no additional synchronisation cost over standard batched diffusion decoding: a batch of $B$ sequences continues until either every sequence reaches $\tau_\text{decode}$ or the maximum step budget is exhausted.

\FloatBarrier
\subsection{Case Studies}
\label{appendix:case}
\begin{figure}[!htbp]
    \centering
\includegraphics[width=\linewidth]{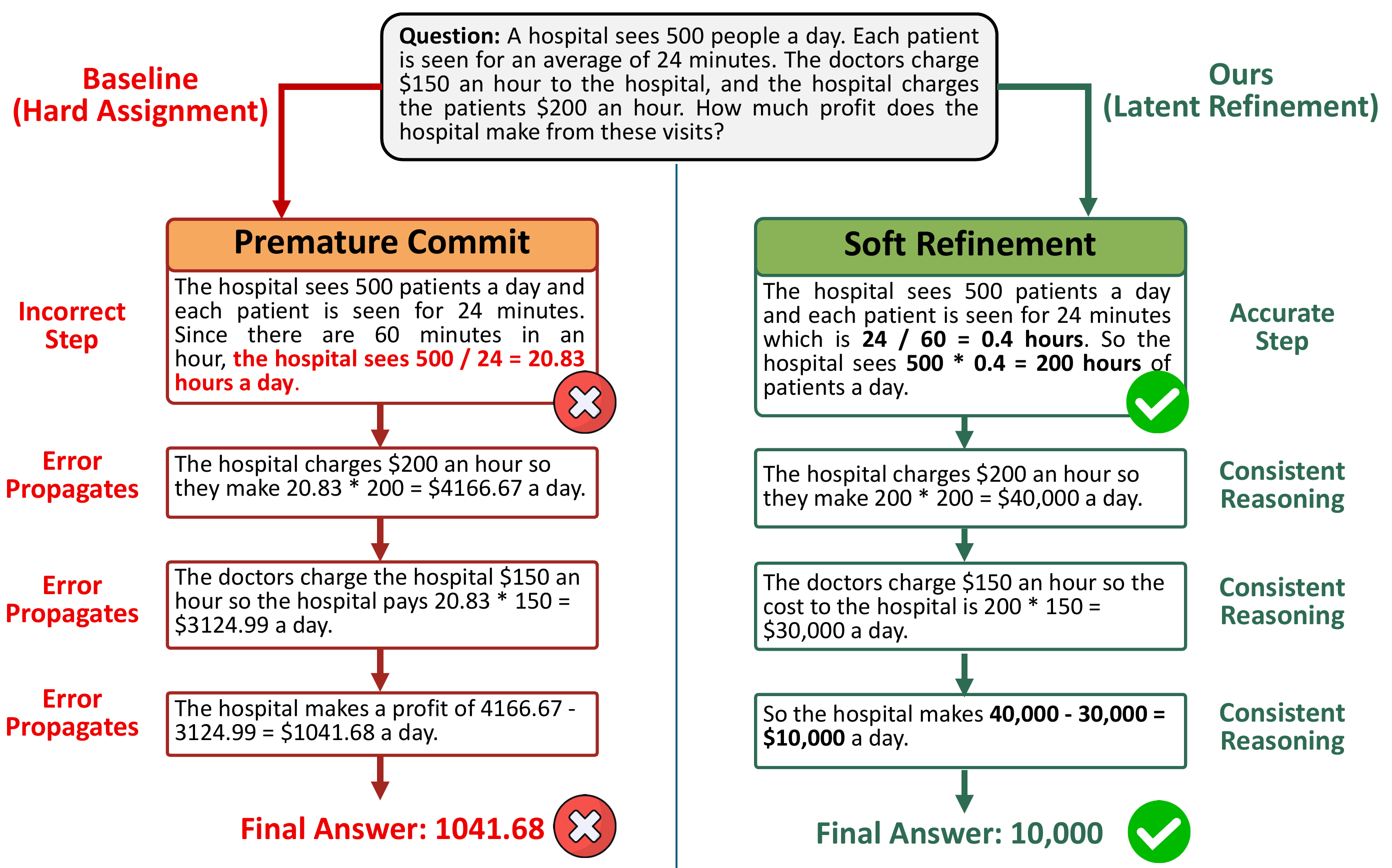}
\caption{Premature Commitment vs. Latent Refinement: A Case Study on Error Propagation (GSM8K)
}
    \label{fig:early_commit}
\end{figure}

\begin{figure}[!htbp]
    \centering
\includegraphics[width=\linewidth]{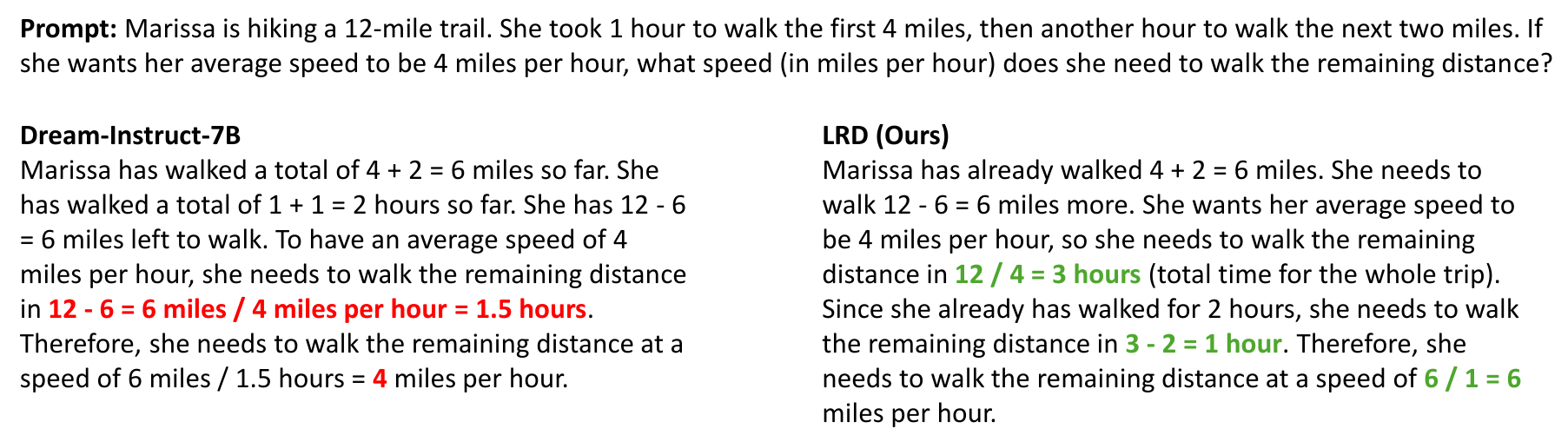}
\caption{Logical Reasoning Error. }
    \label{fig:case1}
\end{figure}

\begin{figure}[!htbp]
    \centering
\includegraphics[width=\linewidth]{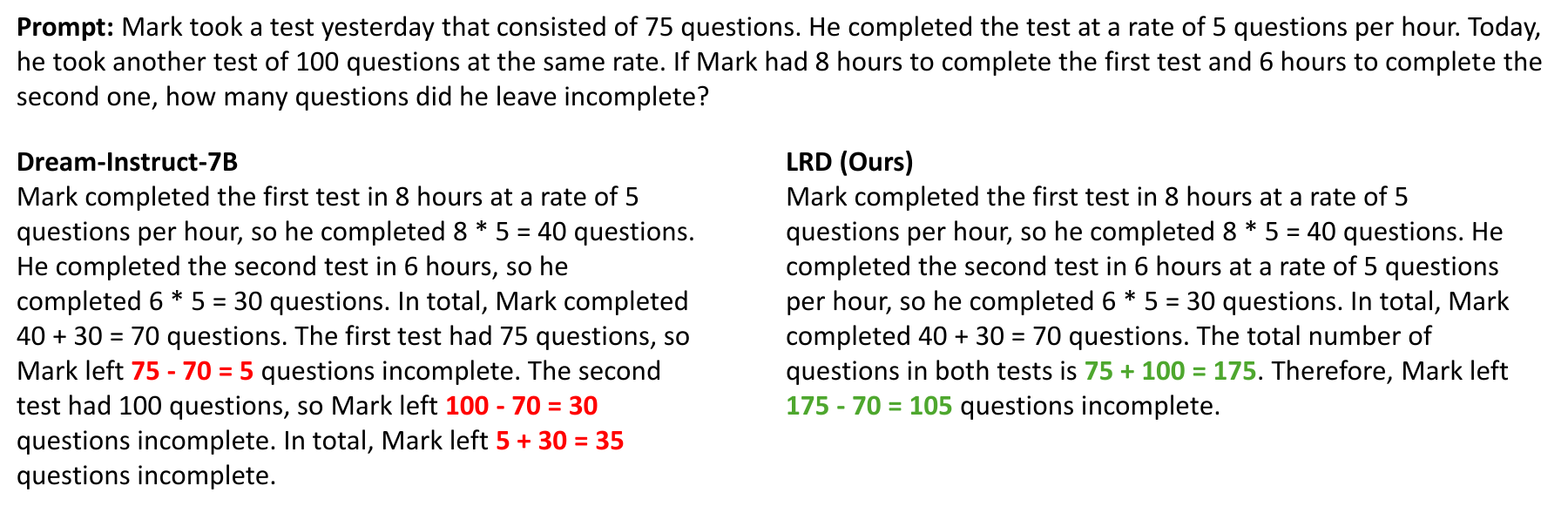}
\caption{Procedural Step Errors. }
    \label{fig:case2}
\end{figure}

\begin{figure}[!htbp]
    \centering
\includegraphics[width=\linewidth]{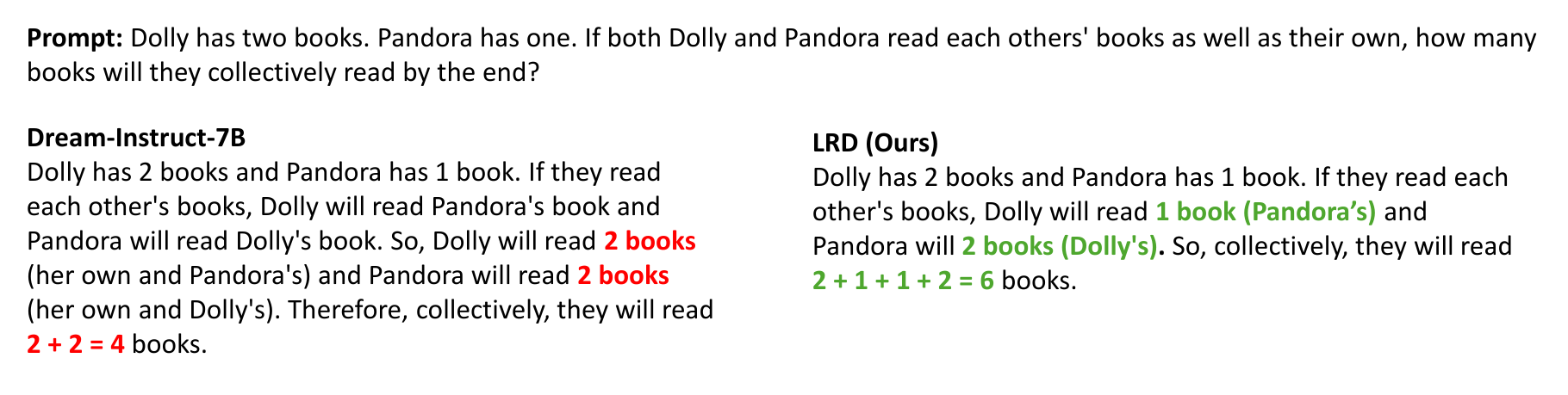}
\caption{Problem Misinterpretation. }
    \label{fig:case3}
\end{figure}

To provide a deeper understanding of the mechanisms underlying model performance, we conduct representative failure cases encountered by the baseline model and contrast them with LRD solutions. This case study elucidates the specific types of reasoning breakdowns that persist in standard models when facing chain-of-thought reasoning problems. We categorise these failures into three distinct types: (1) Logical Reasoning Errors, where correct premises lead to invalid intermediate deductions; (2) Procedural Step Errors, involving omitted or extraneous calculations during state tracking; and (3) Problem Misinterpretation, where the initial mapping from natural language constraints to formal representations is flawed. Figure~\ref{fig:early_commit} shows a GSM8K premature-commitment case, and Figures~\ref{fig:case1}--\ref{fig:case3} illustrate the three error types.

\begin{revision}
\paragraph{Representative corrections and regressions.}
In HumanEval/8 (\texttt{sum\_product}), hard decoding sets \texttt{product = 1} but returns inside the loop without multiplying by the input elements; LRD updates the product for every element and returns after the loop. In a GSM8K interrupted-download problem, hard decoding treats the remaining 120\,GB as the post-update workload and omits the already elapsed 40 minutes, while LRD accounts for the restart and all three intervals ($40+20+100=160$ minutes). Conversely, some LRD-specific regressions alter a locally correct symbol (e.g., changing 68 to 88) or settle on an inconsistent calculation. Improvements are aggregate, rather than universal across individual examples.
\end{revision}

\FloatBarrier
\section{Ablations and Hyperparameters}
\label{appendix:ablations_full}

\FloatBarrier
\subsection{Component Ablations}
\label{appendix:unified_ablation_results}

\begin{table*}[!htbp]
\caption{Component ablation results on decoding variants. We report Accuracy, Speed, and effective token number $E_{\text{token}}$. \textcolor{red}{Red} and \textcolor{LightGreen}{green} numbers indicate changes compared to our full method.}
\label{tab:ablation_all}
\vspace{-0.2em}
\centering
\renewcommand{\arraystretch}{1.0}
\resizebox{\linewidth}{!}{
\begin{tabular}{c l *{12}{l}}
\toprule[1pt]
\multirow{2}{*}{\textbf{Length}} & \multirow{2}{*}{\textbf{Method}}
& \multicolumn{4}{c}{\textbf{Accuracy}} & \multicolumn{4}{c}{\textbf{Speed}} & \multicolumn{4}{c}{$\mathbf{E_{\text{token}}}$} \\
\cmidrule(lr){3-6} \cmidrule(lr){7-10} \cmidrule(lr){11-14}
& & HumanEval & MBPP & GSM8K & MATH500
  & HumanEval & MBPP & GSM8K & MATH500
  & HumanEval & MBPP & GSM8K & MATH500 \\
\midrule

\multirow{5}{*}{256}
& baseline
& \textcolor{DarkGrey}{55.4} & \textcolor{DarkGrey}{57.4} & \textcolor{DarkGrey}{80.8} & \textcolor{DarkGrey}{37.9}
& \textcolor{DarkGrey}{1.0$\times$} & \textcolor{DarkGrey}{1.0$\times$} & \textcolor{DarkGrey}{1.0$\times$} & \textcolor{DarkGrey}{1.0$\times$}
& \textcolor{DarkGrey}{117.2} & \textcolor{DarkGrey}{53.5} & \textcolor{DarkGrey}{132.4} & \textcolor{DarkGrey}{228.4} \\
& Ours
& 61.6$_{+6.2}$ & 59.4$_{+2.0}$ & 83.0$_{+2.2}$ & 40.6$_{+2.7}$
& 1.4$\times$$_{+0.4\times}$ & 2.4$\times$$_{+1.4\times}$ & 1.4$\times$$_{+0.4\times}$ & 1.1$\times$$_{+0.1\times}$
& 108.4$_{-8.8}$ & 49.2$_{-4.3}$ & 128.6$_{-5.8}$ & 226.0$_{-2.4}$ \\
\cmidrule(lr){2-14}
& w/o latent refinement
& 60.1$_{\textcolor{red}{-1.5}}$ & 58.6$_{\textcolor{red}{-0.8}}$ & 82.3$_{\textcolor{red}{-0.7}}$ & 39.4$_{\textcolor{red}{-1.2}}$
& 1.5$\times$$_{\textcolor{LightGreen}{+0.1\times}}$ & 2.5$\times$$_{\textcolor{LightGreen}{+0.1\times}}$ & 1.5$\times$$_{\textcolor{LightGreen}{+0.1\times}}$ & 1.1$\times$$_{\textcolor{LightGreen}{+0.0\times}}$
& 108.7$_{\textcolor{red}{+0.3}}$ & 50.4$_{\textcolor{red}{+1.2}}$ & 129.9$_{\textcolor{red}{+1.3}}$ & 226.8$_{\textcolor{red}{+0.8}}$ \\
& w/o mix embed
& 59.5$_{\textcolor{red}{-2.1}}$ & 58.8$_{\textcolor{red}{-0.6}}$ & 82.7$_{\textcolor{red}{-0.3}}$ & 38.9$_{\textcolor{red}{-1.7}}$
& 1.3$\times$$_{\textcolor{red}{-0.1\times}}$ & 2.2$\times$$_{\textcolor{red}{-0.2\times}}$ & 1.5$\times$$_{\textcolor{LightGreen}{+0.1\times}}$ & 1.0$\times$$_{\textcolor{red}{-0.1\times}}$
& 117.2$_{\textcolor{red}{+8.8}}$ & 49.5$_{\textcolor{red}{+0.3}}$ & 129.4$_{\textcolor{red}{+0.8}}$ & 228.4$_{\textcolor{red}{+2.4}}$ \\
& w/o early stop
& 61.8$_{\textcolor{LightGreen}{+0.2}}$ & 59.4$_{\textcolor{LightGreen}{+0.0}}$ & 83.2$_{\textcolor{LightGreen}{+0.2}}$ & 40.6$_{\textcolor{LightGreen}{+0.0}}$
& 1.0$\times$$_{\textcolor{red}{-0.4\times}}$ & 1.1$\times$$_{\textcolor{red}{-1.3\times}}$ & 1.1$\times$$_{\textcolor{red}{-0.3\times}}$ & 1.0$\times$$_{\textcolor{red}{-0.1\times}}$
& 109.7$_{\textcolor{red}{+1.3}}$ & 51.4$_{\textcolor{red}{+2.2}}$ & 129.9$_{\textcolor{red}{+1.3}}$ & 228.0$_{\textcolor{red}{+2.0}}$ \\
\midrule
\multirow{5}{*}{512}
& baseline
& \textcolor{DarkGrey}{56.1} & \textcolor{DarkGrey}{56.7} & \textcolor{DarkGrey}{80.2} & \textcolor{DarkGrey}{38.6}
& \textcolor{DarkGrey}{1.0$\times$} & \textcolor{DarkGrey}{1.0$\times$} & \textcolor{DarkGrey}{1.0$\times$} & \textcolor{DarkGrey}{1.0$\times$}
& \textcolor{DarkGrey}{116.2} & \textcolor{DarkGrey}{55.7} & \textcolor{DarkGrey}{135.2} & \textcolor{DarkGrey}{378.9} \\
& Ours
& 60.9$_{+4.8}$ & 58.8$_{+2.1}$ & 82.7$_{+2.5}$ & 41.8$_{+3.2}$
& 2.9$\times$$_{+1.9\times}$ & 4.6$\times$$_{+3.6\times}$ & 3.6$\times$$_{+2.6\times}$ & 1.2$\times$$_{+0.2\times}$
& 103.9$_{-12.3}$ & 51.8$_{-4.1}$ & 125.9$_{-9.3}$ & 363.5$_{-15.4}$ \\
\cmidrule(lr){2-14}
& w/o latent refinement
& 59.9$_{\textcolor{red}{-1.0}}$ & 57.8$_{\textcolor{red}{-1.0}}$ & 82.2$_{\textcolor{red}{-0.5}}$ & 41.0$_{\textcolor{red}{-0.8}}$
& 3.1$\times$$_{\textcolor{LightGreen}{+0.2\times}}$ & 4.9$\times$$_{\textcolor{LightGreen}{+0.3\times}}$ & 3.8$\times$$_{\textcolor{LightGreen}{+0.2\times}}$ & 1.2$\times$$_{\textcolor{LightGreen}{+0.0\times}}$
& 106.3$_{\textcolor{red}{+2.4}}$ & 52.6$_{\textcolor{red}{+0.8}}$ & 127.9$_{\textcolor{red}{+2.0}}$ & 363.0$_{\textcolor{LightGreen}{-0.5}}$ \\
& w/o mix embed
& 58.0$_{\textcolor{red}{-2.9}}$ & 57.8$_{\textcolor{red}{-1.0}}$ & 80.7$_{\textcolor{red}{-2.0}}$ & 40.8$_{\textcolor{red}{-1.0}}$
& 2.7$\times$$_{\textcolor{red}{-0.2\times}}$ & 4.3$\times$$_{\textcolor{red}{-0.3\times}}$ & 3.0$\times$$_{\textcolor{red}{-0.6\times}}$ & 1.0$\times$$_{\textcolor{red}{-0.2\times}}$
& 116.2$_{\textcolor{red}{+12.3}}$ & 51.8$_{\textcolor{red}{+0.0}}$ & 126.2$_{\textcolor{red}{+0.3}}$ & 368.9$_{\textcolor{red}{+5.4}}$ \\
& w/o early stop
& 61.2$_{\textcolor{LightGreen}{+0.3}}$ & 58.8$_{\textcolor{LightGreen}{+0.0}}$ & 82.9$_{\textcolor{LightGreen}{+0.2}}$ & 41.9$_{\textcolor{LightGreen}{+0.1}}$
& 1.2$\times$$_{\textcolor{red}{-1.7\times}}$ & 1.1$\times$$_{\textcolor{red}{-3.5\times}}$ & 1.1$\times$$_{\textcolor{red}{-2.5\times}}$ & 1.0$\times$$_{\textcolor{red}{-0.2\times}}$
& 106.2$_{\textcolor{red}{+2.3}}$ & 53.6$_{\textcolor{red}{+1.8}}$ & 127.2$_{\textcolor{red}{+1.3}}$ & 366.0$_{\textcolor{red}{+2.5}}$ \\

\midrule
\multirow{5}{*}{1024}
& baseline
& \textcolor{DarkGrey}{56.0} & \textcolor{DarkGrey}{57.3} & \textcolor{DarkGrey}{81.3} & \textcolor{DarkGrey}{40.1}
& \textcolor{DarkGrey}{1.0$\times$} & \textcolor{DarkGrey}{1.0$\times$} & \textcolor{DarkGrey}{1.0$\times$} & \textcolor{DarkGrey}{1.0$\times$}
& \textcolor{DarkGrey}{90.4} & \textcolor{DarkGrey}{60.5} & \textcolor{DarkGrey}{135.5} & \textcolor{DarkGrey}{482.3} \\
& Ours
& 61.0$_{+5.0}$ & 59.0$_{+1.7}$ & 83.5$_{+2.2}$ & 43.9$_{+3.8}$
& 9.3$\times$$_{+8.3\times}$ & 10.6$\times$$_{+9.6\times}$ & 5.5$\times$$_{+4.5\times}$ & 1.7$\times$$_{+0.7\times}$
& 84.6$_{-5.8}$ & 57.2$_{-3.3}$ & 123.7$_{-11.8}$ & 437.3$_{-45.0}$ \\
\cmidrule(lr){2-14}
& w/o latent refinement
& 60.7$_{\textcolor{red}{-0.3}}$ & 58.8$_{\textcolor{red}{-0.2}}$ & 83.2$_{\textcolor{red}{-0.3}}$ & 42.4$_{\textcolor{red}{-1.5}}$
& 9.3$\times$$_{\textcolor{LightGreen}{+0.0\times}}$ & 10.7$\times$$_{\textcolor{LightGreen}{+0.1\times}}$ & 5.6$\times$$_{\textcolor{LightGreen}{+0.1\times}}$ & 1.7$\times$$_{\textcolor{LightGreen}{+0.0\times}}$
& 83.9$_{\textcolor{LightGreen}{-0.7}}$ & 58.2$_{\textcolor{red}{+1.0}}$ & 125.0$_{\textcolor{red}{+1.3}}$ & 455.4$_{\textcolor{red}{+18.1}}$ \\
& w/o mix embed
& 59.1$_{\textcolor{red}{-1.9}}$ & 58.7$_{\textcolor{red}{-0.3}}$ & 82.9$_{\textcolor{red}{-0.6}}$ & 41.4$_{\textcolor{red}{-2.5}}$
& 9.1$\times$$_{\textcolor{red}{-0.2\times}}$ & 10.4$\times$$_{\textcolor{red}{-0.2\times}}$ & 5.1$\times$$_{\textcolor{red}{-0.4\times}}$ & 1.3$\times$$_{\textcolor{red}{-0.4\times}}$
& 90.4$_{\textcolor{red}{+5.8}}$ & 61.7$_{\textcolor{red}{+4.5}}$ & 130.5$_{\textcolor{red}{+6.8}}$ & 483.5$_{\textcolor{red}{+46.2}}$ \\
& w/o early stop
& 61.4$_{\textcolor{LightGreen}{+0.4}}$ & 59.0$_{\textcolor{LightGreen}{+0.0}}$ & 83.7$_{\textcolor{LightGreen}{+0.2}}$ & 44.2$_{\textcolor{LightGreen}{+0.3}}$
& 1.2$\times$$_{\textcolor{red}{-8.1\times}}$ & 1.1$\times$$_{\textcolor{red}{-9.5\times}}$ & 1.2$\times$$_{\textcolor{red}{-4.3\times}}$ & 1.2$\times$$_{\textcolor{red}{-0.5\times}}$
& 86.2$_{\textcolor{red}{+1.6}}$ & 59.2$_{\textcolor{red}{+2.0}}$ & 126.1$_{\textcolor{red}{+2.4}}$ & 438.9$_{\textcolor{red}{+1.6}}$ \\

\bottomrule[1pt]
\end{tabular}
}
\end{table*}

First, regarding accuracy, removing either latent refinement or mixed embeddings consistently reduces performance.
The absence of mixed embeddings causes the most significant drops (up to $-2.9$ on HumanEval and $-2.5$ on MATH500), demonstrating that the predictive feedback loop is critical for maintaining generation quality.
Second, regarding efficiency and convergence, removing mixed embeddings makes decoding slower, indicating its vital role in helping the model converge earlier.
Conversely, removing latent refinement yields slightly faster decoding, showing that while it improves convergence and stability, the refinement steps incur a minor speed overhead.
Finally, early stopping proves to be the primary driver of speedup; removing it \rev{reduces speedups to roughly $1.0$--$1.2\times$}, with virtually no cost to accuracy.
Overall, latent refinement and mixed embeddings are essential for high-performance convergence, whereas early stopping unlocks massive efficiency gains.

\begin{revision}
\FloatBarrier
\subsubsection{Matched-State Probe of Information Preservation}
\label{appendix:matched_state}

To isolate the effect of feedback from differences in decoding history, we reconstruct each baseline pre-commit state and freeze its commitment set and committed tokens. At the unresolved positions, \emph{Hard} uses the usual \texttt{[MASK]} embeddings, whereas \emph{Soft} replays LRD's recurrent soft-embedding updates along the same baseline trajectory up to that state. We then compare one forward pass from the matched state. Both conditions therefore share the same discrete history and commitments; only the treatment of unresolved positions differs.

The experiment uses Dream-Instruct-7B at generation length 256, with 164 HumanEval and 1,319 GSM8K evaluation examples. Claude Opus 5 was used to localise where each baseline response first became incorrect; the mapping to the corresponding token in the task-verified correct answer was manually checked. At an unresolved position whose shared-state prediction is wrong, we measure the full-vocabulary softmax probability and rank of the correct token $y^*$, and
\begin{equation}
\Delta p=p_{\mathrm{soft}}(y^*)-p_{\mathrm{hard}}(y^*).
\end{equation}
Table~\ref{tab:matched_state} separates baseline-wrong examples according to whether the final LRD response is correct or wrong. Confidence intervals use an example-clustered bootstrap, so multiple positions from one example are resampled together.

\begin{table}[htbp]
\ifshowchanges\color{blue}\captionsetup{font+={color=blue},labelfont+={color=blue}}\fi
\centering
\caption{Matched-state hard/soft probe. Counts are error positions, not independent examples. Intervals are 95\% example-clustered bootstrap confidence intervals; lower rank is better.}
\label{tab:matched_state}
\resizebox{\linewidth}{!}{%
\begin{tabular}{llccccc}
\toprule
Final outcome & Benchmark & Positions & Hard $p(y^*)$ & Soft $p(y^*)$ & $\Delta p$ [95\% CI] & Rank: Hard $\to$ Soft \\
\midrule
Baseline wrong / LRD correct & HumanEval & 73 & 0.24 & 0.37 & $+0.13$ [0.08, 0.18] & $6.9\to4.2$ \\
Baseline wrong / LRD correct & GSM8K & 186 & 0.11 & 0.23 & $+0.12$ [0.08, 0.16] & $46.2\to12.8$ \\
Baseline wrong / LRD wrong & HumanEval & 108 & 0.17 & 0.21 & $+0.04$ [0.01, 0.07] & $11.6\to9.5$ \\
Baseline wrong / LRD wrong & GSM8K & 327 & 0.09 & 0.15 & $+0.06$ [0.03, 0.09] & $51.3\to43.7$ \\
\bottomrule
\end{tabular}}
\end{table}

Soft feedback increases the correct-token probability and improves its rank, with larger effects on examples LRD ultimately corrects. The positive effect when both final responses are wrong separates information preservation from final task success: retaining a useful alternative need not recover the complete trajectory. Rank improvement also indicates that the change is more than a uniform flattening of the output distribution. These matched-state results support the proposed mechanism on the probed error positions, without establishing a pointwise guarantee for every token or example.
\end{revision}

\FloatBarrier
\subsection{Repeated Latent Refinement}
\label{appendix:repeated_refinement}

\begin{table}[!htbp]
\caption{Ablation study on decoding variants at length 512, where \textit{Auto} uses adaptive latent refinement, and \textit{LF$\times$k} enforces $k$ latent refinement steps prior to each token commitment.}
\label{tab:ablation_result}
\centering
\small
\resizebox{0.7\linewidth}{!}{
\begin{tabular}{lllll}
\toprule[1pt]
\textbf{Method} & \textbf{HumanEval} & \textbf{MBPP} & \textbf{GSM8K} & \textbf{MATH500} \\
\midrule
Baseline & \textcolor{DarkGrey}{56.1} & \textcolor{DarkGrey}{56.7} & \textcolor{DarkGrey}{80.2} & \textcolor{DarkGrey}{38.6} \\
Ours     & 60.9$_{\textcolor{LightGreen}{+4.8}}$ & 58.8$_{\textcolor{LightGreen}{+2.1}}$ & 82.7$_{\textcolor{LightGreen}{+2.5}}$ & 41.8$_{\textcolor{LightGreen}{+3.2}}$ \\
Auto     & 59.6$_{\textcolor{LightGreen}{+3.5}}$ & 57.7$_{\textcolor{LightGreen}{+1.0}}$ & 81.5$_{\textcolor{LightGreen}{+1.3}}$ & 41.4$_{\textcolor{LightGreen}{+2.8}}$ \\
LF$\times$1     & 60.4$_{\textcolor{LightGreen}{+4.3}}$ & 57.8$_{\textcolor{LightGreen}{+1.1}}$ & 81.6$_{\textcolor{LightGreen}{+1.4}}$ & 40.2$_{\textcolor{LightGreen}{+1.6}}$ \\
LF$\times$2     & 58.3$_{\textcolor{LightGreen}{+2.2}}$ & 57.2$_{\textcolor{LightGreen}{+0.5}}$ & 81.2$_{\textcolor{LightGreen}{+1.0}}$ & 40.2$_{\textcolor{LightGreen}{+1.6}}$ \\
LF$\times$3     & 57.9$_{\textcolor{LightGreen}{+1.8}}$ & 57.8$_{\textcolor{LightGreen}{+1.1}}$ & 81.8$_{\textcolor{LightGreen}{+1.6}}$ & 39.0$_{\textcolor{LightGreen}{+0.4}}$ \\
LF$\times$4     & 60.8$_{\textcolor{LightGreen}{+4.7}}$ & 57.2$_{\textcolor{LightGreen}{+0.5}}$ & 80.9$_{\textcolor{LightGreen}{+0.7}}$ & 39.6$_{\textcolor{LightGreen}{+1.0}}$ \\
LF$\times$5     & 58.8$_{\textcolor{LightGreen}{+2.7}}$ & 57.6$_{\textcolor{LightGreen}{+0.9}}$ & 80.7$_{\textcolor{LightGreen}{+0.5}}$ & 39.2$_{\textcolor{LightGreen}{+0.6}}$ \\
\bottomrule[1pt]
\end{tabular}}
\end{table}

\textbf{Excessive latent refinement brings no benefit but slows decoding.}
Table~\ref{tab:ablation_result} compares our two-stage strategy (one initial latent refinement followed by standard decoding) with variants that enforce latent refinement at every step, either a fixed number of times (LF$\times k$) or adaptively (Auto). Results show that while all variants outperform the baseline, none surpass our method: enforcing repeated latent refinements from LF$\times2$ to LF$\times5$ generally degrades accuracy, and even adaptive scheduling (Auto) underperforms compared to ours. The reason is that excessive latent refinement adds redundant computation without providing additional guidance once the model has stabilised. In contrast, our two-stage design strikes a better balance by leveraging latent refinement only at the beginning, yielding both higher accuracy and substantially faster decoding.

\FloatBarrier
\subsection{Convergence Analysis}
\label{appendix:convergence_analysis}

\subsubsection{Latent Refinement Convergence Ratio}
\label{appendix:latent_refinement_ratio}

\begin{figure}[!htbp]
\centering
\includegraphics[width=0.6\linewidth]{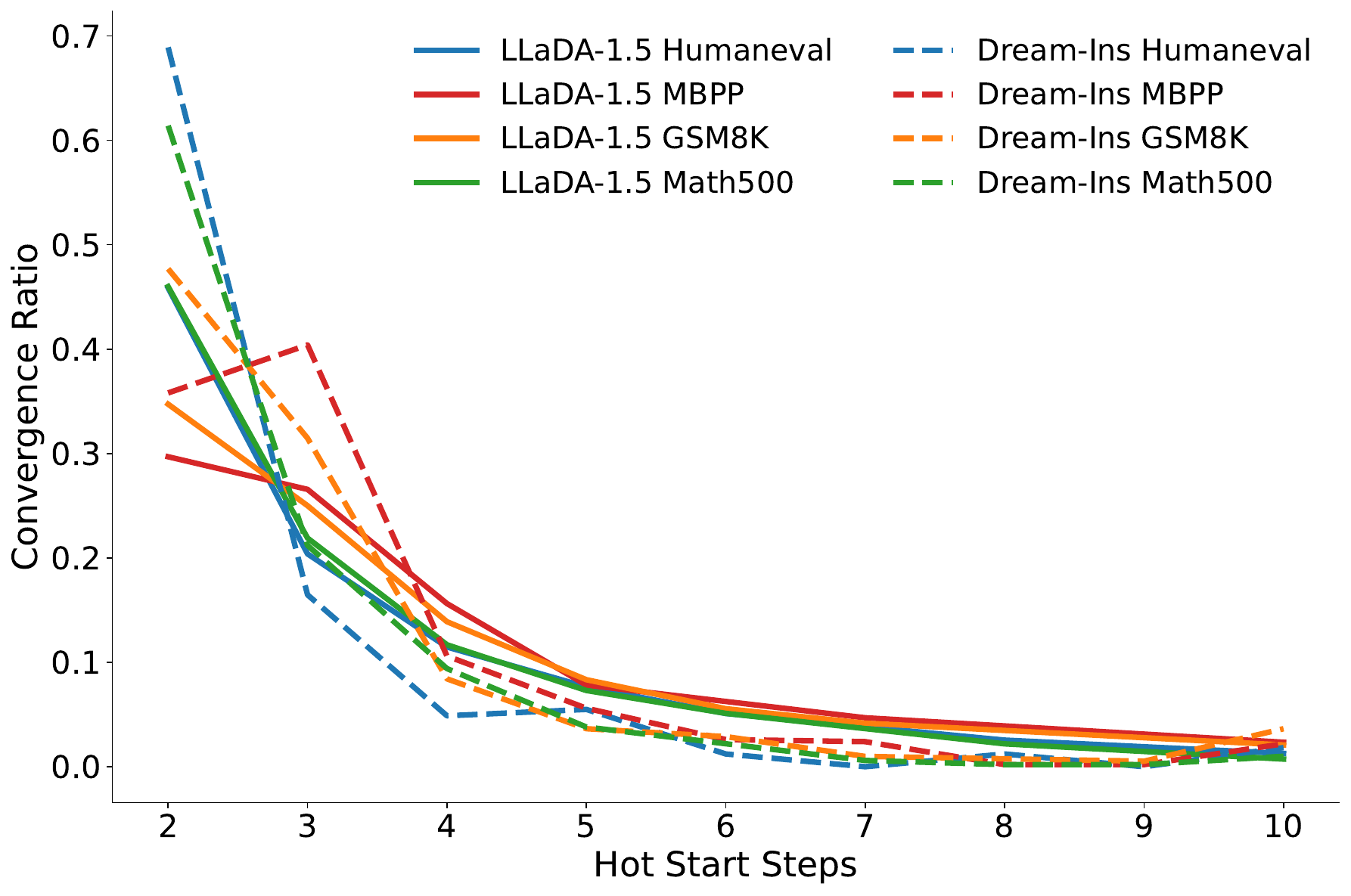}
\caption{Convergence ratios across latent refinement steps for LLaDA-1.5 and Dream-Ins on four benchmarks. Since computing the difference in KL divergence requires at least three consecutive steps, the curves are plotted starting from step 2.}
\label{fig:hotstart}
\end{figure}

Figure~\ref{fig:hotstart} reports the proportion of cases converging at each latent refinement step. Across benchmarks, most runs converge within the first few refinement steps. On HumanEval with Dream-Ins, 68.9\% of samples converge by step 2, and more than 85\% by step 3. Similar trends hold for GSM8K, MBPP, and MATH500, where over 70\% of cases converge within the first three to four steps. These results confirm that latent refinement is highly efficient in practice: most examples stabilise early, reducing the need for excessive refinement iterations and validating the design of our latent refinement mechanism.

\FloatBarrier
\subsubsection{Choice of Convergence Metric}
\label{appendix:convergence_metric}

\begin{figure}[!htbp]
    \centering
    \includegraphics[width=0.6\textwidth]{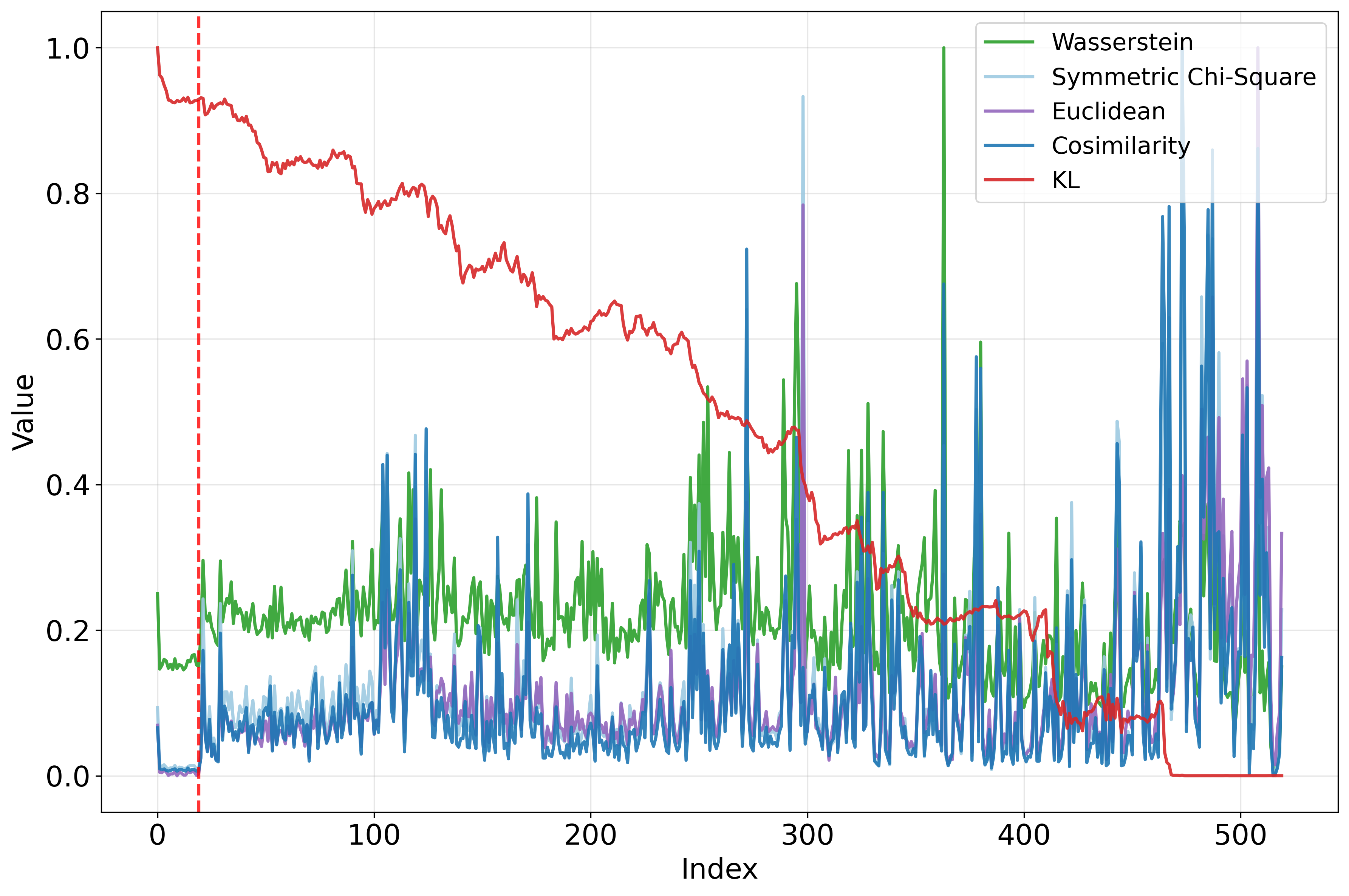}
    \caption{
        Step-wise predictive distributions of different distance metrics on LLaDA-1.5 (HumanEval).
        The red vertical line indicates where decoding begins after a fixed 20-step latent refinement.
    }
    \label{fig:Convergence_Metric}
\end{figure}

We compare several distance metrics for monitoring convergence of the latent belief state on LLaDA-1.5 with HumanEval, as shown in Figure~\ref{fig:Convergence_Metric}. Specifically, we track step-wise distances between consecutive predictive distributions using KL divergence, Wasserstein distance, Chi-Square distance, Euclidean distance, and cosine similarity, with a 20-step latent refinement phase followed by decoding (the \textcolor{red}{red} vertical line marks the transition point).

KL divergence exhibits a clear and stable decreasing trend throughout both the refinement and decoding phases, providing a smooth signal that correlates well with the model’s convergence behaviour. In contrast, the alternative metrics show much higher variance and noise. Based on this comparison, we adopt the KL divergence as our primary convergence indicator for triggering phase transitions and early stopping in LRD.

\begin{revision}
\paragraph{Relation to KLASS.}
Both KLASS~\citep{kim2025klass} and LRD use directed token-distribution KL as a stability signal. KLASS combines per-token KL stability over a history window with confidence to select positions for unmasking in a hard-token/mask state. LRD averages consecutive-step KL over the sequence and uses it to trigger the initial soft-to-hard phase transition and later sequence-level stopping. Thus the distinction is the state and action controlled by KL, and its combination with soft refinement, rather than a new divergence or sequence averaging alone. A small KL residual measures predictive stability, not answer correctness; the metric comparisons in Appendix~\ref{appendix:convergence_metric} provide an empirical choice of monitor.
\end{revision}

\FloatBarrier
\subsection{Hyperparameter Analysis}
\label{appendix:hyperparameter_analysis}

\subsubsection{Top-\texorpdfstring{$p$}{p} Mixing}
\label{appendix:topp_mixing}

\begin{figure}[!htbp]
\centering
\includegraphics[width=0.6\linewidth]{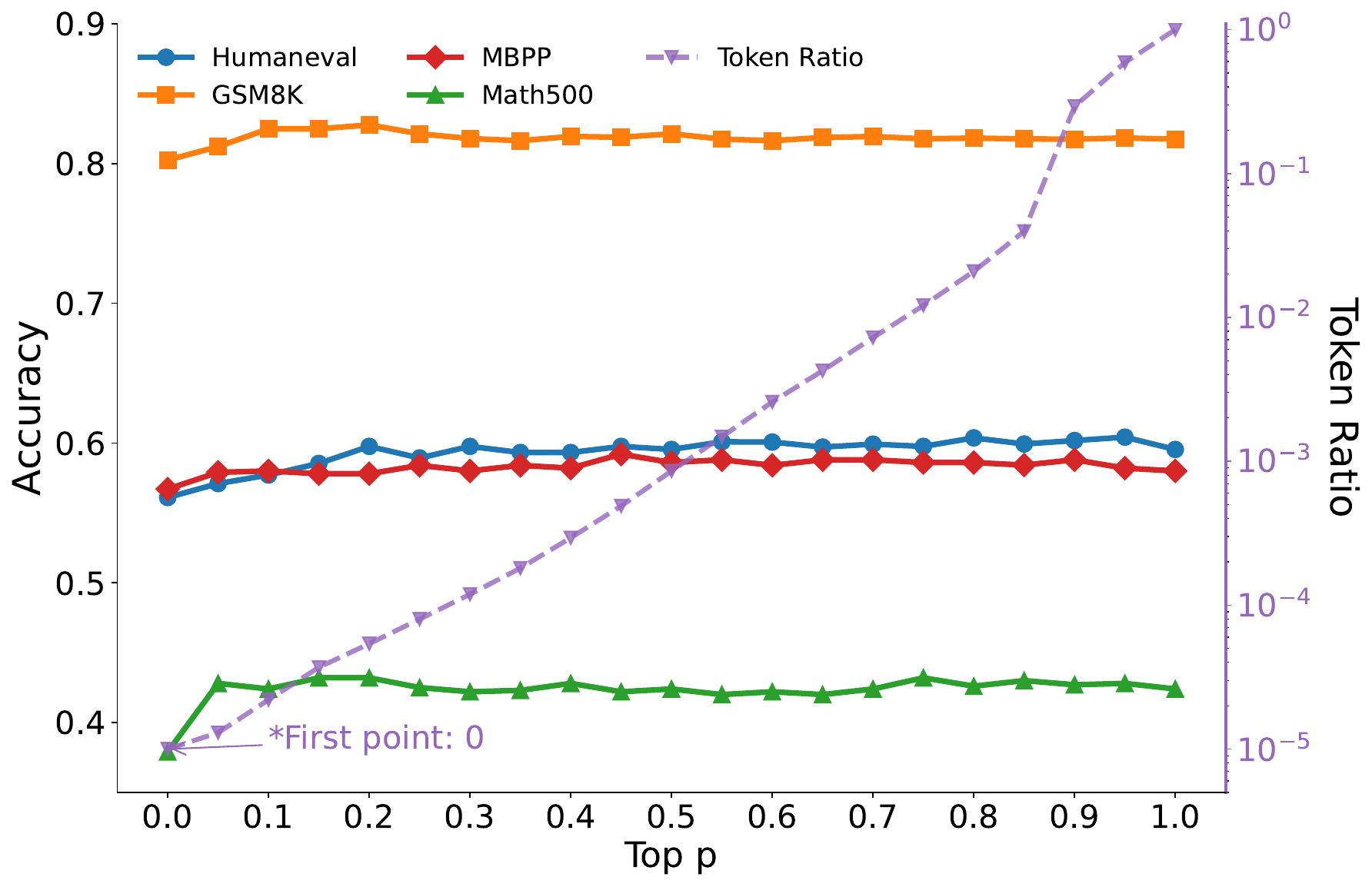}
\caption{Effect of top-$p$ mixing on Dream-Ins across four benchmarks. The purple curve shows the log fraction of tokens included in the mixture.}
\label{fig:topp}
\end{figure}

Figure~\ref{fig:topp} shows that when $p=0$ no mixing occurs and the method degenerates to the baseline, giving the lowest accuracy across all benchmarks. Increasing $p$ quickly improves performance, even though the token ratio curve indicates that only a very small fraction of tokens are mixed at $p\leq0.2$. This suggests that the key factor is enabling mixing rather than the absolute number of tokens included. Beyond $p\approx0.2$, accuracy stabilises and fluctuates slightly, showing that adding more low-probability tokens offers little benefit while introducing potential noise. These results confirm that top-$p$ mixing provides a good balance: minimal mixing is already highly effective, and larger $p$ values do not bring further gains.

\FloatBarrier
\subsubsection{Hyperparameter Robustness}
\label{appendix:hyperparam}

We sweep each of the three LRD hyperparameters $\tau_{\text{refine}}$, $\tau_{\text{decode}}$, and $T_{\text{refine}}$ individually while keeping the others at the defaults $(\tau_{\text{refine}}{=}0.1,\,\tau_{\text{decode}}{=}0.1,\,T_{\text{refine}}{=}20)$ on Dream-Instruct-7B with $L=512$. Table~\ref{tab:hyperparam} reports per-benchmark accuracy on HumanEval, MBPP, GSM8K, and MATH-500, together with the average speed across the four benchmarks.

\begin{table}[!htbp]
\caption{Single-axis hyperparameter sweep on Dream-Instruct-7B ($L=512$). For each row only the listed hyperparameter changes; the others remain at their default. ``Avg Speed'' is the geometric mean of the per-benchmark speedups over the four benchmarks. Default: $\tau_{\text{refine}}=0.1$, $\tau_{\text{decode}}=0.1$, $T_{\text{refine}}=20$.}
\label{tab:hyperparam}
\centering
\small
\resizebox{0.9\linewidth}{!}{
\begin{tabular}{llccccc}
\toprule
Hyperparam & Value & HumanEval & MBPP & GSM8K & MATH-500 & Avg Speed \\
\midrule
$\tau_{\text{refine}}$ & 0.05            & 60.8 & 58.8 & 82.4 & 41.9 & 2.6$\times$ \\
$\tau_{\text{refine}}$ & 0.10 (default)  & 60.9 & 58.8 & 82.7 & 41.8 & 2.8$\times$ \\
$\tau_{\text{refine}}$ & 0.15            & 60.8 & 58.6 & 82.6 & 41.6 & 2.9$\times$ \\
\midrule
$\tau_{\text{decode}}$ & 0.05            & 61.0 & 58.8 & 82.6 & 41.5 & 2.3$\times$ \\
$\tau_{\text{decode}}$ & 0.10 (default)  & 60.9 & 58.8 & 82.7 & 41.8 & 2.8$\times$ \\
$\tau_{\text{decode}}$ & 0.15            & 60.7 & 58.7 & 82.6 & 41.6 & 3.0$\times$ \\
\midrule
$T_{\text{refine}}$    & 5               & 60.8 & 58.7 & 82.7 & 41.7 & 2.7$\times$ \\
$T_{\text{refine}}$    & 10              & 60.9 & 58.7 & 82.7 & 41.7 & 2.8$\times$ \\
$T_{\text{refine}}$    & 15              & 60.9 & 58.7 & 82.7 & 41.8 & 2.8$\times$ \\
$T_{\text{refine}}$    & 20 (default)    & 60.9 & 58.8 & 82.7 & 41.8 & 2.8$\times$ \\
\bottomrule
\end{tabular}}
\end{table}

All three hyperparameters are robust. First, $\tau_{\text{refine}}$ varies accuracy by less than $0.5$ points across $[0.05, 0.15]$, so the Phase~1$\rightarrow$Phase~2 transition threshold is largely insensitive within a reasonable range. Second, $\tau_{\text{decode}}$ primarily controls the speed and accuracy trade-off: tighter thresholds preserve accuracy at the cost of speed (\rev{$2.3\times$}), while looser thresholds buy additional speed at modest accuracy cost (\rev{$3.0\times$}). Third, $T_{\text{refine}}$ is essentially irrelevant once large enough: accuracy varies by \rev{at most $0.1$ points} across $\{5,10,15,20\}$ because more than 85\% of samples converge in $\leq 3$ refinement steps and the cap is rarely reached (cf.\ \rev{Figure~\ref{fig:hotstart}}).

\begin{revision}
\paragraph{Robustness across window lengths.}
\label{appendix:cross_window}
\addcontentsline{toc}{subsubsection}{\protect\numberline{}Robustness across window lengths}

We extend the preceding length-512 sweeps to Dream-Instruct-7B at windows of 256 and 1024, changing only the listed parameter. Under the single-token commitment policy, the full commitment budget equals the window length; actual execution can stop earlier. The defaults remain $\tau_{\mathrm{refine}}=\tau_{\mathrm{decode}}=0.1$ and $T_{\mathrm{refine}}=20$, with $r_f=0.1$ in these sweeps. Average speed is the geometric mean across the four tasks. The defaults remain near the best accuracy across the tested windows, while $\tau_{\mathrm{decode}}$ controls the efficiency--quality trade-off. The refinement cap has little effect once it exceeds the number of steps needed by most inputs.

\begin{table}[htbp]
\ifshowchanges\color{blue}\captionsetup{font+={color=blue},labelfont+={color=blue}}\fi
\centering
\caption{Cross-window hyperparameter sweep on Dream-Instruct-7B at length 256.}
\label{tab:hyperparam256}
\small
\begin{tabular}{llccccc}
\toprule
Parameter & Value & HumanEval & MBPP & GSM8K & MATH500 & Avg. speed \\
\midrule
$\tau_{\mathrm{refine}}$ & 0.05 & 61.5 & 59.3 & 82.8 & 40.7 & 1.4$\times$ \\
& 0.10 & 61.6 & 59.4 & 83.0 & 40.6 & 1.5$\times$ \\
& 0.15 & 61.4 & 59.2 & 82.9 & 40.4 & 1.6$\times$ \\
\midrule
$\tau_{\mathrm{decode}}$ & 0.05 & 61.7 & 59.4 & 83.1 & 40.7 & 1.3$\times$ \\
& 0.10 & 61.6 & 59.4 & 83.0 & 40.6 & 1.5$\times$ \\
& 0.15 & 61.3 & 59.2 & 82.8 & 40.3 & 1.7$\times$ \\
\midrule
$T_{\mathrm{refine}}$ & 5 & 61.4 & 59.2 & 82.8 & 40.4 & 1.5$\times$ \\
& 10 & 61.5 & 59.3 & 82.9 & 40.5 & 1.5$\times$ \\
& 15 & 61.6 & 59.3 & 83.0 & 40.6 & 1.5$\times$ \\
& 20 & 61.6 & 59.4 & 83.0 & 40.6 & 1.5$\times$ \\
\midrule
$r_f$ & 0.05 & 60.8 & 58.9 & 82.5 & 40.0 & 1.4$\times$ \\
& 0.10 & 61.6 & 59.4 & 83.0 & 40.6 & 1.5$\times$ \\
& 0.15 & 61.5 & 59.3 & 82.9 & 40.5 & 1.5$\times$ \\
& 0.20 & 61.1 & 59.1 & 82.6 & 40.2 & 1.5$\times$ \\
\bottomrule
\end{tabular}
\end{table}

\begin{table}[htbp]
\ifshowchanges\color{blue}\captionsetup{font+={color=blue},labelfont+={color=blue}}\fi
\centering
\caption{Cross-window hyperparameter sweep on Dream-Instruct-7B at length 1024. The length-512 sweep is in Table~\ref{tab:hyperparam}.}
\label{tab:hyperparam1024}
\small
\begin{tabular}{llccccc}
\toprule
Parameter & Value & HumanEval & MBPP & GSM8K & MATH500 & Avg. speed \\
\midrule
$\tau_{\mathrm{refine}}$ & 0.05 & 60.9 & 58.9 & 83.3 & 44.0 & 5.1$\times$ \\
& 0.10 & 61.0 & 59.0 & 83.5 & 43.9 & 5.5$\times$ \\
& 0.15 & 60.8 & 58.8 & 83.4 & 43.7 & 5.8$\times$ \\
\midrule
$\tau_{\mathrm{decode}}$ & 0.05 & 61.1 & 58.9 & 83.6 & 44.0 & 4.6$\times$ \\
& 0.10 & 61.0 & 59.0 & 83.5 & 43.9 & 5.5$\times$ \\
& 0.15 & 60.7 & 58.8 & 83.2 & 43.6 & 6.1$\times$ \\
\midrule
$T_{\mathrm{refine}}$ & 5 & 60.8 & 58.8 & 83.3 & 43.7 & 5.4$\times$ \\
& 10 & 60.9 & 58.9 & 83.4 & 43.8 & 5.5$\times$ \\
& 15 & 61.0 & 58.9 & 83.5 & 43.9 & 5.5$\times$ \\
& 20 & 61.0 & 59.0 & 83.5 & 43.9 & 5.5$\times$ \\
\midrule
$r_f$ & 0.05 & 60.2 & 58.4 & 82.8 & 42.9 & 5.0$\times$ \\
& 0.10 & 61.0 & 59.0 & 83.5 & 43.9 & 5.5$\times$ \\
& 0.15 & 60.9 & 58.9 & 83.4 & 43.7 & 5.5$\times$ \\
& 0.20 & 60.6 & 58.7 & 83.1 & 43.4 & 5.4$\times$ \\
\bottomrule
\end{tabular}
\end{table}
\end{revision}

\FloatBarrier
\subsection{Limitations and Outlook}
\label{appendix:limitations}

\rev{LRD improves aggregate performance on the evaluated settings, not every individual response. Soft feedback may overwrite a locally correct symbol or stabilise an incorrect belief, and large mixing ratios can destabilise the model. At very small step budgets the refinement overhead may outweigh its speed benefit (Appendix~\ref{appendix:small_budgets}). Large-window headline speedups depend strongly on sequence-level early stopping; content-normalised gains without that stopping rule are smaller (Appendix~\ref{appendix:content_speed}). The stability results are conditional local bounds for soft updates, not guarantees of correctness, global contraction, or stability under arbitrary hard commitments.}

Diffusion language models remain an emerging generation paradigm. Compared with autoregressive LMs, the strongest open diffusion LMs are still largely research-lab releases rather than products backed by a mature inference ecosystem. As a result, their current absolute accuracy and latency may lag strong AR systems on some workloads, especially when AR models benefit from highly optimised serving stacks, mature KV-cache implementations, speculative decoding, quantisation, kernel-level optimisation, and extensive deployment tuning. Some diffusion-specific accelerators are also not yet uniformly supported across models and hardware backends, making end-to-end speed comparisons sensitive to implementation details.

These limitations should be interpreted as limitations of the present ecosystem rather than evidence against the paradigm. Diffusion LMs offer qualitatively different strengths: bidirectional conditioning, parallel token refinement, flexible commitment schedules, and the ability to improve uncertain positions before finalising discrete tokens. LRD targets this direction by improving the quality of the intermediate belief state while remaining training-free and composable with system-level accelerators. We expect the gap with AR systems to narrow as stronger open dLLMs, better diffusion-specific kernels, cache reuse, parallel decoding libraries, and serving infrastructure become available. In this sense, diffusion language modelling has substantial headroom, and methods such as LRD are intended to make that headroom more accessible by improving both quality and efficiency at inference time.

\FloatBarrier
\subsection{Broader Impacts}
\label{appendix:broader-impacts}

This work introduces Latent Refinement Decoding (LRD), a framework that significantly enhances the inference efficiency and reasoning reliability of diffusion large language models. By reducing generation steps by approximately 70\% and achieving speedups of up to $10.6\times$, LRD substantially lowers the computational cost and energy consumption of non-autoregressive generation, contributing to more sustainable AI deployment and improving accessibility of diffusion LMs for users with limited compute. While our method improves utility on critical tasks such as coding and mathematics, it is a training-free strategy that operates on existing pre-trained base models; therefore, it inherits their underlying biases and safety profiles, and faster inference can also lower the cost of producing misleading or harmful content. We therefore recommend that LRD be deployed alongside the standard alignment and safety guardrails of the underlying model, including content filtering, rate limiting, and provenance/watermarking schemes. Because LRD is training-free and modular, such mitigations remain fully applicable on top of it.

\FloatBarrier
\section{Theoretical Support: Stability, True Posterior, and Consistency}
\label{appendix:theory}

\FloatBarrier
\subsection{Stability Analysis of Mixed Embedding Updates}
\label{appendix:stability}

Our method operates in the embedding space rather than the discrete token space.
At each timestep $t$, we maintain a set of \textit{soft embeddings}
$\mathcal{E}_t = \{\tilde{\mathbf{e}}_t^{(1)}, \ldots, \tilde{\mathbf{e}}_t^{(L)}\}$ defined as
\begin{equation}
\tilde{\mathbf{e}}_t^{(i)} = (1-\alpha_t^{(i)}) \cdot \mathbf{e}_{\texttt{[MASK]}}
+ \alpha_t^{(i)} \cdot \sum_{v \in \mathcal{T}_{t+1}^{(i)}} \bar{p}_{t+1}^{(i)}(v) \cdot \mathbf{e}_v,
\label{eq:mix-embedding-appendix}
\end{equation}
where $\mathbf{e}_{\texttt{[MASK]}}$ denotes the \texttt{[MASK]} embedding, $\mathbf{e}_v$ denotes the embedding of token $v$, $\mathcal{T}_{t+1}^{(i)}$ is the top-$p$ nucleus set at position $i$, and $\bar{p}_{t+1}^{(i)}(v)$ is the renormalised predicted distribution over the nucleus set at position $i$.

To analyse stability, an ideal approach would be to examine the Jacobian of the update operator through its spectral radius. However, in practice this is intractable: transformer structures involve many linear and nonlinear components (layer normalisation, residual connections, multi-head attention), making it nearly impossible to provide a formal global analysis. The effective Jacobian inherits the complexity of the underlying transformer, and its spectral radius (or even its spectral norm) may be large and not easily bounded. As a result, although the iteration often stabilises empirically, a rigorous global convergence guarantee cannot be obtained.

Therefore, in this section, we follow the discussion from existing work \citep{yudin2025payattentionattentiondistribution,hu2024specformerguardingvisiontransformer,dasoulas2021lipschitznormalizationselfattentionlayers}
and focus on \emph{local} Lipschitz continuity. This analysis considers a \textbf{single self-attention layer} without any other operators. Please notice that due to the skip connection, the value addition across layers makes the contraction-like guarantees almost impossible for the transformer structure, however, it will still be beneficial to discuss the stability and Lipschitz continuity in a single-layer self-attention structure to provide intuition to support our method and explain empirical results rather than a complete mathematical proof.

Specifically, the local Lipschitz bound \cite{hu2024specformerguardingvisiontransformer} suggests that for all soft embedding $e_t$ within an $\epsilon$-ball around the origin (i.e.\ $\|e_t \| \leq \epsilon$), where $\epsilon$ bounds the maximum norm of embeddings, the following inequality holds after one-layer self-attention mapping:

\begin{equation}
    \|\mathbf{e}_{t+1}^{s} - \mathbf{e}_{t}^{s}\|_2 \leq K \|\mathbf{e}_{t+1} - \mathbf{e}_{t}\|_2,
\end{equation}

where $\mathbf{e}_{t}^{s}$ is the output of $\mathbf{e}_{t}$ after one-layer self-attention mapping, $K$ is the local Lipschitz constant. \rev{A schematic single-layer bound must include both the value pathway and the input-dependent attention pathway:}
\begingroup\ifshowchanges\color{blue}\fi
\begin{equation}
    K(\epsilon) \;\lesssim\; c_0\|\mathbf{W}^V_h\|_2
    +c_1\|\mathbf{W}^V_h\|_2\|\mathbf{W}^Q_h(\mathbf{W}^K_h)^\top\|_2\epsilon^2,
\end{equation}
\endgroup
\rev{where $c_0,c_1$ depend on the sequence geometry, attention scaling, and chosen norms. Small $\epsilon$ can reduce the input-dependent term, but the value term remains; it does not force $K<1$. This single-layer intuition is separate from the assumed sequence-level confidence condition in Definition~\ref{def:local-lip-seq}.}

\begin{figure*}
    \centering
\includegraphics[width=0.7 \linewidth]{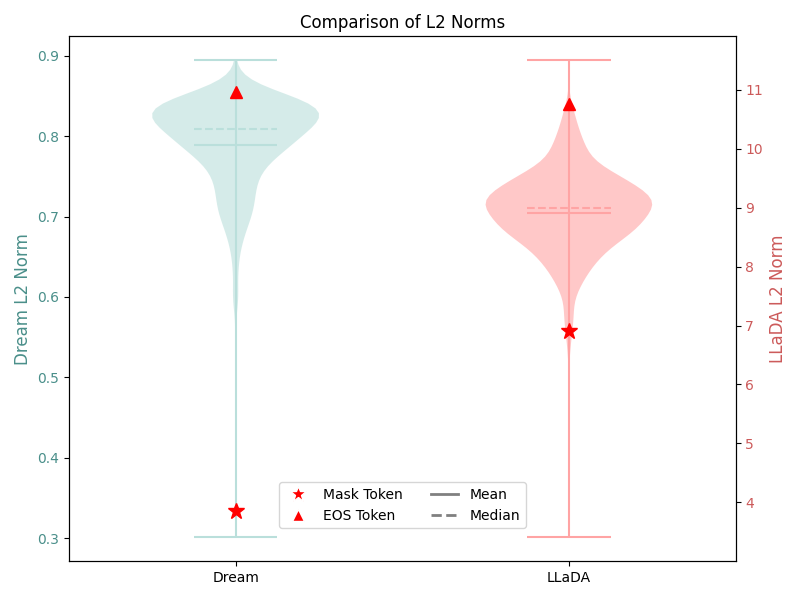}
\caption{Distribution of L2 norms for token embeddings in Dream and LLaDA models. }
    \label{fig:l2_norms}
\end{figure*}

% Here, we consider only a single layer of self-attention then there is no linear addition of value term, which means we only discuss the potential amplification effect on the embedding norm.
Here, we consider only a single layer of self-attention without the residual (skip) connection, which means we only discuss the potential amplification effect of the self-attention mapping on the embedding norm.
Then, the ideal outcome of such a mapping would be a less expansive function with a small $K$, which ensures a locally stable region. However, in transformer blocks the large parameter norms often make this condition difficult to satisfy. Since $\mathbf{W}^Q_h$, $\mathbf{W}^K_h$, and $\mathbf{W}^V_h$ are fixed for a pretrained model, in decoding only structure, the stability in practice can only rely on keeping $\epsilon$ sufficiently small, which is under our control. One fact is that the \texttt{[MASK]} embedding has a very small $\ell_2$ norm of $0.3340$ in 3,584 dimensions (corresponding to a per-dimension RMS of about $0.0055$), regular token embeddings are much larger. %For instance, a typical token embedding has an $\ell_2$ norm of about $0.8721$, which corresponds to an average per-dimension RMS magnitude of approximately $0.0142$.
Figure~\ref{fig:l2_norms} shows the distribution of L2 norms across all token embeddings in both Dream and LLaDA models, confirming that the \texttt{[MASK]} token consistently exhibits substantially lower norms than regular tokens, validating our design choice to use \texttt{[MASK]} as a stable low-norm reference point for mixed embedding updates. Therefore, to guarantee the updated embedding is small enough, we need to ensure the injected norm is significantly small as well. This motivates us to restrict the update within a small weight regarding the injected embeddings.

To connect this bound back to the embedding updates, we require $\tilde{\mathbf{e}}_t^{(i)}$ and $\tilde{\mathbf{e}}_{t+1}^{(i)}$ to lie within an $\epsilon$-ball around the origin, which requires a very small $\epsilon$. Since both are formed as weighted sums of the \texttt{[MASK]} embedding and candidate token embeddings (Equation~\ref{eq:mix-embedding-appendix}), a straightforward way to reduce this distance is to bound the mixing coefficient $\alpha_t^{(i)}$.
Concretely, by the triangle inequality and $\sum_{v \in \mathcal{T}_{t+1}^{(i)}} \bar{p}_{t+1}^{(i)}(v)=1$,
\begin{equation}
\|\tilde{\mathbf{e}}_t^{(i)}\|_2
\le (1-\alpha_t^{(i)})\|\mathbf{e}_{\texttt{[MASK]}}\|_2
+ \alpha_t^{(i)} \max_{v \in \mathcal{T}_{t+1}^{(i)}} \|\mathbf{e}_v\|_2,
\end{equation}
so choosing a sufficiently small $\alpha_t^{(i)}$ keeps $\tilde{\mathbf{e}}_t^{(i)}$ within a small-norm neighbourhood.
Intuitively, this means the search for efficient mixed embeddings remains close to the \texttt{[MASK]} token, with exploration constrained to a small neighbourhood. \rev{This limits input perturbations and empirically yields stable predictive distributions, without establishing a contraction region.}

To simplify, we introduce a base rate $r_f$ and set
$\alpha_t^{(i)} = r_f \cdot \bigl(1 - \bar{H}_{t+1}^{(i)}\bigr)$,
where $\bar{H}_{t+1}^{(i)} \in [0,1]$ is the normalised entropy. Intuitively, higher entropy (more uncertainty) yields a smaller $\alpha_t^{(i)}$, keeping the representation closer to $[\text{MASK}]$, while lower entropy increases $\alpha_t^{(i)}$ and places more weight on the belief-weighted embedding.
% Since $\max_i \alpha_t^{(i)} \leq r_f$, ensuring the difference is within $\epsilon$ reduces to choosing a sufficiently small $r_f$.
Since $\max_i \alpha_t^{(i)} \leq r_f$, ensuring $\|\tilde{\mathbf{e}}_t^{(i)}\|_2 \le \epsilon$ (and similarly for $t{+}1$) reduces to choosing a sufficiently small $r_f$.
Empirically, we find that the method is stable and effective when $r_f$ is small, but fails to converge for large $r_f$ (see Figure~\ref{fig:base_rate}).

We further evaluate the stability of output embeddings before the logit prediction step across adjacent timesteps. Since the token space is sparse and high-dimensional, we use the KL divergence as the metric. This reveals clear convergence during the latent refinement phase when $r_f$ is small, even after deep iteration with multi-layer self-attention in a transformer (see Figure~\ref{fig:hotstart}).

In summary, although a rigorous global convergence guarantee for mixed embedding iterations is intractable due to the nonlinear, high-capacity nature of transformers, our local Lipschitz analysis provides useful theoretical insight. Together with empirical validation, this suggests that while strict guarantees remain challenging, the proposed method is practically stable and effective for reasoning with diffusion LLMs.

\begin{revision}
\FloatBarrier
\subsubsection{Feedback-Operator Stress Test}
\label{appendix:operator_stress}

We apply SM-, Evo-, and RCD-style feedback operators to the same frozen LLaDA-Instruct-8B decoder on GSM8K. These are operator-style stress tests, not evaluations of the corresponding trained methods. Let $\hat\rho_t=\|\Delta\tilde{\mathcal E}_t\|_{2,\infty}$ be the empirical perturbation over unresolved positions, and define the confidence-drop rate $\mathrm{CR}=\Pr[c_i^{t+1}<c_i^t-0.05]$. Table~\ref{tab:operator_stress} reports averages over steps and examples as feedback strength increases from mild to strongest.

\begin{table}[htbp]
\ifshowchanges\color{blue}\captionsetup{font+={color=blue},labelfont+={color=blue}}\fi
\centering
\caption{Feedback strength, embedding perturbation, and confidence drops in a frozen-decoder stress test. Accuracy is a fraction; CR is in units of $10^{-3}$.}
\label{tab:operator_stress}
\small
\begin{tabular}{lccc}
\toprule
Operator & $\hat\rho$: mild $\to$ strongest & CR: mild $\to$ strongest & Accuracy at strongest \\
\midrule
LRD & $0.03\to12.43$ & $0.2\to232.0$ & 0.14 \\
SM-style & $0.29\to11.78$ & $0.7\to213.4$ & 0.19 \\
Evo-style & $0.21\to10.62$ & $1.5\to233.1$ & 0.19 \\
RCD-style & $0.40\to15.79$ & $2.2\to331.8$ & 0.00 \\
\bottomrule
\end{tabular}
\end{table}

Strong feedback increases embedding movement and confidence drops across all tested operators, including LRD. The analysis therefore motivates controlling feedback strength rather than claiming that a particular mixture is universally stable. This stress test supports that qualitative failure mechanism; it is not a numerical verification of every Lipschitz or progress assumption in the conditional TCR bounds. Training may make stronger feedback compatible with a model, whereas LRD controls feedback in frozen-model inference.
\end{revision}

\FloatBarrier
\subsection{Derivation of the True Posterior in the Masking Process}
\label{appendix:posterior-derivation}
We derive Eq.~\ref{eq:true-posterior} for the true posterior distribution in the absorbing masking forward process.
For each position $i$, the forward process is defined as
\[
\Pr(x_t^{(i)} = x_0^{(i)} \mid x_0^{(i)}) = \alpha_t^*,
\qquad
\Pr(x_t^{(i)} = \texttt{[MASK]} \mid x_0^{(i)}) = 1-\alpha_t^*,
\]
with $(\alpha_t^*)_{t=0}^T$ monotonically decreasing. Thus each token can only either remain as its original value $x_0^{(i)}$ or transition to the
special token \texttt{[MASK]}. By Bayes' rule,
\begin{equation}
q^*(x_{t-1}^{(i)} \mid x_t^{(i)}=\texttt{[MASK]}, x_0^{(i)})
= \frac{\Pr(x_t^{(i)}=\texttt{[MASK]} \mid x_{t-1}^{(i)}, x_0^{(i)})
\Pr(x_{t-1}^{(i)} \mid x_0^{(i)})}{\Pr(x_t^{(i)}=\texttt{[MASK]} \mid x_0^{(i)})}.
\end{equation}
There are two possible values for $x_{t-1}^{(i)}$:
\begin{itemize}
    \item
    The probability of $x_{t-1}^{(i)} = x_0^{(i)}$ is $\alpha_{t-1}^*$, and transitioning to mask at step $t$ occurs with probability $1 - \tfrac{\alpha_t^*}{\alpha_{t-1}^*}$.
    Hence the joint probability is $\alpha_{t-1}^* - \alpha_t^*$.
    \item The probability of $x_{t-1}^{(i)} = \texttt{[MASK]}$ is $1-\alpha_{t-1}^*$, and once masked, the token remains masked with probability $1$.
    Hence the joint probability is $1-\alpha_{t-1}^*$.
\end{itemize}
The marginal probability of being masked at step $t$ is $Pr(x_t^{(i)}=\texttt{[MASK]} \mid x_0^{(i)}) = 1 - \alpha_t^*$. So we obtain
\[
q^*(x_{t-1}^{(i)} \mid x_t^{(i)}=\texttt{[MASK]}, x_0^{(i)})
= \frac{\alpha_{t-1}^*-\alpha_t^*}{1-\alpha_t^*}\,\delta_{x_0^{(i)}}
+ \frac{1-\alpha_{t-1}^*}{1-\alpha_t^*}\,\delta_{\texttt{[MASK]}}.
\]

\FloatBarrier
\subsection{Self-consistency view of latent refinement}
\label{appendix:fixed_point_view}

We interpret Phase~1 (\S\ref{sec:method}) as an \emph{iterative self-consistency} refinement carried out at a fixed diffusion stage, rather than an additional reverse-time discretisation.
In Phase~1, at refinement step $t$ and for each position $i\in\{1,\dots,L\}$, we maintain a soft embedding $\tilde{\mathbf e}_t^{(i)}$ defined in Eq.~\ref{eq:mix-embedding}.
Given the current soft embeddings $\{\tilde{\mathbf e}_t^{(i)}\}_{i=1}^L$, the dLLM outputs token marginals $p_t^{(i)}(\cdot)$ over the vocabulary at each position.
We then form the top-$p$ nucleus set $\mathcal{T}_{t+1}^{(i)}$ and its renormalised distribution $\bar p_{t+1}^{(i)}(\cdot)$ (as in Eq.~\ref{eq:mix-embedding}), and project this distribution back to an expected token embedding
$\sum_{v\in\mathcal{T}_{t+1}^{(i)}}\bar p_{t+1}^{(i)}(v)\mathbf e_v$.
Finally, we mix this expected embedding with the mask embedding $\mathbf e_{[\text{MASK}]}$ using the entropy-based coefficient $\alpha_t^{(i)}$ (Eq.~\ref{eq:alpha}), yielding the updated soft embedding $\tilde{\mathbf e}_{t+1}^{(i)}$ for the next refinement step.

Equivalently, Eq.~\ref{eq:mix-embedding} defines an iterative update mapping on the soft state.
Let $\tilde{\mathcal E}_t := \{\tilde{\mathbf e}_t^{(i)}\}_{i=1}^L$ denote the collection of soft embeddings over all positions at refinement step $t$.
We define the one-step refinement operator $\Phi_\theta$ by
\[
\tilde{\mathcal E}_{t+1} \;=\; \Phi_\theta(\tilde{\mathcal E}_t),
\]
where $\Phi_\theta$ denotes: (i) run the dLLM to obtain $\{p_t^{(i)}(\cdot)\}_{i=1}^L$ conditioned on $\tilde{\mathcal E}_t$, and (ii) update each position according to Eq.~\ref{eq:mix-embedding}.
Note that $\Phi_\theta$ includes entropy-based gating and top-$p$ truncation, and we do not assume global contraction or a formal convergence guarantee in embedding space.

Phase~1 applies this operator repeatedly until the distributional residual between consecutive predictions becomes small, measured by the KL gap in Eq.~\ref{eq:kl-gap}:
\[
\mathcal D_{\mathrm{KL}}^{(t)}=\frac{1}{L}\sum_{i=1}^L D_{\mathrm{KL}}\!\left(p_t^{(i)} \,\|\, p_{t+1}^{(i)}\right).
\]
When $\mathcal D_{\mathrm{KL}}^{(t)}<\tau_{\mathrm{refine}}$, the predicted marginals change little under an additional refinement update. We treat this as a practical surrogate for reaching an approximately self-consistent belief state under the current soft context (rather than a strict convergence certificate in embedding space).

Crucially, this refinement is intended to reduce mismatch between the soft state $\tilde{\mathcal E}_t$ and the model-implied predictive distributions at the \emph{same} diffusion stage, rather than to advance the diffusion time index.
After this initialisation, Phase~2 performs discrete commitments while continuing the usual denoising trajectory.
This perspective is related in spirit to iterative self-consistency procedures used in implicit-depth models and to diffusion formulations that incorporate inner-loop refinement within denoising dynamics. \citep{bai2019deq,bai2024fpdm}

\begin{revision}
\paragraph{Relation to predictor--corrector sampling.}
\label{appendix:corrector_connection}
The Diffusion Duality~\citep{sahoo2025duality} connects uniform-state discrete diffusion to Gaussian relaxations, while $\Psi$-samplers~\citep{deschenaux2026psi} introduce predictor--corrector mechanisms for discrete diffusion. A standard absorbing masked sampler cannot revise a token after irreversible commitment; LRD delays such commitment and lets unresolved positions revise one another through soft context. At a fixed diffusion stage, its per-position mixture can be written as
\[
q_{\alpha}(\cdot)=\alpha\bar p+(1-\alpha)\delta_{\texttt{[MASK]}},
\qquad
\tilde{\mathbf e}=\mathbb E_{q_{\alpha}}[\mathbf e_x].
\]
This has the mixture form of a mask-corruption marginal with the unavailable clean token replaced by the current truncated belief, and LRD feeds its embedding first moment back to the denoiser. In this functional sense, Phase~1 resembles deterministic mean-field corrector-like refinement. The correspondence does not identify an exact reverse-time transition kernel: LRD neither samples a predictor--corrector transition nor remasks committed tokens, and its entropy-adaptive weight is not a calibrated diffusion-time coefficient. It therefore does not inherit the marginal-preservation guarantees of a predictor--corrector sampler. Small embedding norms motivate local stability but do not establish that mixed inputs are in the model's training distribution. The matched-state probe in Appendix~\ref{appendix:matched_state} tests their empirical usefulness separately.

\end{revision}

\FloatBarrier
\subsection{Proofs and empirical verification of the TCR bounds}
\label{appendix:tcr-proofs}

This section gives complete proofs for the statements in Section~\ref{sec:tcr}: the per-step embedding bound (Lemma~\ref{lem:rho}), the high-probability bound on backward updates (Theorem~\ref{thm:back}), and the lower bound on expected TCR (Corollary~\ref{cor:tcr}). \rev{We also describe a protocol for measuring the quantities in Definition~\ref{def:local-lip-seq} and the gap condition in Corollary~\ref{cor:tcr}; this protocol is distinct from the qualitative stress test in Appendix~\ref{appendix:operator_stress}.} Figure~\ref{fig:tcr-verification} visualises the two quantities used in this verification: the signed gap to the commitment threshold and the resulting threshold-oriented convergence rate.

\rev{\textbf{Scope of the perturbation bound.} The bound $\rho$ applies to soft-to-soft updates with the committed context held fixed. For an actual Phase~2 transition, let $\mathcal E_t$ include both soft and committed embeddings. Its input perturbation is $\|\mathcal E_{t+1}-\mathcal E_t\|_{2,\infty}$, which also includes any newly committed token. A soft-to-hard jump can remain non-zero as $r_f\to0$, so it is not controlled by the soft-only $\rho$. Applying a confidence bound to that transition requires a Lipschitz condition for the full input and its actual perturbation. The plots of threshold progress are empirical evidence, not verification of every assumption or of a positive numerical lower bound.}

\begin{figure}[htbp]
    \centering

    \begin{subfigure}{0.48\textwidth}
        \centering
        \includegraphics[width=\textwidth]{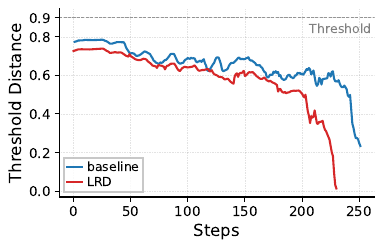}
        \caption{Signed gap to the commitment threshold}
        \label{fig:commit-threshold-gap}
    \end{subfigure}
    \hfill
    \begin{subfigure}{0.48\textwidth}
        \centering
        \includegraphics[width=\textwidth]{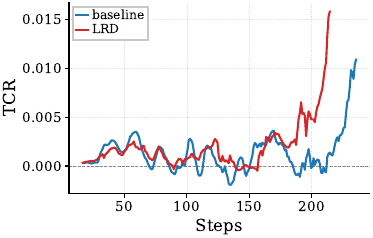}
        \caption{Threshold-oriented convergence rate}
        \label{fig:tcr-curve}
    \end{subfigure}

    \caption{Empirical quantities for TCR verification on LLaDA (on MBPP dataset, commitment threshold = 0.9). Left: the average signed gap to the commitment threshold, where lower/high values indicate positions moving closer to/away from commitment. Right: the corresponding TCR curve, where positive/negative values indicate progress toward/backward commitment. Both figures demonstrate our framework is able to speed-up the commitment in parallel decoding.}
    \label{fig:tcr-verification}
\end{figure}

\paragraph{Proof of Lemma~\ref{lem:rho}.}

\begin{revision}
Recall the soft-embedding rule
$\tilde{\mathbf e}_i^{(t)}=\mathbf e_{\texttt{[MASK]}}+\alpha_t^{(i)}(\mathbf m_t^{(i)}-\mathbf e_{\texttt{[MASK]}})$,
where $\mathbf m_t^{(i)}$ is a convex combination of token embeddings and $\alpha_t^{(i)}\in[0,r_f]$. Hence $\|\mathbf m_t^{(i)}\|_2\le E_{\max}$, and
\begin{equation}
\|\tilde{\mathbf e}_i^{(t)}-\mathbf e_{\texttt{[MASK]}}\|_2
\le r_f(E_{\max}+E_{\texttt{[MASK]}}).
\label{eq:lem-rho-detail}
\end{equation}
Applying the triangle inequality through the same mask anchor at two consecutive steps gives
\[
\|\tilde{\mathbf e}_i^{(t+1)}-\tilde{\mathbf e}_i^{(t)}\|_2
\le 2r_f(E_{\max}+E_{\texttt{[MASK]}})=\rho.
\]
For distribution-sensitive estimates, the bounded-expectation inequality is
\begin{equation}
\Big\|\sum_v(p(v)-q(v))\mathbf e_v\Big\|_2
\le 2E_{\max}\,\mathrm{TV}(p,q),
\label{eq:coupling-tv}
\end{equation}
where $\mathrm{TV}(p,q)=\tfrac12\sum_v|p(v)-q(v)|$. The factor 2 follows from this convention for total variation; the mask-anchor proof above establishes the same stated $\rho$ without that inequality.\hfill$\square$
\end{revision}

\begin{remark}[Tighter local bound]\label{rem:tighter}
In late refinement steps the nucleus narrows and adjacent token embeddings are highly correlated. Letting \rev{$D_t^{(i)}$ be the diameter of the token embeddings over $\operatorname{supp}(\bar p_t^{(i)})\cup\operatorname{supp}(\bar p_{t+1}^{(i)})$} and $\bar{\mathbf{e}}_t^{(i)}\coloneqq\sum_v\bar p_t^{(i)}(v)\mathbf{e}_v$, the same argument gives the tighter bound
\(
\|\tilde{\mathbf e}_i^{(t+1)}-\tilde{\mathbf e}_i^{(t)}\|_2
\le |\Delta\alpha_t^{(i)}|\,\|\bar{\mathbf{e}}_t^{(i)}-\mathbf{e}_{\texttt{[MASK]}}\|_2
+\alpha_{t+1}^{(i)}D_t^{(i)}\mathrm{TV}\!\big(\bar p_{t+1}^{(i)},\bar p_t^{(i)}\big),
\)
which is typically much smaller than $\rho$ once the model is close to commitment.
\end{remark}

\paragraph{Proof of Theorem~\ref{thm:back}.}
Let $\mathcal{L}_i^{(t)}$ denote the Lipschitz event
$\big\{|c_i^{(t+1)}-c_i^{(t)}|\le K\,\|\Delta\tilde{\mathcal E}_t\|_{2,\infty}\big\}$. By Definition~\ref{def:local-lip-seq}, $\Pr_{\mathcal{P}}(\mathcal{L}_i^{(t)})\ge 1-\eta$. Lemma~\ref{lem:rho} applied position-wise gives $\|\Delta\tilde{\mathcal E}_t\|_{2,\infty}\le\rho$, so on $\mathcal{L}_i^{(t)}$,
\(
|c_i^{(t+1)}-c_i^{(t)}|\le K\rho.
\)
The map $c\mapsto[\tau-c]_+$ is $1$-Lipschitz, so $|d_i^{(t+1)}-d_i^{(t)}|\le K\rho$ on the same event. Therefore the event $\{d_i^{(t+1)}-d_i^{(t)}>\delta\}$ with $\delta>K\rho$ implies the failure $(\mathcal{L}_i^{(t)})^c$, whose probability is at most $\eta$. The expectation bound follows by linearity over $i\in M_t$.\hfill$\square$

\paragraph{Proof of Corollary~\ref{cor:tcr}.}
Recall the progress event
$G_i^{(t)}\coloneqq\{c_i^{(t+1)}-c_i^{(t)}\ge\gamma\}\cap\{c_i^{(t)}\le\tau-\gamma\}$
and the Lipschitz event $\mathcal{L}_i^{(t)}$ from the proof of Theorem~\ref{thm:back}, with $\Pr_{\mathcal{P}}(G_i^{(t)})\ge q_t$ and $\Pr_{\mathcal{P}}((\mathcal{L}_i^{(t)})^c)\le\eta$. The three events $G_i^{(t)}\cap\mathcal{L}_i^{(t)}$, $(G_i^{(t)})^c\cap\mathcal{L}_i^{(t)}$, $(\mathcal{L}_i^{(t)})^c$ are mutually exclusive and exhaustive. We bound $d_i^{(t)}-d_i^{(t+1)}$ on each:

\emph{(i) On $G_i^{(t)}\cap\mathcal{L}_i^{(t)}$.} The side condition $c_i^{(t)}\le\tau-\gamma$ gives $d_i^{(t)}=\tau-c_i^{(t)}\ge\gamma$. If $c_i^{(t+1)}\ge\tau$ then $d_i^{(t+1)}=0$ and $d_i^{(t)}-d_i^{(t+1)}=\tau-c_i^{(t)}\ge\gamma$; otherwise $d_i^{(t+1)}=\tau-c_i^{(t+1)}$ and $d_i^{(t)}-d_i^{(t+1)}=c_i^{(t+1)}-c_i^{(t)}\ge\gamma$. Either way $d_i^{(t)}-d_i^{(t+1)}\ge\gamma$.

\emph{(ii) On $(G_i^{(t)})^c\cap\mathcal{L}_i^{(t)}$.} Lemma~\ref{lem:rho} gives $\|\Delta\tilde{\mathcal E}_t\|_{2,\infty}\le\rho$, so on $\mathcal{L}_i^{(t)}$ we have $|c_i^{(t+1)}-c_i^{(t)}|\le K\rho$. Since $[\tau-\cdot]_+$ is $1$-Lipschitz, $|d_i^{(t+1)}-d_i^{(t)}|\le K\rho$, hence $d_i^{(t)}-d_i^{(t+1)}\ge -K\rho$.

\emph{(iii) On $(\mathcal{L}_i^{(t)})^c$.} Since $d_i^{(t)},d_i^{(t+1)}\in[0,\tau]$, trivially $d_i^{(t)}-d_i^{(t+1)}\ge-\tau$.

Combining the three cases via the indicator decomposition,
\begin{equation}
d_i^{(t)}-d_i^{(t+1)}\;\ge\;\gamma\,\mathbf{1}\!\big[G_i^{(t)}\cap\mathcal{L}_i^{(t)}\big]
\;-\;K\rho\,\mathbf{1}\!\big[(G_i^{(t)})^c\cap\mathcal{L}_i^{(t)}\big]
\;-\;\tau\,\mathbf{1}\!\big[(\mathcal{L}_i^{(t)})^c\big].
\label{eq:cor-decomp}
\end{equation}
Taking expectations under $\mathcal{P}$ and using the Bonferroni-type bound
$\Pr_{\mathcal{P}}(G_i^{(t)}\cap\mathcal{L}_i^{(t)})\ge\Pr_{\mathcal{P}}(G_i^{(t)})+\Pr_{\mathcal{P}}(\mathcal{L}_i^{(t)})-1\ge q_t-\eta$,
together with $\Pr_{\mathcal{P}}((G_i^{(t)})^c\cap\mathcal{L}_i^{(t)})\le\Pr_{\mathcal{P}}((G_i^{(t)})^c)\le 1-q_t$ and $\Pr_{\mathcal{P}}((\mathcal{L}_i^{(t)})^c)\le\eta$, we obtain
\(
\mathbb{E}\big[d_i^{(t)}-d_i^{(t+1)}\big]\ge\gamma(q_t-\eta)-K\rho(1-q_t)-\tau\eta.
\)
Averaging over $i\in M_t$ yields the claimed bound on $\mathbb{E}[\mathrm{TCR}^{(t)}]$.\hfill$\square$

\paragraph{Empirical estimation of $\eta$.}
The failure probability $\eta$ in Definition~\ref{def:local-lip-seq} is operationally defined; the following four-step protocol turns it into a measurement.

\begin{enumerate}[leftmargin=2em,topsep=2pt,itemsep=2pt]
\item Fix $r_f$ and let $\rho=2r_f(E_{\max}+E_{\texttt{[MASK]}})$.
\item Sample $N$ triples $(\text{prompt},i,t)$ from the validation set during LRD decoding, recording $c_i^{(t)},c_i^{(t+1)}$ and the sequence-level perturbation $\|\Delta\tilde{\mathcal E}_t\|_{2,\infty}$.
\item For each candidate $K$, compute the empirical failure rate
\[
\hat\eta(K)=\frac{1}{N}\sum_{n=1}^{N}\mathbf{1}\!\Big[\big|c_{i_n}^{(t_n+1)}-c_{i_n}^{(t_n)}\big|>K\,\big\|\Delta\tilde{\mathcal E}_{t_n}\big\|_{2,\infty}\Big].
\]
\item Choose $K^*$ as the smallest $K$ such that $\hat\eta(K)\le\eta_0$ for a target tolerance $\eta_0$ (we use $\eta_0=0.05$).
\end{enumerate}

This procedure (i)~grounds Definition~\ref{def:local-lip-seq} in observable quantities, (ii)~makes the trade-off between $K$ and $\eta$ explicit, and (iii)~converts the high-probability statement from a postulate into a measurement that depends only on $r_f$ and the validation distribution.

\paragraph{Verifying the gap condition.}
On the same validation triples, compute the empirical progress probability
\(
\hat q_t(\gamma)=\tfrac{1}{N}\sum_n\mathbf{1}\!\big[c_{i_n}^{(t_n+1)}-c_{i_n}^{(t_n)}\ge\gamma\;\wedge\;c_{i_n}^{(t_n)}\le\tau-\gamma\big]
\)
for $\gamma\in\{0.01,0.05,0.10\}$, mirroring the side condition in the progress event $G_i^{(t)}$ of Corollary~\ref{cor:tcr}. \rev{Together, these measurements would permit numerical evaluation of the lower bound in Corollary~\ref{cor:tcr}.}

\paragraph{Interpretation.}
The pair $(\rho,\eta)$ provides an honest design dial: shrinking $r_f$ shrinks both $\rho$ (mechanically, via Lemma~\ref{lem:rho}) and typically $\eta$ (because confidence becomes a smoother function of the input perturbation), at the cost of a smaller per-step gap $\gamma$. Conversely, large $r_f$ makes $q_t\gamma$ large but inflates $\eta$. This trade-off matches what is observed in our base-rate ablation (Figure~\ref{fig:base_rate}), where the method is stable for small $r_f$ and diverges for large $r_f$, and motivates the choice $r_f\in[0.1,0.2]$ used throughout the experiments.

\clearpage

\end{document}